%% file: main.tex
\documentclass[11pt]{report}

\input{preamble/packages}
\input{preamble/macros}

\begin{document}

% Custom title page
\begin{titlepage}
    \begin{center}
    \vspace*{6em}
    {\LARGE\bfseries Early science acceleration experiments with GPT-5\par}
    \vspace{1em}
    Sébastien Bubeck$^{1}$, Christian Coester$^{2}$, Ronen Eldan$^{1}$, Timothy Gowers$^{3}$, Yin Tat Lee$^{1}$, \par
    Alexandru Lupsasca$^{1,4}$, Mehtaab Sawhney$^{5}$, Robert Scherrer$^{4}$, Mark Sellke$^{1,6}$, \par Brian K. Spears$^{7}$,
    Derya Unutmaz$^{8}$, Kevin Weil$^{1}$, Steven Yin$^{1}$, Nikita Zhivotovskiy$^{9}$
    \vspace{1em}
    {\small \\
        $^{1}$OpenAI \\
        $^{2}$University of Oxford \\
        $^{3}$Collège de France and University of Cambridge \\
        $^{4}$Vanderbilt University \\
        $^{5}$Columbia University \\
        $^{6}$Harvard University \\
        $^{7}$Lawrence Livermore National Laboratory \\
        $^{8}$The Jackson Laboratory \\
        $^{9}$University of California, Berkeley \\
    }

    \vspace{2em}
    November 20, 2025
    \vspace{3em}

    {\bfseries Abstract}\par
    \vspace{1em}
    \end{center}
    \noindent
    AI models like GPT-5 are an increasingly valuable tool for scientists, but many remain unaware of the capabilities of frontier AI. We present a collection of short case studies in which GPT-5 produced new, concrete steps in ongoing research across mathematics, physics, astronomy, computer science, biology, and materials science. In these examples, the authors highlight how AI accelerated their work, and where it fell short; where expert time was saved, and where human input was still key. We document the interactions of the human authors with GPT-5, as guiding examples of fruitful collaboration with AI. Of note, this paper includes four new results in mathematics (carefully verified by the human authors), underscoring how GPT-5 can help human mathematicians settle previously unsolved problems. These contributions are modest in scope but profound in implication, given the rate at which frontier AI is progressing.
    \vfill
\end{titlepage}

\renewcommand{\thechapter}{\Roman{chapter}}
\setcounter{tocdepth}{1}
\tableofcontents

% IMPORTANT: use \subfile, not \input

\subfile{content/00-Introduction}

\newpage

\chapter{\,Independent rediscovery of known results at the scientific frontier}
\label{sec:IndependentDiscovery}

\subfile{content/01-Bubeck}

\newpage

\subfile{content/02-Lupsasca}

\newpage

\subfile{content/03-Unutmaz}

\newpage

\chapter{\,Deep literature search}
\label{sec:DeepLiteratureSearch}

\subfile{content/04-Zhivotovskiy}

\newpage

\subfile{content/05-SawhneySellke}

\newpage

\subfile{content/12-Anonymous}

\newpage

\chapter{\,Working in tandem with AI}
\label{sec:WorkingWithAI}

\subfile{content/06-Gowers}

\newpage

\subfile{content/07-Scherrer}
\newpage

\subfile{content/08-Spears}

\newpage

\chapter{\,New scientific results obtained with AI}
\label{sec:NewResults}

\subfile{content/09-SawhneySellke}

\newpage

\subfile{content/10-Coester}

\newpage

\subfile{content/11-BubeckSellkeYin.tex}

\newpage

\subfile{content/11.5-BubeckSellkeYin.tex}

\newpage

\subfile{content/99-Conclusion}

\DeclareNameAlias{sortname}{given-family}
\DeclareNameAlias{author}{given-family}

\printbibliography

\end{document}

%% file: preamble/packages.tex
\usepackage[margin=1in]{geometry}
\usepackage{amsmath, amssymb, amsthm}
\usepackage{microtype}
\usepackage{fullpage}
\usepackage[utf8]{inputenc}
\usepackage{csquotes}
\usepackage[colorlinks=true,allcolors=blue]{hyperref}
\usepackage[nameinlink,noabbrev]{cleveref}
\usepackage{graphicx}
\usepackage{subcaption}
\usepackage{booktabs}
\usepackage{mathtools}
\usepackage{siunitx}
\usepackage{wrapfig}
\usepackage{dsfont, a4wide, amsfonts}
\usepackage{fancyhdr, xspace, psfrag, setspace, supertabular, color}
\usepackage{bbm}
\usepackage{stmaryrd}
\usepackage{txfonts}
\usepackage{enumerate}
\usepackage{enumitem}
\usepackage{tikz}
\usepackage{subfiles}
\usepackage{authblk}
\usepackage[most]{tcolorbox}
\usepackage[sorting=nyt,backend=biber,maxbibnames=99,citestyle=alphabetic,style=alphabetic]{biblatex}
\usepackage{xcolor}
\usepackage{pdfpages}
\usepackage{physics}
\usepackage{upgreek}
\usepackage{microtype}
\usepackage{lmodern}
\usepackage[framemethod=TikZ]{mdframed}
\usepackage[T1]{fontenc}
\usepackage{listings}
\usetikzlibrary{positioning,arrows.meta,fit}

\usetikzlibrary{calc,angles,quotes,arrows.meta,decorations.markings}

\usepackage{float}

%% file: preamble/macros.tex
\newcommand{\secprefix}{}

\newtheorem{theorem}{Theorem}
\newtheorem{corollary}{Corollary}
\newtheorem{conjecture}{Conjecture}
\newtheorem{lemma}{Lemma}
\newtheorem{proposition}{Proposition}
\newtheorem{remark}{Remark}
\numberwithin{theorem}{section}
\numberwithin{corollary}{section}
\numberwithin{lemma}{section}
\numberwithin{proposition}{section}
\numberwithin{remark}{section}

\newcommand{\weirdplus}{\ensuremath{{}^{+}}}
\newcommand{\weirdminus}{\ensuremath{{}^{-}}}

\let\originalleft\left
\let\originalright\right
\renewcommand{\left}{\mathopen{}\mathclose\bgroup\originalleft}
\renewcommand{\right}{\aftergroup\egroup\originalright}
\mathcode`\*="8000
{\catcode`\*=\active\gdef*{\mathclose{}\,\mathopen{}}}

\newcommand{\br}[1]{\left[#1\right]}

\newcommand{\pd}{\mathop{}\!\partial}

\def\R{\mathbb{R}}
\def\T{\mathbb{T}}

\def\N{\mathbb{N}}
\def\P{\mathbb{P}}

\def\b1{\mathbbm{1}}

\tcbset{
  gptbox/.style={
    enhanced,
    sharp corners,
    boxrule=0.5pt,
    colback=gray!5,
    colframe=gray!40,
    left=6pt, right=6pt, top=4pt, bottom=4pt,
    fonttitle=\bfseries,
  }
}

\definecolor{chatbg}{RGB}{248,248,248}
\definecolor{userbg}{RGB}{230,244,255}
\definecolor{assistantbg}{RGB}{240,240,240}
\definecolor{userlabel}{RGB}{0,102,204}
\definecolor{assistantlabel}{RGB}{0,0,0}

\definecolor{codegreen}{rgb}{0,0.6,0}
\definecolor{codegray}{rgb}{0.5,0.5,0.5}
\definecolor{codepurple}{rgb}{0.58,0,0.82}
\definecolor{codebg}{rgb}{0.95,0.95,0.92}

\mdfdefinestyle{ChatBoxFrame}{%
    linecolor=black,
    outerlinewidth=0.1pt,
    roundcorner=20pt,
    innertopmargin=\baselineskip,
    innerbottommargin=\baselineskip,
    innerrightmargin=20pt,
    innerleftmargin=20pt,
    backgroundcolor=white}
\newenvironment{chatbox}
{%
      \begin{mdframed}[style=ChatBoxFrame]%
}
{%
      \end{mdframed}%
}

\newenvironment{userturn}
{%
\small
    \tcbset{colframe=userbg,colback=userbg}
    \begin{tcolorbox}[breakable,enhanced,flush right,width=0.8\textwidth]
}
{%
  \end{tcolorbox}%
}

\newenvironment{assistantturn}
{%
\small
    \tcbset{colframe=gray!10,colback=gray!10}
    \begin{tcolorbox}[breakable,enhanced,flush right,width=\textwidth]
    
}
{%
    \end{tcolorbox}
}

\tcbuselibrary{skins,breakable}

\definecolor{debuguserbgX}{RGB}{225,245,255}
\definecolor{debugassistantbgX}{RGB}{240,240,255}

\tcbset{
  debugchatstyleX/.style={
    breakable,
    enhanced,
    skin=enhancedmiddle,
    colback=white,
    colframe=black,
    arc=20pt,
    boxrule=0.4pt,
    left=20pt,right=20pt,top=10pt,bottom=10pt,
    oversize=0pt
  }
}

\newtcolorbox{debugchatboxX}[1][]{debugchatstyleX,#1}

\tcbset{
  debuguserstyleX/.style={
    breakable,
    enhanced,
    width=0.8\textwidth,
    halign=flush right,
    colback=debuguserbgX,
    colframe=debuguserbgX,
    arc=10pt,
    boxrule=0pt,
    left=6pt,right=6pt,top=6pt,bottom=6pt
  }
}
\newtcolorbox{debuguserturnX}[1][]{debuguserstyleX,#1}

\tcbset{
  debugassistantstyleX/.style={
    breakable,
    enhanced,
    width=0.8\textwidth,
    halign=left,
    colback=debugassistantbgX,
    colframe=debugassistantbgX,
    arc=10pt,
    boxrule=0pt,
    left=6pt,right=6pt,top=6pt,bottom=6pt
  }
}
\newtcolorbox{debugassistantturnX}[1][]{debugassistantstyleX,#1}

%% file: content/00-Introduction.tex
\chapter*{\vspace{-3cm}Introduction}
\addcontentsline{toc}{chapter}{Introduction}
\label{sec:Introduction}

Over the past few years, large language models have become increasingly useful for tasks such as writing, programming, and planning. More recently, they have started to be capable of contributing intellectually to scientific research. With GPT-5, we see early signs that, guided by an expert, the model can propose helpful ideas, perform deep literature searches, and even produce complete new proofs. This report documents internal and external cases in which GPT-5 contributed to scientific progress, and how it fell short. Our aim is to highlight what is possible today, what is still out of reach, and implications for the future of scientific research.

We focus on concrete examples spanning mathematics, physics, astronomy, computer science, biology, and materials science. Since the subject matter varies widely by section, each begins with an explanation of the background and motivation before discussing the contribution made by GPT-5. We also provide ChatGPT conversation transcripts when possible, and describe the human input that was needed, in order to make clear where the model currently adds value and where expert oversight remains essential. Several related accounts in a similar spirit have also surfaced in recent months, including \cite{feldman2025g,diez2025mathematical,ivanisvili2025counterexample,alexeev2025forbidden,jang2025point,salim2025acceleratingmathematicalresearchlanguage,georgiev2025exploration}. We note that the latter paper, on AlphaEvolve by Google, is slightly different in flavor, as it focuses on search problems with a well-defined objective function that can be hill-climbed. By contrast, we (and the other cited papers) are focused on a general-purpose system that can answer any type of query. The two approaches are complementary, each providing unique advantages to scientists.

We are also careful to point out the limits of current AI models. GPT-5 is imperfect: it can confidently make mistakes, ardently defend them, and confuse itself (and us) in the process. Results may depend on fine details of both the initial prompt and follow-up responses, and thus can be challenging to reproduce. Despite these limitations, we see real progress. GPT-5 can search broad conceptual spaces, integrate diverse information sources, and iterate quickly. It can tirelessly propose new ideas, help turn an imprecise idea into a concrete result, and sanity-check or extend a given line of thought. In some areas, such as literature search, it is often uniquely effective. 

Our aim is not to claim more than the evidence allows. It is to show, with specific examples, what GPT-5 can and cannot do today, and to give a clear path for how researchers can use it to accelerate scientific discovery while keeping standards high. We believe GPT-5 already provides substantial value for scientific researchers today, and will become an even more powerful tool tomorrow.
The rest of the paper is organized as follows:
\begin{itemize}
    \item Chapter \ref{sec:IndependentDiscovery} collects examples in which GPT-5 independently rediscovers known results at the research frontier in math, physics, and biology.
    \item Chapter \ref{sec:DeepLiteratureSearch} highlights the ability of GPT-5 to perform ``deep literature search''. Its ability to focus on the core concepts rather than the words used to describe them surmounts barriers of language between scientific disciplines to uncover seemingly forgotten or hard-to-find connections.
    \item Chapter \ref{sec:WorkingWithAI} displays examples of human researchers working in tandem with GPT-5 to accelerate their research workflow.
    \item Chapter \ref{sec:NewResults} presents examples of GPT-5 obtaining novel research-level results.
\end{itemize}

%% file: content/01-Bubeck.tex
\section{Improved step-size condition in a recent convex optimization result -- S{\'e}bastien Bubeck}
\label{sec:\secprefix-01}

This section can be viewed as a warm-up for the rest of the paper. The experiment was conducted on August 20th 2025, at a time when it was not yet clear whether GPT-5 could be used to push the frontier in research level questions. The goal of this experiment was more modest than what is described in several of the other sections, namely it was merely to see if GPT-5 could reproduce the main result of a recent paper, specifically the main theorem in \cite{barzilai2025convex}. This particular paper was chosen for several reasons:
\begin{itemize}
\item First, obviously, the paper is not in the training data for GPT-5 as it is too recent. But even more importantly, the main theorem is about a genuinely new question for a very classical field (convex optimization), and in particular no trace of this question can be found in the training data either.
\item The paper has three versions on arxiv (v1, v2, v3), and crucially v1 presents a suboptimal result which is then refined (and made optimal) in v2. The challenge I was interested in is: \emph{given v1, can GPT-5 rederive v2}?
\item In particular, if GPT-5 could rederive v2 from v1, it would indicate that it could have counterfactually accelerated the scientific discovery process.
As we will see, GPT-5 did not manage to fully rederive the v2 result, but it basically went half-way between v1 and v2. (See also the comment at the end of this section on deriving v2 from scratch with models that can reason for longer.)
\end{itemize}

Remarkably, this was my first attempt at an experiment of this type (i.e., I did not try any other paper than \cite{barzilai2025convex}).

\subsection{Description of the problem}

The most basic algorithm to optimize an objective function $f : \mathbb{R}^d \rightarrow \mathbb{R}$ is gradient descent with constant step-size $\eta$, where the iterates are given by 
$$x_{n+1} = x_n - \eta \nabla f(x).$$
A classical result in convex optimization is that, for smooth convex functions, convergence happens if and only if the step size is smaller than $2/L$ where $L$ is the smoothness constant (i.e., the Lipschitz constant of the gradient map $x \mapsto \nabla f(x)$). In \cite{barzilai2025convex} the authors study a new variant of the convergence question, which as far as we know (both the authors of that paper and the present author) was never studied before: instead of merely asking for convergence, under what condition is the \emph{curve traced by the visited objective function values itself convex}, i.e., when is the piecewise linear function going through the points $\{(n, f(x_n))\}_{n \in \mathbb{N}}$ a convex function itself?
\newline

In v1 of the paper, the authors prove that $\eta \leq 1/L$ is a sufficient condition, and they also show that $\eta \leq 1.75 / L$ is a necessary condition. In v2 of the paper, they close this gap and show that in fact $\eta \leq 1.75 / L$ is indeed sufficient.

\subsection{GPT-5 acting as a collaborator: from v1 getting close to v2}

\begin{figure}[h!]
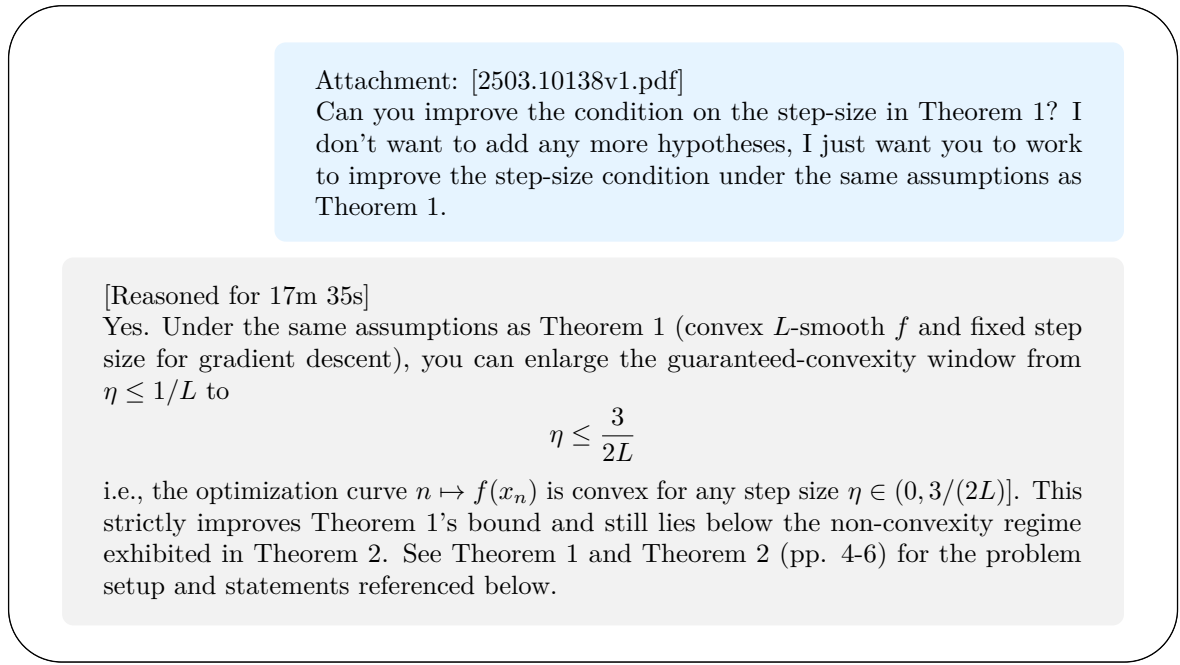

\begin{chatbox}
\begin{userturn}
Attachment: [2503.10138v1.pdf]
\newline
Can you improve the condition on the step-size in Theorem 1? I don't want to add any more hypotheses, I just want you to work to improve the step-size condition under the same assumptions as Theorem 1.
\end{userturn}

\begin{assistantturn}
[Reasoned for 17m 35s]
\newline
Yes. Under the same assumptions as Theorem 1 (convex $L$-smooth $f$ and fixed step size for gradient descent), you can enlarge the guaranteed-convexity window from $\eta \leq 1/L$ to
$$\eta \leq \frac{3}{2L}$$
i.e., the optimization curve $n \mapsto f(x_n)$ is convex for any step size $\eta \in (0, 3/ (2L)]$. This strictly improves Theorem 1's bound and still lies below the non-convexity regime exhibited in Theorem 2. See Theorem 1 and Theorem 2 (pp. 4-6) for the problem setup and statements referenced below.
\end{assistantturn}
\end{chatbox}
\caption{The interaction with GPT-5 Pro to derive the improved bound.}
   \label{fig:convex_opt_1}
\end{figure}

As discussed above, the experiment I was interested in is \emph{given v1, can GPT-5 rederive v2}? To check this, I uploaded the PDF of the v1 paper, and simply asked GPT-5 (Pro) to try to improve the result. The entire interaction is captured in Figure \ref{fig:convex_opt_1} above. In particular, one can see that GPT-5 claims to have improved the condition from $\eta \leq 1/L$ to $\eta \leq 1.5/L$, thus approaching the optimal bound (but not quite getting there) of $\eta \leq 1.75 / L$. But is this claim substantiated? It is indeed, and the proof given by GPT-5 is shown in Figure \ref{fig:convex_opt_2}, which the present author has verified to be correct. Moreover, I inspected the reasoning traces to make sure that no web search was performed (as this could have revealed the existence of v2), and furthermore the proof given by GPT-5 is quite different from the one in v2. Indeed, the GPT-5 proof can be viewed as a more canonical variant of the v1 proof, whereas the (human) v2 proof requires a clever weighting of different inequalities for certain cancellations to happen.
\newline

To say it plainly, such a result (improving from $1/L$ to $1.5/L$) could probably have been achieved by \emph{some} experts in the field in a matter of hours, and likely for \emph{most} experts it would have taken a few days. This is the type of science acceleration that we will see time and again in this report.
\newline

We end this section by noting that our internal models, which can think for a few hours, were able to derive the optimal bound $1.75/L$ from scratch (i.e., without providing v1 of the paper in context, but simply asking the main question studied here directly).

\begin{figure}
    \centering
\begin{chatbox}
\begin{assistantturn}
\begin{theorem}[Improved step-size condition]\label{thm:improved-stepsize}
Let $f:\mathbb{R}^d\to\mathbb{R}$ be convex and $L$-smooth, and consider gradient descent
\[
x_{k+1}=x_k-\eta \nabla f(x_k), \qquad k\ge 0.
\]
If $\eta\in(0,\tfrac{3}{2L}]$, then the sequence $\{f(x_k)\}_{k\ge 0}$ is convex, in the sense that the decreases
\[
D_k := f(x_k)-f(x_{k+1})
\]
form a nonincreasing sequence: $D_{k+1}\le D_k$ for all $k\ge 0$.
\end{theorem}

\begin{proof}
Write $g_k:=\nabla f(x_k)$ and $\Delta_k:=g_{k+1}-g_k$. We compare two consecutive decreases $D_k$ and $D_{k+1}$.

\emph{(1) A lower bound for $D_k$.}
For convex $L$-smooth $f$, the Bregman divergence obeys
\[
\frac{1}{2L}\,\|\nabla f(x)-\nabla f(y)\|^2 \;\le\; f(x)-f(y)-\langle \nabla f(y),x-y\rangle .
\]
With $x=x_k$, $y=x_{k+1}$ and $x_k-x_{k+1}=\eta g_k$ this gives
\begin{equation}\tag{A}\label{eq:A}
D_k \;\ge\; \eta\,\langle g_{k+1},g_k\rangle \;+\; \frac{1}{2L}\,\|\Delta_k\|^2 .
\end{equation}

\emph{(2) An upper bound for $D_{k+1}$.}
By convexity, $f(x)-f(y)\le \langle \nabla f(x),x-y\rangle$. With $x=x_{k+1}$, $y=x_{k+2}$ and $x_{k+1}-x_{k+2}=\eta g_{k+1}$,
\begin{equation}\tag{B}\label{eq:B}
D_{k+1} \;\le\; \eta\,\|g_{k+1}\|^2 .
\end{equation}

\emph{(3) Subtract and use cocoercivity.}
Subtracting \eqref{eq:B} from \eqref{eq:A} yields
\[
D_k-D_{k+1}
\;\ge\;
\eta\,\langle g_{k+1},g_k-g_{k+1}\rangle+\frac{1}{2L}\|\Delta_k\|^2
=
-\eta\,\langle g_{k+1},\Delta_k\rangle+\frac{1}{2L}\|\Delta_k\|^2 .
\]
Since $g_{k+1}=g_k+\Delta_k$, we have $\langle g_{k+1},\Delta_k\rangle=\langle g_k,\Delta_k\rangle+\|\Delta_k\|^2$, hence
\[
D_k-D_{k+1}
\;\ge\;
-\eta\,\langle g_k,\Delta_k\rangle+\Bigl(\frac{1}{2L}-\eta\Bigr)\|\Delta_k\|^2 .
\]
For convex $L$-smooth $f$, gradients are $1/L$-cocoercive:
\[
\frac{1}{L}\,\|\Delta_k\|^2 \;\le\; \langle \Delta_k,\,x_{k+1}-x_k\rangle
= \langle \Delta_k,\,-\eta g_k\rangle
= -\eta\,\langle g_k,\Delta_k\rangle .
\]
Therefore,
\[
D_k-D_{k+1}
\;\ge\;
\Bigl(\frac{1}{L}+\frac{1}{2L}-\eta\Bigr)\|\Delta_k\|^2
=
\Bigl(\frac{3}{2L}-\eta\Bigr)\|\Delta_k\|^2 .
\]
If $\eta\le \tfrac{3}{2L}$, the right-hand side is nonnegative, so $D_{k+1}\le D_k$ for all $k$, which is equivalent to the convexity of the sequence $\{f(x_k)\}_{k\ge 0}$. 
\end{proof}
\end{assistantturn}
\end{chatbox}
   \caption{GPT-5 Pro's proof of the sufficient condition $\eta \leq 1.5/L$.}
    \label{fig:convex_opt_2}
\end{figure}

%% file: content/02-Lupsasca.tex
\section{Discovering new black hole symmetries with GPT-5 -- Alex Lupsasca}
\label{sec:\secprefix-02}

We document, in a standardized format, a theoretical calculation in black hole physics performed by an AI.
GPT-5~Pro (re)derived nontrivial Lie point symmetries---including an $\mathsf{SL}(2,\mathbb{R})$ algebra---of the stationary, axisymmetric wave equation on a Kerr background.
The model initially failed on the curved-space problem, but then succeeded after a flat-space warm-up, ultimately producing the correct symmetry generators that underpin the recent results in \cite{Lupsasca2025} (to which the model did not have access).

\subsection{The problem in context}

Astrophysical black holes are characterized by their mass $M$ and angular momentum $J=aM$.
These two quantities completely determine the black hole spacetime geometry: it is described by the Kerr metric with parameters $(M,a)$.

We study massless, stationary, axisymmetric waves on a rotating (Kerr) black hole.
In Boyer-Lindquist coordinates $(t,r,x,\phi)$ with $x=\cos{\theta}$, the governing equation is the scalar Laplace operator restricted to these symmetries,
\begin{align}
    \label{eq:StataxiLaplacian}
    \pd_r\br{\Delta(r)\pd_r\psi(r,x)}+\pd_x\br{(1-x^2)\pd_x\psi(r,x)}=0,
    \qquad\Delta(r)=r^2-2Mr+a^2.
\end{align}
This is a linear second-order partial differential equation in two variables.
Its solutions encode the static tidal response of black holes; their asymptotics determine horizon deformability (``Love numbers'').
In general relativity, black holes exhibit vanishing (static) Love numbers, a surprising rigidity often traced to hidden symmetries of the field equations.

A key question is thus: what are the symmetries of Eq.~\eqref{eq:StataxiLaplacian}?

\paragraph{Human route.} 
Classically, one would apply Lie's algorithm for reduction of order to Eq.~\eqref{eq:StataxiLaplacian}.
In this procedure, one first posits a symmetry generator $X=\xi^r\pd_r+\xi^x\pd_x+\eta\pd_\psi$, prolongs $X$ to derivatives, then imposes invariance of Eq.~\eqref{eq:StataxiLaplacian}, and finally solves the resulting overdetermined system for the unknown functionss $(\xi^r,\xi^x,\eta)$.
This is arduous in a curved black hole background (coefficients depend on $r$ and $x$ via $\Delta$ and $1-x^2$).
In recent work \cite{Lupsasca2025}, I carried out this analysis and was surprised to find three nontrivial Lie point symmetries that generate an $\mathsf{SL}(2,\mathbb{R})$ algebra.
These symmetries include a dilatation $H_0$ associated with a scale invariance that can be used to explain the absence of tidal response (vanishing Love numbers) for black holes in general relativity.

\paragraph{Additional references.}
For background on Kerr and wave equations: \cite{Kerr1963,Teukolsky1973,Wald1984}.
For black-hole tidal response and Love numbers, our main technical reference is \cite{Lupsasca2025} and citations therein.

\subsection{The task given to GPT-5 Pro: goal, prompts, and model behavior}

\paragraph{Goal.}
Elicit from GPT-5~Pro the Lie point symmetries of \eqref{eq:StataxiLaplacian}, ideally reproducing the $\mathsf{SL}(2,\mathbb{R})$ generators and their curved-space coefficients recently obtained in \cite{Lupsasca2025}.

\paragraph{Outcome.} After $\sim$5 minutes of internal reasoning, the model (incorrectly) reported that the equation had no continuous symmetries beyond trivial scalings.

\clearpage

\begin{chatbox}
\begin{userturn}
What are the Lie point symmetries of the PDE \eqref{eq:StataxiLaplacian}?
\end{userturn}
\begin{assistantturn}
[Reasoned for 5m 13s]
    
[Incorrect answer: GPT-5 Pro finds no symmetries and wrongly asserts that none exist.]
\end{assistantturn}
\end{chatbox}

\paragraph{Warm-up prompt and outcome (success on flat space).}
To scaffold the task, we initialized a new instance of the model and first posed the flat-space limit $M,a\to0$ in cylindrical coordinates $(\rho,z)$ with $\rho=r\sqrt{1-x^2}$, $z=rx$:
\begin{align}
\label{eq:FlatLaplacian}
\Bigl(\pd_\rho^2+\pd_z^2+\frac{1}{\rho}\pd_\rho\Bigr)\psi(\rho,z)=0.
\end{align}

\paragraph{Outcome.} After 10 minutes and 27 seconds, the model produced all the symmetries, including three that generate $\mathsf{SL}(2,\mathbb{R})$:
\begin{align}
\label{eq:FlatSymmetries}
H_+=\pd_z,\qquad
H_0=\rho\,\pd_\rho+z\,\pd_z+\frac{1}{2},\qquad
H_-=
2\rho z\,\pd_\rho-(\rho^2-z^2)\,\pd_z+z.
\end{align}
Here $H_-$ is the nontrivial special conformal generator; obtaining it suggests that the model executed (or emulated) a real symmetry computation rather than guessing.

\begin{chatbox}
    \begin{userturn}
        What are the Lie point symmetries of \eqref{eq:FlatLaplacian}?
    \end{userturn}
    \begin{assistantturn}
        [Reasoned for 10m 27s]

        [Correct answer including generators \eqref{eq:FlatSymmetries}.]
    \end{assistantturn}
    \begin{userturn}
        What are the Lie point symmetries of \eqref{eq:StataxiLaplacian}?
    \end{userturn}
    \begin{assistantturn}
        [Reasoned for 18m 9s]

        [Correct answer including generators \eqref{eq:Generators}.]
    \end{assistantturn}
\end{chatbox}

\paragraph{Second curved-space attempt and outcome (success).}
We then gave the same instance of GPT-5 the same prompt as before: ``What are the Lie point symmetries of \eqref{eq:StataxiLaplacian}?''

\paragraph{Outcome.} Within $\sim$18 minutes, the model produced the correct curved-space generators closing into $\mathsf{SL}(2,\mathbb{R})$:
\begin{subequations}
\label{eq:Generators}
\begin{align}
H_+&=\frac{x\,\Delta\,\pd_r+(r-M)(1-x^2)\,\pd_x}{(r-M)^2-(M^2-a^2)x^2},\\
H_0&=\frac{(r-M)\,\Delta\,\pd_r+(M^2-a^2)x(1-x^2)\,\pd_x}{(r-M)^2-(M^2-a^2)x^2}+\frac{1}{2},\\
H_-&=\frac{(M^2-a^2)x\,\Delta\,\pd_r-(r-M)(1-x^2)\!\left[\Delta-(M^2-a^2)x^2\right]\pd_x}{(r-M)^2-(M^2-a^2)x^2}+x\,\Delta\,\pd_r+(r-M)\,x.
\end{align}
\end{subequations}

\paragraph{Analysis.}
In summary, here is
\begin{itemize}
    \item \textbf{What GPT-5 got right:} The relatively simple flat-space symmetries; the full nontrivial curved-space coefficients; the $\mathsf{SL}(2,\mathbb{R})$ structure.
    \item \textbf{What GPT-5 got wrong (along the way).} The cold start on Eq.~\eqref{eq:StataxiLaplacian} incorrectly concluded ``no symmetries.'' The model appears to have needed to ``warm up'' via the simpler flat-space problem \eqref{eq:FlatLaplacian} sharing the same symmetry structure.
\end{itemize}

\paragraph{Log.}
For verification and reproducibility purposes, this conversation with GPT-5 can be accessed here: \cite{LupsascaGPT}.

\subsection{Result, implications, and next steps}

\paragraph{Result.}
GPT-5~Pro (re)discovered the curved-space $\mathsf{SL}(2,\mathbb{R})$ symmetry generators \eqref{eq:Generators} of Eq.~\eqref{eq:StataxiLaplacian}.
This matches the key structural insight of \cite{Lupsasca2025}.
Practically, once the symmetry is known, downstream results (e.g., constraints on tidal response and the vanishing of static Love numbers in this sector) follow with comparatively modest analysis.

\paragraph{Reflection on the interaction.}
Two observations:
\begin{enumerate}
\item \textbf{Scaffolding mattered.} The model failed “cold” but succeeded rapidly after a closely related warm-up. This suggests retrieval or internal pattern activation can be primed by presenting a simpler member of the same symmetry class.
\item \textbf{Algorithmic plausibility.} The final generators are too structured to be a lucky guess. The model likely executed (implicitly) a mix of: recognizing conformal invariance in the flat equation, hypothesizing a curved analogue, and/or exploiting a coordinate map that simplifies Eq.~\eqref{eq:StataxiLaplacian} toward Eq.~\eqref{eq:FlatLaplacian}.
\end{enumerate}

\paragraph{Implications}
\begin{itemize}
\item \textbf{AI as a symmetry engine.} With minimal domain scaffolding, current models can carry out nontrivial Lie-symmetry discovery for PDEs with non-constant coefficients.
\item \textbf{Research velocity.} Given such capabilities, the time from idea to publishable result can compress from months to days once the right prompts and scaffolds are in place.
\item \textbf{Generalization opportunity.} The same workflow (warm-up on simplified problems, then lift) can be applied to more complex problems of physical interest in black hole theory and beyond.
\end{itemize}

\paragraph{Takeaway.} GPT-5~Pro, when properly scaffolded, uncovered the $\mathsf{SL}(2,\mathbb{R})$ symmetry content of a curved-space PDE central to black-hole tidal response. This supports a broader thesis: contemporary LLMs can act as practical assistants for symmetry discovery and analytic structure mining in theoretical physics.

%% file: content/03-Unutmaz.tex
\section {Mechanistic analysis and outcome prediction for in vitro immune system experiments using GPT-5 Pro -- Derya Unutmaz, M.D.}
\label{sec:\secprefix-03}

Here, I demonstrate that GPT-5 Pro successfully analyzed a figure from an experiment with human T cells cultured with 2-deoxy-D-glucose (2-DG) that showed an increased proinflammatory Th17 cell subset. This experiment was performed in our lab several years ago, but the mechanism remained unclear. GPT-5 Pro provided the key mechanism that could explain these findings and, in addition, made highly relevant experimental suggestions. The mechanistic insight and further hypothesis to dissect these findings were highly valuable and not immediately obvious, despite our deep expertise in this field. In a subsequent unpublished figure, GPT-5 Pro interpreted flow cytometry data of the checkpoint inhibitors PD-1 and LAG-3 on cytotoxic T cells after transient glycolysis inhibition, inferring that 2-DG was reprogramming inhibitory receptor expression through combined effects on glycosylation and through attenuated T cell receptor signaling. It further correctly predicted that a brief 2-DG pulse during CAR-T cell generation from these cells would enhance their cytotoxicity towards target cancer cell lines, which we had internally validated in unpublished results. Together, these examples illustrate how GPT-5 Pro can function as a true mechanistic co-investigator in biomedical research, compressing months of reasoning into minutes, uncovering non-obvious hypotheses, and directly shaping experimentally testable strategies.

\subsection{Problem Context}

We had been studying how subsets of human immune cells called T cells respond to modifications in their glucose metabolism. There was significant evidence that glucose uptake and energy metabolism influence the differentiation of T cells into effector subsets, which can enhance protection against infections or drive pathologies such as chronic inflammation and autoimmunity, and could have implications during cancer immunotherapy. To dissect this mechanism, we used a competitive analog of glucose, 2-deoxy-D-glucose (2-DG), as an inhibitor of glucose uptake in T cells. Because T cells are highly heterogeneous, we first sorted them into {CD4\weirdplus} and {{CD8\weirdplus}} subsets, which are helper and cytotoxic T cells, respectively. In the initial experiment, we used {CD4\weirdplus} T cells and activated them in the presence of varying doses of 2-DG or under low-glucose conditions. After two days, these treatments were washed out to allow the T cells to proliferate in the presence of interleukin-2 (IL-2), a growth-factor cytokine. The idea was that any impact of glucose inhibition would occur early in the experiment, become imprinted during priming, and, after two weeks of expansion, manifest as differences in T-cell differentiation and function. 

\subsection{The task given to GPT-5: goal, prompts, and model behavior.}

The figure in the prompt below is derived from flow cytometry data on cells cultured with 2-DG. I uploaded it as part of the prompt, which led to the following conversation with GPT-5. While we had published a similar findings as a preprint, this specific figure or experiment had not been previously published. 

\clearpage

\begin{chatbox}
  \begin{userturn}
    \includegraphics[width=0.8\textwidth]{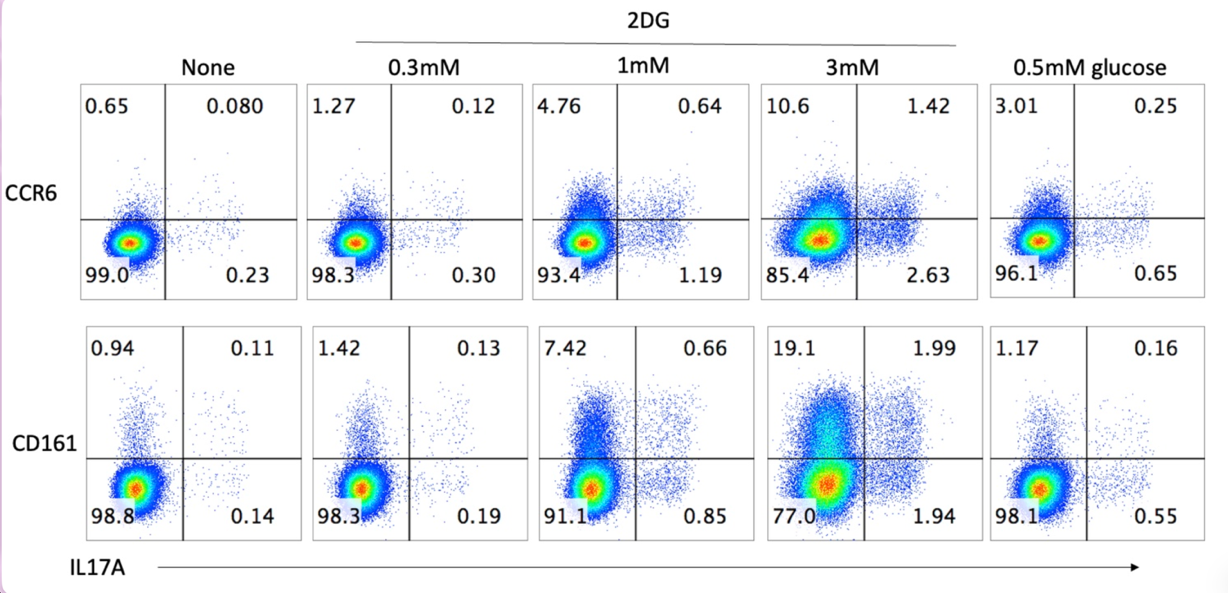}
    
    Context and question: We tested how transient glycolysis inhibition during human T cell priming affects later Th17 function. Sorted {CD4\weirdplus} T cells were stimulated with anti-CD3 and anti-CD28 in the presence of specified concentrations of 2-deoxy-D-glucose (2-DG). Cells were then expanded for two weeks without 2-DG, re-stimulated, and assayed for IL-17A together with CCR6 or CD161. Analyze the provided flow cytometry dataset obtained after stimulating these two-week cultured cells with PMA plus ionomycin for 6 hours, followed by intracellular IL-17A staining and surface staining for CCR6 and CD161. Explain the results, derive mechanisms, extract high-level conclusions, assess impact, and recommend follow-up experiments. Also state how interpretations would differ if the same design started from sorted naive {CD4\weirdplus} T cells instead of memory cells.
    \begin{itemize}
        \item A brief executive summary of the main findings.
        \item A quantitative readout of dose responses for IL-17A, CCR6, and CD161, including percent positive, median fluorescence intensity, and effect sizes with confidence intervals across donors.
        \item Clear plots or tables comparing 2-DG doses to control, both at priming and after the two-week washout, indicating whether effects persist after removal of 2-DG.
        \item An interpretation of how 2-DG altered Th17 biology, considering all potential mechanisms.
        \item High-level conclusions and potential impact for immunotherapy, autoimmunity, and other immune responses.
        \item A concise plan for follow-up experiments that will help decode the potential mechanisms and identify the precise mechanism.
        \item A comparison of expected outcomes if the experiment started with naive {CD4\weirdplus} T cells versus memory {CD4\weirdplus} T cells, rather than total {CD4\weirdplus} T cells that contain a mixture of naive and memory subsets, including suggested follow-up experiments to resolve mechanism.
        \item Key follow-up experiments to establish mechanism and causality, with potential outcomes.
    \end{itemize}
  \end{userturn}
\end{chatbox}

\clearpage

\begin{chatbox}
  \begin{assistantturn}
    [Reasoned for 17m]
    \newline \newline 
    Briefly exposing human {CD4\weirdplus} T cells to 2DG during the first activation only leaves a durable, dose-dependent imprint detectable two weeks later: the frequencies of {IL-17A\weirdplus}, {CCR6\weirdplus}, and {CD161\weirdplus} cells all rise, with the largest increases at 3mM 2DG. Marker increases (CCR6, CD161) are substantially larger than the increase in IL-17A itself, implying an expansion or imprinting of Th17-lineage memory {CCR6\weirdplus}/{CD161\weirdplus}) that is poised for, but not uniformly executing, IL-17A production. A lowglucose control (0.5 mM) partly mimics the effect but is weaker than 2DG at equal nominal concentrations, suggesting contributions from glycolysis restriction and N-linked glycosylation interference (a known 2DG off-target) rather than energy limitation alone. The effect persists after drug washout, indicating cellular reprogramming, selective survival/expansion of Th17-lineage cells, or both.
  \end{assistantturn}

\end{chatbox}

After 17 minutes of reasoning, GPT-5 Pro produced multiple histogram figures like these two examples, that perfectly analyzed the data, along with an outstanding executive summary that captured the key mechanistic points.

\begin{table}[ht]
\centering
\begin{tabular}{c c}
\includegraphics[width=0.48\linewidth]{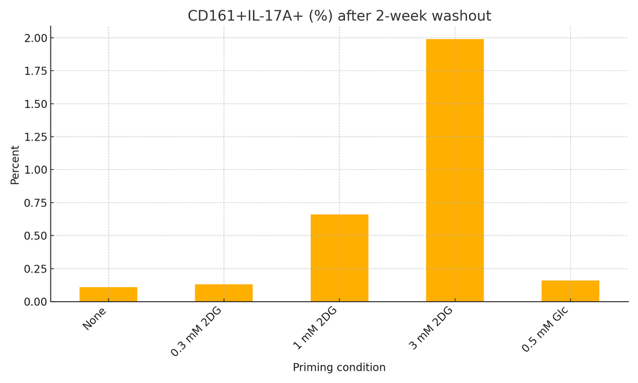} &
\includegraphics[width=0.48\linewidth]{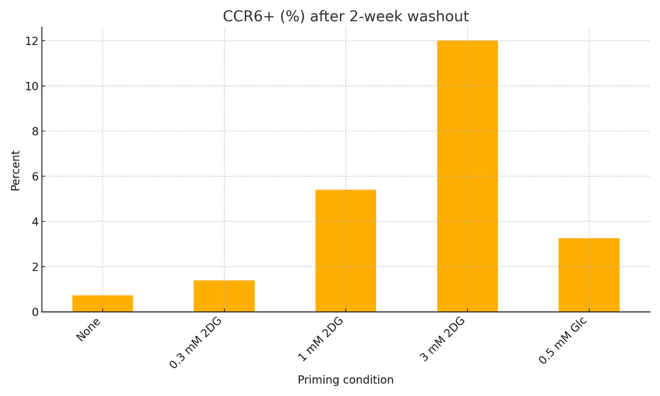} \\
\end{tabular}
\end{table}

It also generated a quantitative dose–response readout for the figure (output not shown) and a CSV file containing all underlying data used to generate the histograms. GPT-5 Pro then provided an interpretation of how 2-DG treatment altered Th17 biology. The mechanistic inferences from GPT-5 Pro were pivotal, and some were ideas we had not previously considered. In retrospect, in particular, the proposed mechanism of reduced IL-2 signaling via interference with N-linked glycosylation made clear biological sense, because it could directly explain the disinhibition of Th17 cell differentiation under 2-DG treatment. However, this mechanistic hypothesis had not occurred to us previously, as making this connection was not immediately obvious. Other mechanistic hypotheses were also highly relevant and plausible. 

\clearpage

\begin{chatbox}
  \begin{assistantturn}
  \small
      \textbf{What the data say:} \newline \newline
      A single 2DG exposure at priming led, two weeks later, to (i) a 13× increase in {IL-17A\weirdplus} cells, (ii) a 16–20× increase in Th17-lineage markers (CCR6, CD161), and (iii) disproportionately more {CCR6\weirdplus}/{CD161\weirdplus} cells than {IL-17A\weirdplus} cells. This pattern fits an \textbf{enrichment and/or imprinting of Th17 lineage memory} rather than a uniform gain in acute effector cytokine output across all cells. Many {CCR6\weirdplus}/{CD161\weirdplus} cells are {IL-17A\weirdminus} at re-stimulation, a hallmark of \textbf{poised Th17 memory} that can produce IL-17A under the right cues (for example, IL-1$\beta$ + IL-23).
      \newline \newline 
      \textbf{Mechanistic inferences.} Several nonexclusive mechanisms could generate this durable skew:
      \vspace{0.5em}
    \begin{enumerate}[leftmargin=1.5em]
        \item \textbf{Reduced IL2 signaling via N-linked glycosylation interference.} 2-DG at mM levels perturbs Nglycan synthesis, which in activated T cells reduces proper surface expression of glycoproteins---including \textbf{IL2R$\alpha$ (CD25)}. Diminished IL2/STAT5 signaling \textbf{disinhibits Th17 differentiation} and limits conversion of Th17 cells to Th1-like states. The stronger effect of 2DG than low glucose supports a role for \textbf{glycosylation} beyond mere ATP/pyruvate restriction.
        \item \textbf{Acute metabolic checkpointing that locks in a lineage program.} Transient glycolytic inhibition can shift activation metabolism toward \textbf{mitochondrial oxidation}, altering mTORC1/HIF1$\alpha$ activity and chromatin modifiers (acetylCoA, $\alpha$-ketoglutarate), leaving an \textbf{epigenetic memory} at Th17-relevant loci (RORC, IL23R, CCR6). The persistence after washout is most consistent with such a program.
        \item \textbf{Selective survival/expansion.} 2DG may preferentially limit proliferation or survival of non-Th17 subsets (e.g., Th1), effectively enriching preexisting \textbf{{CCR6\weirdplus}/{CD161\weirdplus} memory} present in the sorted total {CD4\weirdplus} pool. The larger fold increase in CD161/CCR6 than in IL-17A favors this explanation as at least a contributor.
        \item \textbf{Altered TCR/CD28 signal strength.} Perturbed glycosylation also modulates TCR and co-stimulatory receptor function during priming. Weaker IL-2-rich effector programs and preserved ROR$\gamma$t/AHR programs would tilt toward the Th17 lineage.
    \end{enumerate}

     \vspace{1em}
    \textbf{High level conclusions and potential impact:} A onetime metabolic perturbation during priming imprints a durable Th17-leaning memory phenotype in human {CD4\weirdplus} T cells. Because Th17/Tc17 states can display superior persistence and tissue trafficking, this imprint could be useful for adoptive T cell therapies (including CAR-T) when a Th17-like program is desired. Conversely, it warns that transient glycolysis inhibition---whether intentional (e.g., 2-DG as an anti-metabolite) or incidental (nutrient limitation in bioreactors)---might increase Th17-lineage memory and thereby exacerbate Th17-mediated autoimmunity if applied in vivo. The clear persistence after washout argues we are not seeing acute metabolic suppression but rather a biologically meaningful reprogramming.
  \end{assistantturn}
\end{chatbox}

GPT-5 Pro then proposed key follow-up experiments. We had already performed the first of these experiments,  the mannose rescue experiment, weeks after the first experiment. The result matched GPT-5 Pro’s prediction exactly. However, although we have not published this specific figure, we did publish a preprint in BioRxiv that included the mannose rescue experiment in a similar story. Therefore, while impressive, a caveat for this suggestion is that GPT-5 Pro may have known about this finding and made the connection with this result. 

That said, GPT-5 Pro also suggested additional, even more important experiments targeting N-linked glycosylation using inhibitors, as well as experiments to interrogate the IL-2 axis during priming, neither of which we have performed nor published in the context of this specific finding. We are also not aware of any other publication performing these experiments for Th17 cell differentiation status after 2-DG treatment. In addition, the suggestions by GPT-5 Pro to map the functional plasticity of Th17 cells and their epigenetic imprinting are striking, because they require an extremely deep understanding of Th17 cell biology.

\begin{chatbox}
  \begin{assistantturn}
To discriminate among mechanisms and establish causality, the most informative next steps are:
\textbf{Partition “glycolysis block” vs “glycosylation block.”} Repeat the priming step with: 
\vspace{0.5em}
\begin{enumerate}
    \item 2-DG$\pm$\textbf{mannose rescue} (restores Nglycosylation without restoring glycolysis)
    \item \textbf{specific glycolysis inhibitors} that do \textbf{\textit{not} inhibit Nglycosylation (e.g., PFKFB3 inhibitor 3PO or iodoacetate at carefully titrated, nontoxic doses), and}
    \item \textbf{Nglycosylation inhibitors} (e.g., tunicamycin) at low doses. If mannose reverses the 2-DG effect while glycolysis-only inhibitors do not phenocopy it, glycosylation is the driver.
\end{enumerate}
\vspace{0.5em}

\textbf{Interrogate the IL2 axis during priming.} Quantify surface \textbf{CD25} maturation (PNGase F shift), pSTAT5, and IL2 consumption. Rescue experiments with \textbf{exogenous IL}\textbf{2}, or \textbf{IL}\textbf{2/anti}\textbf{IL}\textbf{2} immune complexes that enforce STAT5 signaling, should counteract Th17 bias if IL2 pathway suppression is causal.

\vspace{0.5em}

\textbf{Track proliferation/survival and composition.} Use CellTrace/Ki67 and Annexin V at 24–96 h after priming to determine whether 2-DG is \textbf{enriching} {CCR6\weirdplus}/{CD161\weirdplus} cells through selective growth/survival. Include absolute cell counts so percentages can be interpreted.

\vspace{0.5em}

\textbf{Define the epigenetic imprint.} Perform \textbf{ATAC}\textbf{seq} and targeted \textbf{CUT\&Tag} for H3K27ac/H3K4me3 at RORC, IL-17A/F, IL23R, CCR6, and PRDM1 after priming $\pm$2-DG and again at day 14. Lasting chromatin accessibility gains would confirm a programmed memory.

\vspace{0.5em}

\textbf{Map functional plasticity.} After the two-week washout, restimulate with \textbf{IL}-\textbf{1$\beta$ + IL}-\textbf{23} (Th17 stabilizing) or \textbf{IL}\textbf{12} (Th1 skewing) to see whether 2DG-primed cells are preferentially \textbf{locked} into or \textbf{resistant} to Th1 conversion.

\vspace{0.5em}

\textbf{Rule out MAIT contamination.} Because CD161 is also expressed by MAIT cells, add TRAV12 (V$\alpha$7.2) and MR1tetramer staining. This will confirm the cells are bona fide Th17-lineage {CD4\weirdplus} T cells.

\vspace{0.5em}

\textbf{Metabolism readouts.} Seahorse ECAR/OCR at 24–48h post-priming to verify the intended glycolytic restriction and compensatory mitochondrial changes; measure mTORC1 readouts (pS6) and HIF1$\alpha$.''
\end{assistantturn}
\end{chatbox}

In the follow-up prompt, I also uploaded the results of the mannose rescue experiment we had published. GPT-5 Pro confirmed previous suggestions and further refined the mechanistic model to explain these results together with the similar follow-up experiments suggested (output not shown).

\begin{chatbox}
  \begin{assistantturn}
\textbf{Integrating both datasets and the naïve vs memory result}

Your original titration showed a monotonic 2-DG dose–response and persistent increases in CCR6/CD161 well after washout; your new experiment shows that mannose fully normalizes or nearly normalizes these increases. Together with your subset sort, three strong inferences follow.

\vspace{0.5em}

\begin{enumerate}[leftmargin=1.5em]
\item The 2-DG imprint is driven primarily by impaired N-linked glycosylation during priming. Mannose bypasses the 2-DG block in the dolichol-linked oligosaccharide pathway and rescues normal glycoprotein maturation. The near-complete rescue at 3mM mannose means glycolysis inhibition is not the principal cause of the Th17 skew.
\item The responding cells are memory, not naïve, {CD4\weirdplus} T cells. The absence of effect in naïve cells rules out a simple “2DG instructs naïve Th17 differentiation” model under your conditions. Memory cells likely contain preexisting {CCR6\weirdplus}/{CD161\weirdplus} Th17-lineage clones; 2-DG preserves or expands them during priming by attenuating IL-2/STAT5 and/or by dampening TCR/CD28 glycoprotein function, both of which otherwise push cells away from Th17 toward Th1/CTL-like fates.
\item Programming plus selection best explains the magnitude and durability. Short term glycosylation stress at priming can both select for Th17-lineage memory cells (differential proliferation/survival) and program chromatin at Th17 loci. The persistence after two weeks without drug argues that at least part of the effect is epigenetically stabilized.
\end{enumerate}

\vspace{0.5em}

\textbf{Mechanistic model that fits all observations}\textbf{. }During the first 24–48h of activation, 3mM 2DG impairs N-glycosylation of multiple surface receptors (notably IL2R$\alpha$/CD25, TCR, CD28, potentially IL-6R and IL-23R). Hypoglycosylated receptors are retained or function less efficiently at the membrane, which reduces IL-2/STAT5 signaling and blunts strong co-stimulation. Because IL-2/STAT5 antagonizes Th17 and promotes Th1/Treg fates, this temporary reduction biases activated memory cells toward retaining or consolidating a Th17 program. Mannose restores Nglycan synthesis, normalizing receptor glycosylation and thereby reversing the imprint. Naïve cells, which depend heavily on IL-2 for survival and lack precommitted Th17 chromatin, do not show the same skew; any reduction in IL-2 signaling at priming in naïve cells tends to reduce expansion rather than drive Th17 features.

\vspace{0.5em}

This model predicts that, in 2-DG-primed memory cultures: surface CD25 will show lower expression and altered glycoform size; pSTAT5 will be reduced; ATACseq will show enhanced accessibility at RORC, IL-23R, CCR6; and proliferation/survival curves will show relative sparing or enrichment of {CCR6\weirdplus}/{CD161\weirdplus} cells.
\end{assistantturn}
    
\end{chatbox}

This final section on impact, implications, and high-level interpretation offered particularly remarkable insight into these hypothesis and proposed experiments.

\clearpage

\begin{chatbox}
  \begin{assistantturn}
  
Because naïve cells did not respond, the 2DG effect does not instruct de novo Th17 programming from naïve precursors under these conditions. Instead, it selects and stabilizes Th17-lineage features in memory cells already carrying partial RORC/IL-23R/CCR6 programs. This distinction has practical consequences. If your aim is to manufacture Th17-leaning T cell products for therapy, enriching memory {CD4\weirdplus} (and potentially {CD8\weirdplus}) compartments before a short 2-DG priming pulse will be far more effective than applying 2-DG to naïve enriched starting material. Conversely, if you are attempting to avoid Th17 skew in a culture process, maintaining normal glycosylation (adequate mannose/glucose; avoiding 2-DG exposure during early activation) is essential.

\vspace{0.5em}

\textbf{Impact and applications.} For immunotherapy, a brief, controlled glycosylation perturbation at priming could be used to generate Th17/Tc17-leaning memory with improved persistence and trafficking, then returned to normal glycosylation by washout or mannose supplementation. For autoimmunity, the data warn that 2-DG --or any intervention that transiently depresses N-glycosylation during T cell activation---could exacerbate Th17 immunity, especially within existing memory pools. For vaccine design, manipulating glycosylation during priming may tune helper subset balance without permanent genetic changes.''

\end{assistantturn}
\end{chatbox}

I then uploaded an unpublished flow cytometry figure showing the effect of 2-DG on the checkpoint inhibitors PD-1 and LAG-3 on activated {CD8\weirdplus} naive and memory T cells (Figure I.3 shown below) with similar prompt to the original. GPT-5 Pro perfectly analyzed and described the data in the figure:

\begin{chatbox}
  \begin{assistantturn}
A brief, transient exposure to 3 mM 2‑DG during anti‑CD3/CD28 activation profoundly remodels early checkpoint receptor expression on human {CD8\weirdplus} T cells measured on day 4. In naïve {CD8\weirdplus} cells, 2‑DG collapses the {PD‑1\weirdplus} {LAG‑3\weirdplus} population from $\sim94\%$ control to $\sim0.8\%$, leaving most cells double‑negative. In memory {CD8\weirdplus} cells, 2‑DG reduces LAG‑3 strongly (from ~92 percent to $\sim57\%$ LAG‑3\weirdplus) and decreases the double‑positive subset to $\sim 55\%$, with total {PD‑1\weirdplus} only modestly reduced (from $\sim99\%$ to $\sim89\%$). In glucose‑free media (glycolysis restriction without glycosylation block), naïve cells show only a partial reduction of LAG‑3 (to $\sim38\%$) and little change in PD‑1; memory cells are largely unchanged. Thus, 2‑DG’s effects are far stronger than glucose deprivation and are consistent with impaired N‑linked glycosylation rather than energy shortage. The pattern dovetails with your earlier findings in {CD4\weirdplus} T cells: a brief, early glycosylation perturbation leaves a durable imprint that shifts fate and checkpoint expression.
\end{assistantturn}
\end{chatbox}

\begin{figure}[ht]
\centering
\includegraphics[width=0.65\linewidth]{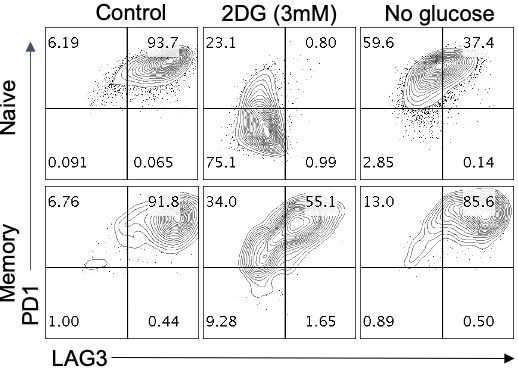}
\caption{Naive and memory CD8+ T cells were activated with anti-CD3/CD28 beads in media alone, 3mM 2DG, or in glucose-free media.Representative flow cytometry plots of PD1 and LAG3 surface expression on day 4- post activated naive (top) and memory T cells (bottom) in indicated conditions were shown.}
\end{figure}

In the follow-up, GPT-5 Pro provided another key mechanistic interpretation of why LAG-3 and PD-1 could be lower on {{CD8\weirdplus}} T cells due to glycosylation impairment, given that both are heavily N-glycosylated. It also provided a second and very insightful mechanistic explanation in which reduced signal strength could attenuate TCR signaling during 2-DG exposure during the activation phased of these cells and thus this modulation in signaling could have prevented upregulation of PD-1 and LAG-3, especially in naive T cells. Remarkably, we have direct and unpublished experiments showing this exact reduction in signal strength, which is remarkably much more pronounced in naive T cells compared to memory subset. Both PD-1 and LAG-3 are key checkpoint inhibitors that dampen responses against cancer cells. Indeed, the major cancer immunotherapy treatments target PD-1 or its ligand to enhance the responses of {{CD8\weirdplus}} T cells toward cancer cells.

\begin{chatbox}
  \begin{assistantturn}

  \vspace{0.5em}

\textbf{Mechanistic interpretation}. 
2‑DG is a glucose analog that both limits glycolysis and interferes with N‑linked glycosylation, a post‑translational modification required for correct folding, trafficking, and surface stability of many receptors. LAG‑3 and PD‑1 are heavily N‑glycosylated. The drastic, mannose‑rescuable effects you already demonstrated in CD4\weirdplus) cells, combined with the present observation that no‑glucose fails to phenocopy 2‑DG, point to glycosylation impairment as the dominant mechanism here. Two layers likely operate:
  \vspace{0.5em}
  
	1.	Direct receptor effect. Hypoglycosylation reduces proper surface expression and stability of LAG‑3 (and to a lesser extent PD‑1, especially in naïve cells), rapidly lowering measured positivity.
\vspace{0.5em}

	2.	Signal‑strength effect. By hypoglycosylating TCR/CD28 and IL‑2Ralpha (CD25) during priming, 2-DG attenuates early TCR/co‑stimulatory and IL‑2/STAT5 signals that otherwise drive robust PD‑1/LAG‑3 induction. This is more evident in naïve cells, which rely on strong primary signals; memory cells, with lower activation thresholds, still upregulate PD‑1 but lose much of their LAG‑3.

  \vspace{0.5em}

\end{assistantturn}
\begin{assistantturn}

\textbf{High‑level conclusions and impact}

 A short, early 2-DG pulse is a powerful dial that reduces inhibitory checkpoint display (especially LAG‑3) on {CD8\weirdplus}) T cells, with a stronger effect in naïve than in memory cells. The biology is glycosylation‑dependent rather than energy‑deficit‑dependent. Because PD‑1/LAG‑3 co‑expression correlates with early dysfunction/exhaustion, transient 2-DG during priming may produce {CD8\weirdplus} products that are less prone to early exhaustion and better poised for sustained activity after drug washout.
 
 \vspace{0.5em}

Immunotherapy implications. For CAR‑T or TCR‑engineered products derived from memory {CD8\weirdplus} cells, a brief 2-DG exposure during the first 24–48 hour could generate cells with lower LAG‑3 and slightly lower PD‑1, potentially enhancing resistance to exhaustion under antigen pressure and improving serial killing over days. This matches your CD4\weirdplus) memory dataset, where early 2‑DG favored type‑17‑like, memory‑leaning states that persist after washout. Conversely, for autoimmunity the data caution that interfering with glycosylation during priming could produce effector‑competent, checkpoint‑low {CD8\weirdplus}) responses that exacerbate pathology.

\end{assistantturn}
\end{chatbox}

Importantly, I asked GPT-5 Pro to simulate an experiment in which we would genetically engineer these {CD8\weirdplus} T cells with anti-CD19 CAR molecules to target {CD19\weirdplus} lymphomas, and to predict how prior 2-DG treatment would impact the cytotoxicity of these cells toward the cancer cells. Because we had already performed this experiment, also unpublished, I was able to validate the prediction from GPT-5 Pro, and it was nearly perfect in predicting that cytotoxicity would be enhanced. It also proposed mechanistic insights and follow-up experiments that we would need to perform before publishing these results. In this regard, GPT-5 Pro made sufficient contributions to this work to the extent that it would warrant its inclusion as a co-author in this new study.

\begin{chatbox}
\begin{assistantturn}

\textbf{What this predicts for anti‑CD19 CAR memory {CD8\weirdplus} cells (simulation)} 

Memory {CD8\weirdplus}) cells are activated, transduced with an anti‑CD19 CAR, expanded for 2 weeks in IL‑2, washed free of drug, then co‑cultured with {CD19\weirdplus} lymphoma cells for 3 days. Outcomes are read as specific lysis and serial‑killing capacity. To isolate functional differences, analyses gate on {CAR\weirdplus} cells, to control for any 2-DG effects on transduction efficiency. Mechanistic expectations applied to CAR‑T context. The 2-DG pulse yields lower LAG‑3 and modestly lower PD‑1 on day 4, which typically translates to less activation‑induced dysfunction over repeated target encounters. Although acute glycolysis supports immediate cytotoxic degranulation, your exposure is transient and occurs two weeks before the assay; metabolic capacity should recover, while the checkpoint/lineage program imprinted early is retained. Therefore, initial killing at 24 h may be similar or slightly lower, but cumulative 72 h killing should be higher with 2-DG priming, especially at low E:T ratios.

\end{assistantturn}
\end{chatbox}

\subsection{Key takeaways and implications from this case study of GPT-5 Pro}

In summary, GPT-5 Pro provided remarkable key insights and hypothesis based on our experimental data. If we had had these interpretations and the recommended next experimental plan from GPT-5 Pro, we would have resolved or hypothesized the mechanistic insights within minutes upon data analysis. Importantly, we still missed key insights provided by GPT-5 Pro, despite our deep expertise of this field. In addition, GPT-5 Pro suggested several experiments that we will need to perform to publish these findings, that's a co-investigator level contribution. More specific takeaways and implications are:

\begin{enumerate}
\item \textbf{Precision interpretation of complex biology.} GPT-5 Pro rapidly connected the observed phenotypes to a mechanistic hypothesis that distinguished glycolysis blockade from impaired N-linked glycosylation, proposed the IL-2 receptor pathway as the driver, and predicted that memory rather than naïve T cells carry the effect.

\item \textbf{Testable next steps that held up in the lab.} GPT-5 Pro produced a clean decision tree of experiments, including the mannose rescue to restore glycosylation, metabolic readouts, and epigenetic assays. This shows the model can generate highly relevant hypotheses that are testable in wet-labs.

\item \textbf{Mechanism-first thinking that avoids false trails.} By separating selection effects from programming effects, and by proposing controls that disentangle glycosylation from energy restriction, GPT-5 Pro reduced the risk of chasing attractive but potentially unnecessary experiments, which would have wasted many months of testing.

\item \textbf{AI-guided bioengineering for cell therapies.} By correctly predicting that a brief 2-DG exposure during priming would lower PD-1/LAG-3, preserve cytotoxic potential, and enhance serial killing in anti-CD19 CAR memory {CD8\weirdplus} T cells, GPT-5 Pro illustrates how foundation models can propose concrete, testable tweaks to CAR-T cell development protocols that enhance therapeutic performance against cancer or autoimmune diseases by rapidly iterating a variety of conditions in silico, before wet-lab validation.

\item \textbf{Long-term impact on biomedicine.} As models like GPT-5 Pro become native to lab operating systems, we should expect: faster mechanism discovery across immunology, oncology, and metabolism; cheaper negative results because failed branches are pruned in silico; more reproducible science due to selecting better hypothesis and well desisgned experimental approaches. The net effect will be a much higher discovery rate per experiment and a shorter route from observation to discovery to intervention, thus profoundly accelerating the biomedical scientific process.
\end{enumerate}

%% file: content/04-Zhivotovskiy.tex
\section{From density estimation and convex geometry 
to multi-objective
optimization -- Nikita Zhivotovskiy}
\label{sec:\secprefix-04}

The experiment discussed here originally dates to late August 2025, before a number of subsequent applications circulated more broadly. The goal was to apply GPT-5 not to obtain a direct literature answer, but to see whether a newly proved geometric statement would immediately surface adjacent literature and applications. This was useful to the authors because plausible extensions were unclear and would otherwise require asking a number of experts (and luck) to find the right connections. 

Our preliminary conclusion from this experiment is that, given only a core mathematical statement, GPT-5 can rapidly surface nontrivial and technically aligned links across areas (here, multiobjective optimization and approximate Pareto sets), providing context for new applications.

First, we record the statement that motivated the experiment. Let $K$ be a compact subset of $\mathbb{R}_+^d$ and $\alpha \ge 1$. We say that a subset $A \subset K$ is an \emph{$\alpha$-ratio cover} of $K$ if for every $\theta=(\theta_1,\ldots,\theta_d)\in K$ there exists $\phi=(\phi_1,\ldots,\phi_d)\in A$ such that $\theta_j \le \alpha\,\phi_j$ for all $j=1,\ldots,d$.

\begin{theorem}[Compton, Lugosi, Mourtada, Qian, Zhivotovskiy, 2025]
\label{thm:ratiocover}
For every $d \ge 1$ and every convex, compact set $K \subset \mathbb{R}_+^d$, there exists a subset $A \subset K$ with at most $2^{8d}$ elements that is a $32$-ratio cover of $K$.
\end{theorem}

Note that GPT-5 was not involved in the proof of the result above. The original motivation for Theorem~\ref{thm:ratiocover} is the statistical problem of density estimation for mixtures. Let $\{f_1,\ldots,f_d\}$ be densities on $\mathbb{R}^p$ and, for mixing weights $\theta\in K\subset \Delta^{d-1}$ with $\Delta^{d-1}=\{\theta\in\mathbb{R}_+^d:\sum_{i=1}^d \theta_i=1\}$, consider the mixture $p_\theta(x)=\sum_{i=1}^d \theta_i f_i(x)$. If $A$ is an $\alpha$-ratio cover of $K$, then for any $\theta\in K$ there exists $\phi\in A$ with $\theta\le \alpha\,\phi$ coordinatewise; consequently, $p_\theta(x)\le \alpha\,p_\phi(x)$ for all $x$, hence $\frac{p_\theta(x)}{p_\phi(x)}\le \alpha$ pointwise and, in particular, $\mathrm{KL}(p_\theta\|p_\phi)\le \log \alpha$. This helps with KL-based covering bounds for $\{p_\theta:\theta\in K\}$.

When obtaining a result of the form of Theorem~\ref{thm:ratiocover}, one naturally asks whether it is known and where it applies. What GPT-5 was helpful to do, once the authors had Theorem~\ref{thm:ratiocover}, was to surface related results that broaden the scope into theoretical computer science. To keep the experiment simple and clean:
\begin{enumerate}
\item the work \cite{compton2025geometry} was not public at that time, so its statements and possible connections were not available to the model;
\item for this note, the queries were rerun in a temporary, new session to avoid contamination from prior searches/interactions.
\end{enumerate}

\begin{figure}[h]
\begin{chatbox}
  \begin{userturn}
[Theorem \ref{thm:ratiocover} and the preceding paragraph are put in context.]
\newline

What are the similar known quantitative results? List the most relevant results in different areas (geometry, statistics). Make a short list with important/central papers for references/topics
  \end{userturn}

[Reasoned for 8m 4s]
  \begin{assistantturn}
 
[...]

Your ``$\alpha$‑ratio cover'' is essentially the (multiplicative) $\varepsilon$‑approximate Pareto set notion used in multicriteria optimization (dominance up to a $(1+\varepsilon)$-factor in each coordinate).

\begin{itemize}
    \item Papadimitriou–Yannakakis (FOCS 2000). Introduced the modern framework: conditions under which a polynomial‑size $(1+\varepsilon)$-approximate Pareto set exists and can be computed; foundational for multiplicative coordinate‑wise coverage.
    \item [...]
\end{itemize}
\end{assistantturn}

\end{chatbox}
\caption{An excerpt from GPT-5's reply. The full response can be found in \cite{ZhivotovskiyGPT}.}
\label{fig:nikita}
\end{figure}

\begin{figure}[h!]
    \centering
    \includegraphics[width=0.7\textwidth]{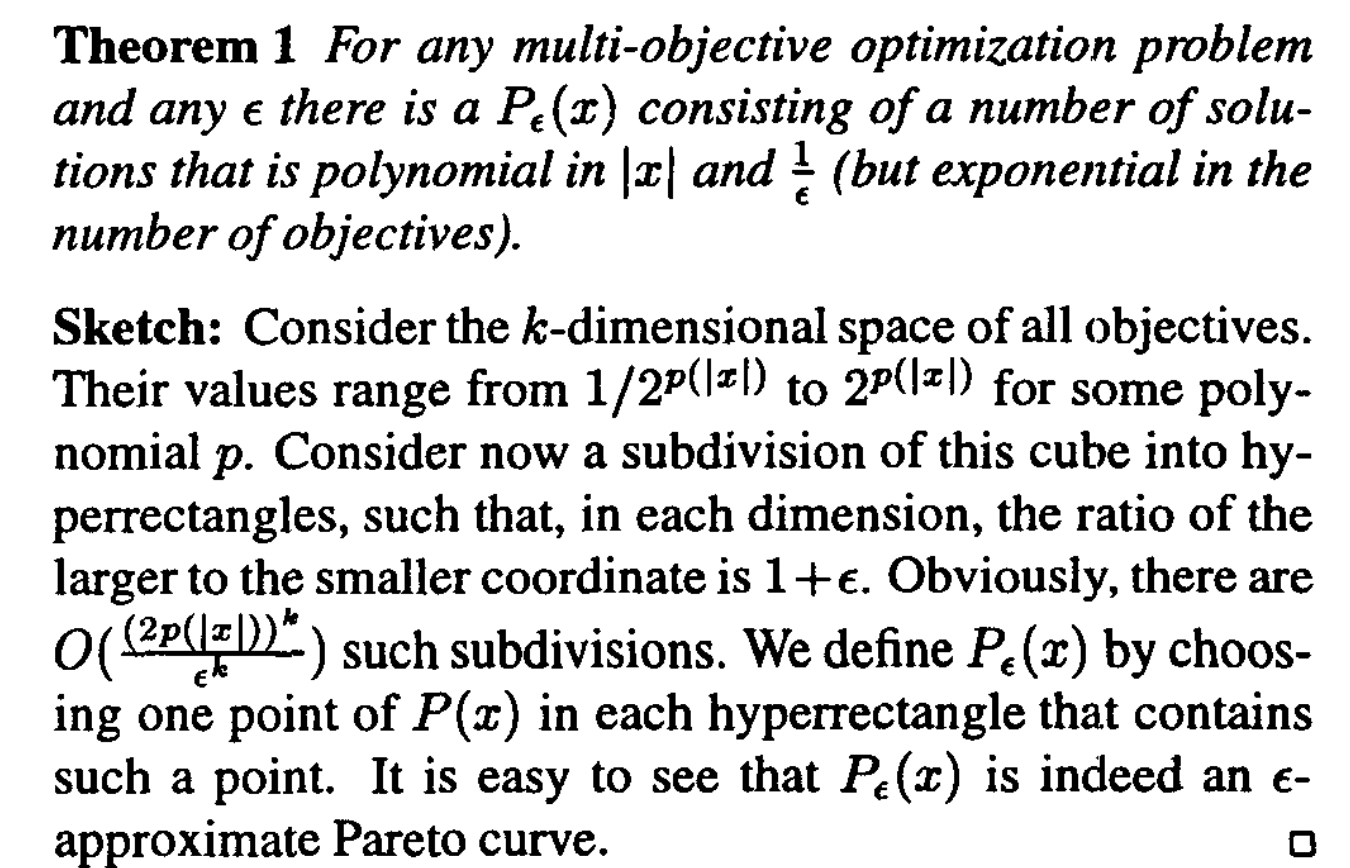}
    \caption{A result of \cite{papadimitriou2000approximability} closely related to Theorem~\ref{thm:ratiocover}.}
    \label{fig:papadimitriou}
\end{figure}

As can be seen in Figure \ref{fig:nikita}, among a series of other relevant results, the August 2025 exchange prominently highlighted the seminal work of \cite{papadimitriou2000approximability} on multiobjective optimization.
At first sight (see Figure \ref{fig:papadimitriou}), this seems unrelated to Theorem~\ref{thm:ratiocover} and could be mistaken for a hallucination. However, unpacking their proof shows that their result can be phrased as a coordinatewise $(1+\varepsilon)$-ratio cover on $K\subset[1/R,R]^d$ of size $O\,\left(\left(\log R/\varepsilon\right)^{d-1}\right)$. In contrast, under convexity of $K$, Theorem~\ref{thm:ratiocover} yields a constant factor $32$ and a size bound $2^{8d}$ independent of $R$. In the Papadimitriou--Yannakakis framework, $K$ denotes the set of achievable objective vectors; convexity is not assumed in general (though convex instances do arise).

A substantial part of \cite{papadimitriou2000approximability} is devoted to constructing near‑optimal $\varepsilon$‑approximate Pareto sets. Motivated by this connection, we extended Theorem~\ref{thm:ratiocover} to $(1+\varepsilon)$‑covers of size $O\,\big((\log(1/\varepsilon)/\varepsilon)^{d-1}\big)$ that, for convex $K$, are independent of the range parameter $R$. Consequently, in convex instances this removes the $\log R$ factor from the classical $O\,\big((\log R/\varepsilon)^{d-1}\big)$ bounds and provides a benchmark for convex multiobjective optimization.

%% file: content/05-SawhneySellke.tex
\newpage

\section{Erd\H{o}s problems (part 1/2) -- Mehtaab Sawhney and Mark Sellke}
\label{sec:\secprefix-05}

\subsection{Introduction}
Paul Erd\H{o}s was a prolific mathematician whose mathematical impact ranges across diverse areas of mathematics including combinatorics, number theory and analysis. Along with publishing more than 1500 mathematical papers, Erd\H{o}s posed a substantial number of mathematical conjectures with several of these conjectures becoming central problems in mathematics. 

Recently Thomas Bloom \cite{BloWeb} created an online index of Erd\H{o}s problems, which has greatly facilitated the process of identifying and solving the problems and opened the door to numerous online collaborations.
The role of large language models in our story already begins here with the creation of the website in early 2024, which relied on a previous version of ChatGPT (\url{https://www.erdosproblems.com/faq}) to help write the code underlying the website.
Each listed problem has its own webpage, with a status (roughly speaking, either ``open'' or ``solved'') as well as a discussion thread for potential collaborations.
The list is continually growing, with 685 ``open'' problems out of 1105 total as of October 31, 2025; the problems posed by Erd\H{o}s over the years are scattered across many disorganized sources.
Unfortunately the solutions are also often similarly scattered, and so it can be very challenging to determine for sure whether a particular problem is actually open, or if it was instead solved decades ago and forgotten to most mathematicians. This has resulted in a non-trivial amount of futile effort by mathematicians trying to solve problems that turned out to already be known. For instance as a graduate student, the first author spent a bit of time thinking about \cite[Problem~\#490]{BloWeb} and even mentioned it an informal meeting with other graduate students as a beautiful problem. Eventually the first author was able to locate a reference by \cite{Sze76} solving the problem, with the vague title ``On a problem of P. Erd\H{o}s''.
Thus, in practice a problem being listed as ``open'' roughly indicates that at least $1$ professional mathematician attempted and failed to find a previously published solution by searching the internet.
We discuss our successful attempts to improve the status of this database using GPT-5, which led to the following progress:
\begin{itemize}
    \item 
    Locating previously published solutions to $10$ problems not previously marked as known: 223, 339, 494, 515, 621, 822, 883 (part 2/2), 903, 1043, 1079.
    \item 
    Reporting noteworthy partial progress in the existing literature for $10$ other problems: 32, 167, 188, 750, 788, 811, 827, 829, 1017, 1011.
    \item 
    Correcting a misprint in the statement of problem 1041.
    \item 
    Generating a new idea for Problem \#848, which, together with previous suggestions by online commenters van Doorn, Weisenberg, and Cambie, enabled us to settle the problem.
\end{itemize}

In this section we focus on several examples from the first two bullets above.
Problem \#848 and its solution are presented later in Section~\ref{sec:\secprefix-09}.

\subsection{Literature search}
The database of Erd\H{o}s problems serves as a testbed for the abilities of GPT-5. 
For every problem listed as ``open'', we asked GPT-5 to locate a solution if it existed, and otherwise to report significant partial progress that merits being mentioned on the website. 
This was done using a combination of the ChatGPT interface, the OpenAI API, and internal OpenAI tools; the latter two allow automated queries of many problems in parallel, and automated filtering of responses also using GPT-5; aside from the automated aspect, we expect similar results for any particular problem could have been obtained purely using the ChatGPT interface. 
The prompts used were completely problem agnostic, consisting of a link to or a screenshot of the problem, and an instruction to search the internet for a solution or partial progress, despite the ``open'' designation on the website.\footnote{We were not alone in such endeavors; see e.g. \url{https://www.erdosproblems.com/forum/thread/737}, \url{https://www.erdosproblems.com/forum/thread/936}, \url{https://www.erdosproblems.com/forum/thread/1008}.}

The references provided were generally accurate and easy to verify manually.
We did not observe any cases in which GPT-5 pretended to have found a correct reference (though in some cases it was overly enthusiastic about partial progress it had found).
We also emphasize that the problem numbers used correspond to their labeling on the database, beginning in early 2024. 
In particular these numbers cannot be used in websearches to locate pre-2024 literature on a given problem.

Typical practice for a mathematician attempting to find literature is either following long chains of references or attempting to ``guess the correct search phrase'' hoping to find relevant literature. With GPT-5, one can skip this artificial middle step and simply plug in the statement! Furthermore in mathematical practice, one is often tasked with locating literature in a separate field ``which surely exists''; GPT-5 can do so provided often vague descriptions or candidate ``theorem'' statements.

We highlight a few representative cases. The first is problem \cite[Problem~\#339]{BloWeb}, the surprising success of which was the impetus for our efforts.

\begin{center}
\fbox{%
\parbox{15cm}{Problem \#339 (\cite{EG80}): \emph{Let $A\subseteq \mathbb N$ be an additive basis of order $r$, i.e. such that every sufficiently large integer is the sum of at most $r$ elements of $A$. Must the set of integers representable as the sum of \textbf{exactly $r$
 distinct} elements from $A$
 have positive lower density?} (I.e., must this set occupy a non-vanishing fraction of the integers $\{1,2,\dots,N\}$ up to $N$, for all sufficiently large $N$?)}%
}
\end{center}

A difficulty in searching for references is that Problem \#339 is raised in a 100-page paper of Erd\H{o}s--Graham \cite{EG80} which lists many other problem and has roughly $700$ citations.
Sorting through to find which of these are relevant for the specific problem above would be extremely time-consuming for a human.
However, GPT-5 Pro was able to find the solution in the first query we tried, given only a screenshot of the problem webpage. 
The discussion in which GPT-5 Pro found the solution is here: \cite{SawhneyGPT}. 
We note that before the solution was found, the problem attracted some discussion among participants on the online forum \url{https://www.erdosproblems.com/forum/thread/339}.

\subsubsection*{Further Examples}

Here we present three more examples, which further illustrate the challenges present in mathematical literature search.
We reiterate that the numbers below correspond only to the recently created online database of Erd\H{o}s problems, and thus are not useful for searching the pre-2024 literature.

\begin{center}
\fbox{%
\parbox{15cm}{Problem \#515 (\cite{Erdos1961}): \emph{Let $f(z)$
 be an entire function, not a polynomial. Does there exist a path $\gamma$ tending to infinity such that, for every $\lambda>0$, the integral
$\int_{\gamma} |f(z)|^{-\lambda} dz$
is finite?}}%
}
\end{center}

\begin{center}
\fbox{%
\parbox{15cm}{Partial result toward Problem \#750 (\cite{erdos1969problems}): \emph{Fix $\varepsilon>0$.
Must there exist a graph $G$ of infinite chromatic number so that every $m$-vertex subgraph has an independent set with at least $(\frac{1}{2}-\varepsilon)m$ vertices?}}%
}
\end{center}

\begin{center}
\fbox{%
\parbox{15cm}{Problem \#1043 (\cite{erdos1958metric}): \emph{Let $f\in\mathbb C[x]$ be a monic polynomial. Must there exist a straight line $\ell$
 such that the projection of $\{z:|f(z)|\leq 1\}$ onto $\ell$ has length at most $2$?
}}%
}
\end{center}

GPT-5 located references solving the above problems, respectively in \cite{lewis1984growth}, \cite{erdos1982almost}, \cite{pommerenke1961metric}.
Identifying these decades-old papers required GPT-5 to go far beyond the functionality of a search engine, and indeed to read each of these papers in detail and apply a genuine understanding of mathematics.

In the first case, the work \cite{lewis1984growth} actually proves something more general than the statement of Problem \#515. 
Namely, the authors study integrals of $e^{-F(z)}$ where $F:\mathbb C\to\mathbb R$ is a \emph{subharmonic} function, meaning its average value on any circle is smaller than its value at the center.
It is a classical fact that $F(z)=\log |f(z)|$ is subharmonic when $f$ is entire, which means that the results of \cite{lewis1984growth} do imply Problem \#515.
However this connection is not the focus of the paper, and indeed neither Erd\H{o}s nor his above problem are mentioned.
This use of different vocabulary for similar underlying concepts is a common challenge in searching the mathematical literature, and would have made \cite{lewis1984growth} quite hard to find using a standard websearch.
Indeed the main feature of Problem \#515 is the integral expression, which would be quite challenging to search for directly.
GPT-5 in fact later located a corroborating reference, a (256 page!) survey \cite{hayman2018research} that mentions the connection between \cite{lewis1984growth} and Problem \#515 on page 27.

Additionally, there is a subtle technical point in connection between \cite{lewis1984growth} and the problem it solves, which is that the function $\log|f(z)|$ has a singularity wherever $f(z)=0$.
This means that to apply the result from \cite{lewis1984growth} to Problem 515, it is crucial that their definition of a ``subharmonic'' function allows such singularities.
The paper requires a careful reading to confirm that their definition of subharmonic is sufficiently general for the required implication; here as well, GPT-5 assisted us in pointing out parts of the paper that made the intended definition clear.

For Problem \#750, the result from \cite{erdos1982almost} actually concerns the slightly different problem of finding a \emph{bipartite} subgraph, and shows such subgraphs exist with $(1-\varepsilon)m$ vertices. GPT-5 both found this reference and observed that taking the larger side of the bipartition solves the Erd\H{o}s problem in question. This (easy) deduction was done by GPT-5 during its literature search without additional prompting, and to our knowledge came from GPT-5's own reasoning.
Needless to say, an pre-LLM websearch would be unable to supply this missing argument to connect the two results.

\begin{figure}[t!]
    \centering
    \includegraphics[width=0.9\textwidth]{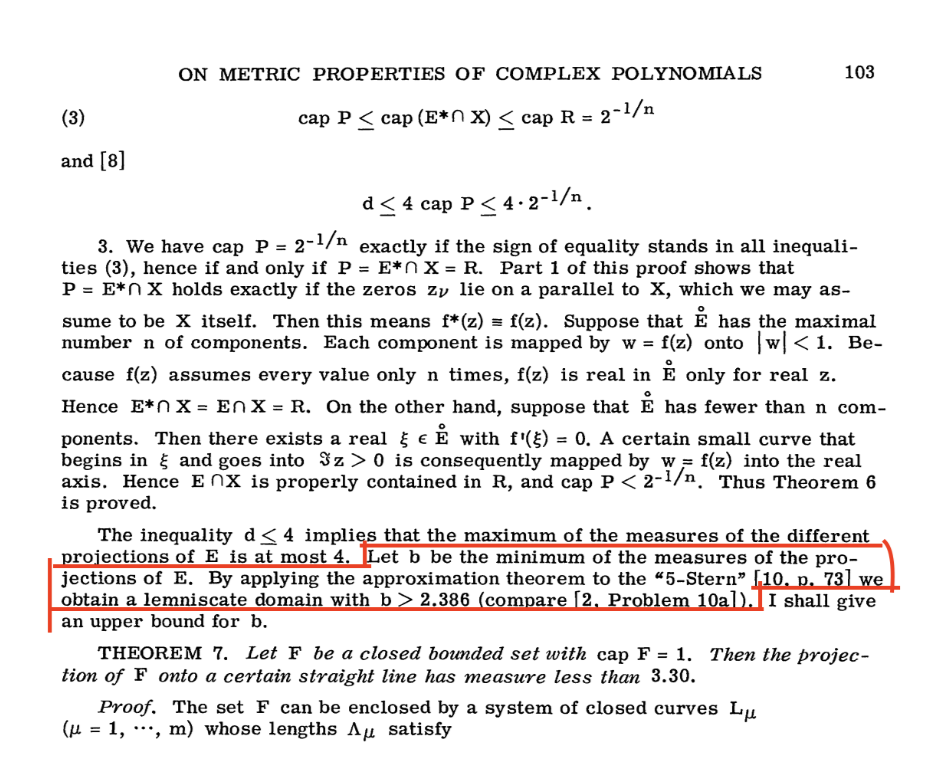}
    \caption{Pommerenke's solution to Problem \#1043, appearing on page 7 of \cite{pommerenke1961metric}. The proof follows by combining the ``approximation theorem'' from \cite{pommerenke1961metric} and Pommerenke's previous work \cite{pommerenke1959kapazitat} (cited as [10, p. 73] in the red block above). See \url{https://www.erdosproblems.com/forum/thread/1043} for a more detailed outline.}
    \label{fig:my_image_label}
\end{figure}

Problem \#1043 is stated in \cite{erdos1958metric} together with 15 other problems; several of these including 1043 were solved in \cite{pommerenke1961metric}.
Unlike the others, Problem \#1043 is addressed in a brief side comment between two theorems, and was consequently overlooked by many including the MathSciNet review of \cite{pommerenke1961metric}.
Additionally it turns out that \cite{pommerenke1959kapazitat} (displayed as reference [10] in the screenshot of \cite{pommerenke1961metric}) actually contains the main work in solving Problem \#1043, and is written in German. GPT-5 translated and explained the proofs from \cite{pommerenke1959kapazitat} to us so that we could verify them ourselves.

\subsection{Looking forward}
Motivated in part by our efforts, other researchers including Thomas Bloom and Terence Tao soon started posting \emph{negative literature searches} to the Erd\H{o}s problem website, reporting that GPT-5 and other large language models were \emph{unable} to locate an existing solution to various problems. This provides a convenient, crowd-sourceable ``soft certificate'' that the solution is unlikely to appear in the published literature. See e.g.\ \url{https://www.erdosproblems.com/forum/thread/689}, \url{https://www.erdosproblems.com/forum/thread/915},
\url{https://www.erdosproblems.com/forum/thread/990}.

The landscape of mathematical results is substantially broader than any one mathematician plausibly has available at their fingertips. GPT-5 therefore provides the practicing mathematician a new mechanism to access the collective breadth of the mathematical literature.

%% file: content/12-Anonymous.tex
\section{Clique-avoiding codes: a cautionary tale}
\label{sec:\secprefix-12}

\newcommand{\bin}{\binom{n}{2}}
\newcommand{\1}{\mathbf{1}}
\newcommand{\orf}{\mathrm{OR}}
\newcommand{\thr}{\mathrm{Thr}}
\newcommand{\zo}{\{0,1\}}

\emph{The following was communicated by Venkatesan Guruswami and Parikshit Gopalan, edited by us and included with their permission}

A binary linear code is a subspace of $\mathbb F_2^N \simeq \{0,1\}^N$ that avoids any nonzero vector of low Hamming weight. Here we consider a variant where the code must avoid a collection of structured vectors, namely all cliques in an $n$-vertex graph. This question is originally due to \cite{GY14}, motivated by the study of maximally recoverable codes, although it was never published. 

Let us define the problem formally.
Let \([n]=\{1,\dots,n\}\) and index the \(\binom{n}{2}\) coordinates of \(\mathbb F_2^{\binom{n}{2}}\) by unordered pairs \(\{i,j\}\) with \(1\le i<j\le n\).
For a vertex subset \(S\subseteq[n]\), let \(\chi_S\in \mathbb F_2^{\binom{n}{2}}\) denote the indicator of the edge set of the clique on \(S\), i.e., $\chi_S(\{i,j\})=\mathbf{1}\big[i\in S \ \text{and}\ j\in S\big]$.
A binary linear code \(C\subseteq \mathbb F_2^{\binom{n}{2}}\) is \emph{clique-avoiding} if
\(
\chi_S\notin C
\)
for every \(S\subseteq[n]\) with \(|S|\ge 2\).

\smallskip
The question posed to GPT-5 was the following:

\begin{figure}[ht]
\begin{chatbox}
\begin{userturn}
What is the minimum number of parity checks (i.e., the minimum co-dimension) denoted $r = r(n)$, of a clique-avoiding code as a function of the number of vertices $n$?
\end{userturn}
\end{chatbox}
\end{figure}

A simple probabilistic (random coding) argument, which GPT-5 also found right away (with some additive $O(1)$ slack), gives the following upper bound:
\begin{lemma}
\label{claim:simple}
We have
\(r(n) \le n\). 
\end{lemma}

It is tempting to think that the random coding bound is tight, but there is no obvious lower bound. This was the question we posed to GPT-5 (without really thinking about it ourselves). Initially, it was convinced that this bound was tight (up to an additive constant), and tried to convince us of this using a sequence of buggy arguments, resorting to linear algebra and proof by authority:

\begin{itemize}
\item The most amusing was a hallucinated response to the effect that one of us had asked this question on TCS Stack Exchange, and the answer given to them was $n$. Both these claims are incorrect. 
\item Another incorrect argument nevertheless contained the germ of the main idea. It was to consider quadratic forms $Q_1,\ldots, Q_r$ and then look at the {\em syndrome} $S(x) = (Q_1(x), \ldots, Q_r(x))$ and claim that $x \mapsto S(x)$ is an injection, hence $r \geq n$. This claim is also false. This assertion vaguely resembles arguments used for standard syndromes which are linear. 
\item When challenged about this, GPT-5 came up with an elegant (and correct) algebraic argument based on the classical Chevalley-Warning theorem, which established Theorem~\ref{thm:lower-bound} below. (On other attempts, it would come up with the correct formulation using quadratic polynomials modulo $2$, but then derive from it weaker results like $r \geq \lceil (n - \log(n))/2\rceil$.)
\end{itemize}

\begin{theorem}
\label{thm:lower-bound}
    For all $n \ge 2$, $r(n) \geq  \lfloor n/2\rfloor$.
\end{theorem}

We later discovered a construction (without using GPT-5) showing that the lower bound is \emph{exactly} tight: the simple probabilistic bound above is suboptimal, and in fact $r(n)=\lfloor n/2\rfloor$ for all $n$.

\subsection{An unexpected twist}

We were very excited to have an exact solution to a question which we had been curious about for years, especially because one direction of the proof was due to an AI. 
However, we soon learned that the same tight bound (as well as a different proof of the matching upper bound) had appeared on Arxiv nearly 3 years previously in the short paper \cite{alon2024graph}.
In fact, GPT-5's proof of the lower bound was exactly the same as Alon's! 
Thus, it appears that GPT-5 reproduced Alon's proof and passed it along to us without realizing its source.
As it turns out, a fresh attempt at querying GPT-5 Pro was able to recover this source, as depicted in Figure~\ref{fig:alon}.

\begin{figure}[ht]
\begin{chatbox}
\begin{userturn}
Do you know anything about the following question? This concerns the minimum co-dimension of a binary linear code $C \subset F_2^{n \choose 2}$ which avoids the indicators of all cliques (when the codeword positions are naturally viewed as edges of the complete graph on n vertices). 
\end{userturn}

\begin{assistantturn}
   \paragraph{Answer.}
Yes. The minimum possible co-dimension of a binary linear code 
\(C \subset \mathbb{F}_2^{\binom{n}{2}}\) that contains no clique 
indicators is exactly
\[
\boxed{\lfloor n/2 \rfloor}.
\]

\paragraph{Explanation (sketch).}

A linear code \(C\) of co-dimension \(r\) is the common kernel of 
\(r\) parity checks.  
Equivalently, we select graphs \(G_1,\dots,G_r\) on \([n]\), and 
a vector \(x \in \mathbb{F}_2^{\binom{n}{2}}\) is in \(C\) iff
\[
\langle x, 1_{E(G_i)} \rangle = 0 \quad \text{for all } i.
\]
Avoiding all clique indicators means that for every nonempty vertex
set \(T \subseteq [n]\), the clique \(K_T\) satisfies
\[
|E(K_T) \cap E(G_i)| \equiv 1 \pmod{2}
\quad\text{for at least one } i.
\]

\paragraph{Upper bound: \(\mathrm{codim}(C)\le \lfloor n/2\rfloor\).}

For odd \(n\), partition \([n-1]\) into blocks 
\(B_i = \{2i-1, 2i\}\), so there are \((n-1)/2\) blocks.  
Define \(G_i\) to consist of all triangles with base \(B_i\) and 
apex any vertex \(j > 2i\).  
One checks that every clique \(K_T\) has odd intersection with at 
least one \(G_i\), hence its indicator is excluded from \(C\).  
Thus co-dimension \((n-1)/2\) suffices.  
A small modification handles even \(n\), giving the general bound
\[
\mathrm{codim}(C) \le \lfloor n/2 \rfloor.
\]

\paragraph{Lower bound: \(\mathrm{codim}(C)\ge \lfloor n/2\rfloor\).}

Suppose \(r < n/2\).  
A Chevalley–Warning type argument produces a nonempty even-sized 
vertex set \(T\subseteq [n]\) such that 
\(|E(K_T)\cap E(G_i)|\) is even for every \(i\).  
Then the clique indicator of \(K_T\) satisfies all parity checks 
and lies in \(C\), a contradiction.  
Hence \(r \ge n/2\) for even \(n\), and therefore
\[
\mathrm{codim}(C) \ge \lfloor n/2 \rfloor.
\]

\paragraph{Conclusion.}

Both bounds match, so the exact minimum co-dimension is
\[
\boxed{\lfloor n/2 \rfloor}.
\]

For a full, elegant presentation, see N.\ Alon, \emph{Graph-Codes},
Theorem~1.6.

\end{assistantturn}
\end{chatbox}
\caption{An interaction in which GPT-5 Pro was able to locate the reference \cite{alon2024graph} solving our problem. This query was made after we first became aware of \cite{alon2024graph}.}
\label{fig:alon}
\end{figure}

Our experience illustrates a pitfall in using AI: although GPT-5 possesses enormous internal knowledge and the capability to locate even more using the internet, it may not always report the original information sources accurately.
This has the potential to deceive even seasoned researchers into thinking their findings are novel.
We expect that our experience is not unique, and urge others to take special care in attribution when working with LLM-assisted proofs.

%% file: content/06-Gowers.tex
\section{Recent experiences of using LLMs as research partners -- Timothy Gowers}
\label{sec:\secprefix-06}

This is a brief report on one or two recent experiences of using LLMs while actually doing mathematics, by which I mean thinking about research-level problems at an earlyish stage where I have some ideas but have not yet found an approach that feels likely to work. I can summarize the situation so far by saying that I have not (yet) had the experience of an LLM making a decisive contribution, but I have found them useful in the following ways.
\begin{enumerate}
\item If I suspect that a definition or mathematical observation I have come up with is in the literature, an LLM can often help me find it. GPT-5 seems to be significantly better at this than GPT-4. In particular, with GPT-5 my experience has been that the references are rarely hallucinated, and even the hallucinations can turn out to be pointers to references that exist and are useful.
\item I have sometimes formulated well-defined subproblems that look as though they shouldn't be too hard but also look as though they would take a bit of time to solve, and GPT-5 has solved them for me in a matter of seconds.
\item I have had reasonably precise ideas for solving problems that I have run past GPT-5 Pro, which has explained to me why they cannot work.  
\end{enumerate}

On the negative side, if I ask more open-ended questions, or offer more sketchy ideas for proof attempts, then that seems to encourage the more annoying characteristics of LLMs to come to the fore: they will tell me that my ideas do indeed work, and will write something that supposedly fleshes out the details but that does not withstand close scrutiny. This applies even to GPT-5 Pro, to which I have had access for the last few days (at the time of writing).

Taking this together, my current assessment of LLMs is that they are just beginning to be useful as research collaborators: one can bounce ideas off them in the way that one can with a human collaborator and one gets a response extremely quickly. As with human collaborators, even the less good ideas of an LLM can sometimes stimulate me to make progress. (I think of this as the ``That clearly doesn't work ... but wait a minute!" phenomenon.) So we have reached the stage where LLMs can speed up the process of thinking about a problem, especially if that problem is a little outside one's primary domain of expertise, but we have not yet reached the stage where an LLM is likely to have the main idea for solving a difficult problem. (That last statement has to be interpreted carefully. I may find a problem difficult because I am unfamiliar with the domain-specific tricks needed to solve it, but if the problem would be found easy by somebody with the relevant expertise, then it may well be easy for an LLM as well. Thus, by ``difficult'' I mean a problem for which the standard bag of tricks in the relevant area does not suffice.) 

I shall give two examples here of recent interactions with GPT-5 and GPT-5 Pro. I stress that in each case, the main contribution of the LLM was to save me a small amount of time: the observations it made and proofs it provided are ones that I would have expected to find for myself before all that long, but with the help of an LLM I did not to have to make any effort at all and I got the answers within seconds. 

It is interesting to think about whether I \emph{should} have made more effort before turning to an LLM for help. Everybody will have to make their own decision about this. My current feeling is that this issue is a bit like the issue of whether one should use calculators to do arithmetic. One is clearly a much better user of calculators if one knows how to do the calculations oneself and uses the calculators only as a time-saving device. Similarly, in these early days, I feel much more comfortable asking an LLM to help me if I think it is carrying out a task for which I have a good idea of how I would set about it and that I could almost certainly do without help, given a bit of time. I would be less comfortable simply feeding it a difficult problem and hoping for the answer (and also, for now, much less confident that I would get a useful answer if I did so). 

\section*{Example 1: a compact subset of $L_2$}

I was trying to solve a problem where I had a somewhat complicated non-linear map $T:L_2\to L_2$ and wanted to prove that the iterates $f, Tf, T^2f,\dots$ of a certain function $f$ converge. From what I knew about $T$, it was fairly clearly enough to show that the iterates had a convergent subsequence, so I was looking for a compact subset $K\subset L_2$ that I could argue contained all the iterates. 

I knew that in an arbitrary Banach space, one source of compact sets is convex hulls of null sequences, and after a little thought I realized that I could prove that in $\ell_2$, if I took any non-negative square summable sequence $s$, then the set $\{a\in\ell_2:|a|\leq s\}$ was compact. I had reason to believe that the iterates of $f$ would exhibit something like Gaussian decay, so I wondered whether a similar result was true for the set of functions bounded above by a Gaussian. 

I soon realized that that could not be the case, because one can take a sequence of more and more rapidly oscillating functions, and those will have no convergent subsequence. 

However, I had reason to believe that the iterates of $f$ under the map $T$ would not start oscillating more and more, so I did not lose hope. It occurred to me that the oscillations should make their presence felt on the Fourier transform side, and as the operation $T$ interacted well with the Fourier transform, it might be enough for me if the set of functions $f$ such that both $f$ and $\hat f$ are bounded above by a Gaussian is compact. 

This felt plausible to me, so I asked GPT-5 about it. After 22 seconds, it provided me with a proof, which relied on a lemma that I had not heard of called the Kolmogorov-Riesz lemma. The proof looked correct, given the lemma, and the lemma looked plausible. I checked online and found that it did indeed exist. I don't know how long it would have taken me to realize that the lemma was the right intermediate step (after which I could have used GPT-5 for its semantic search capabilities) but I estimate that this saved me anything between an hour and three hours of work. 

There is an epilogue to this story. I am collaborating on the project with two research students, who were also good at pointing out drawbacks with my proposals for proof approaches. In particular, one of them pointed out that the condition that $f$ and $\hat f$ should both be bounded above by a standard Gaussian would be too strong, since Hardy's uncertainty principle implies that in this case $f$ must itself be a multiple of a standard Gaussian. Thanks to this comment, I realized that there was a mismatch between the condition I thought I could obtain -- that both $f$ and $\hat f$ are bounded above by a Gaussian $\mathrm e^{-cx^2}$, but not necessarily with $c=1/2$ -- and the condition that I had asked GPT-5 about. So I then asked GPT-5 whether one still had compactness under this weaker assumption. Again, very quickly it told me the answer, which in this case was one that I would have realized for myself fairly quickly, but not as quickly as GPT-5. Unfortunately, the answer was negative: the Hermite functions (which are eigenfunctions of the Fourier transform) form a sequence satisfying the condition and having no convergent subsequence.  

I was fooled by this, because each Hermite function is a polynomial multiplied by $\mathrm e^{-x^2/2}$ and therefore decays more quickly than $e^{-cx^2}$ if $c<1/2$. However, while writing the previous paragraph I started to have my doubts, since what I was after was a sequence of functions all dominated by $Ce^{-cx^2}$ for some fixed $C$ and $c$, and although each Hermite function is \emph{eventually} dominated by $e^{-cx^2}$ when $c<1/2$, that is a weaker statement. Since I could now ask GPT-5 Pro I asked whether there could be a single function of the form $C\mathrm e^{-cx^2}$ with $c>0$ that dominates all Hermite functions. It told me that the answer was no and gave me the following convincing proof. The Hermite functions satsify the recurrence relation
\[xh_n=\sqrt{\frac{n+1}2}h_{n+1}+\sqrt{\frac n2}h_{n-1}.\]
Since the Hermite functions are orthonormal it follows that 
\[\int_{-\infty}^\infty x^2h_n(x)^2\,\mathrm dx=\frac{n+1}2+\frac n2=n+\frac 12.\]
It follows that for sufficiently large $n$ it is at least $C^2\int x^2\mathrm e^{-cx^2}\,\mathrm dx$ (since the latter is finite and independent of $n$), which proves that for sufficiently large $n$ $|h_n|$ cannot be dominated by $Ce^{-cx^2}$. 

I then asked GPT-5 Pro whether if $0<c<1/2$ the set of functions $f$ such that $|f|$ and $|\hat f|$ are dominated by $Ce^{-cx^2}$ is compact. It told me the answer was yes and gave a convincing looking proof, but one that I felt I needed to check, especially as it said that a certain step of the deduction followed from a ``semigroup version" of Hardy's uncertainty principle but gave no details about what that version actually said. When I asked for more detail it gave me a hallucinated reference but by an author who had written on closely related topics. I complained that I couldn't find the paper, and this time it gave me a reference to a paper that was highly relevant, though it got the names of the authors wrong. Unless I am making another mistake, this paper contains exactly the result I wanted: under the giving conditions it gives an upper bound for the Hermite coefficients of $f$ and this upper bound decays at an exponential rate. Thus, there is a null sequence with a convex hull that contains all such functions $f$, giving us the desired compactness. 

The final twist in the story is another remark by my human collaborators. I had been slightly careless in assuming that the set of all $f$ such that $|f|$ and $|\hat f|$ are bounded above by $Ce^{-cx^2}$ is invariant under $T$, so the compactness result we now have is not obviously sufficient for our purposes. Nevertheless, we have taken a significant step forward in our thinking about the problem, and a back-and-forth with GPT-5 and GPT-5 Pro greatly sped up the process.

\section*{Example 2: graphs with uncharacteristically small clique number}

A very nice problem I recently heard about asks whether there is an algorithm with the following properties.
\begin{enumerate}
\item It takes as its input a graph with $n$ vertices.
\item If the graph contains no clique of size $(3/2)\log_2n$, then it outputs 1.
\item For almost all graphs (meaning a $1-o(1)$ proportion, but even 99\% would be interesting), it outputs~0.
\item It runs in polynomial time.
\end{enumerate}
The reason one might hope for such an algorithm to exist is that random graphs have clique number approximately $2\log_2n$ with high probability, and will typically have many cliques of size $(3/2)\log_2n$, which will be highly dispersed, so getting rid of all of them seems difficult. Therefore, one expects that a graph with no clique of size $(3/2)\log_2n$ will have to have ``unexpected structure" of some kind that can be picked up by an efficient algorithm. (By contrast, adding an unexpectedly large clique is easy to achieve: one just adds it in at random. Such ``planted cliques" appear to be very hard to find if they are of size smaller than $\sqrt n/\log n$.)

I decided to try to think of ways of creating a class of graphs that did not have cliques of size $(3/2)\log_2n$ but that nevertheless looked random and therefore hard to detect efficiently. Each idea I had, GPT-5 Pro was able to knock down in seconds (always softening the blow first by telling me that it liked the idea). 

\subsection*{First idea: use linear algebra}

A well known result in combinatorics, and a surprising one when you first see it, is that the number of subsets of $\{1,2,\dots,n\}$ you can have of odd size if all their intersections are of even size is $n$. The proof is one line: interpret each subset as a 01-sequence in the usual way, and regard the sequence as a vector in $\mathbb F_2^n$. The parity of the intersection of two sets becomes the $\mathbb F_2$-inner product of the two corresponding vectors, and the usual proof that an orthogonal sequence of vectors is linearly independent can be easily adapted to show that sets of odd size and even intersections correspond to linearly independent vectors, and hence that there are at most $n$ of them.

If we form a graph where the vertices are all sets of odd size and two vertices are joined by an edge if their intersection has even size, then the above result tells us that there can be no clique of size greater than $n$. Since there are $N=2^{n-1}$ vertices, this graph has no clique of size $(3/2)\log_2N$ (for sufficiently large $N$). 

So far, it is easily distinguishable from a random graph: for example, it is much more regular than a random graph will typically be. I proposed to deal with that problem by passing to a random subset of the vertices of size $2^{\alpha n}$ for some $\alpha$ reasonably close to 1. 

GPT-5 Pro pointed out to me that the rank over $\mathbb F_2$ of such a matrix would be at most $n$, whereas the rank of the adjacency matrix of a random graph on $2^{\alpha n}$ vertices would typically be close to $2^{\alpha n}$, so there was a very easy algorithm for picking out my proposed class of matrices. 

\subsection*{Second idea: use angles between unit vectors} 

My first idea had failed because it was in some sense too algebraic, but it occurred to me that I knew another result about vectors that implies the existence of a graph without large cliques. It states that the largest number of vectors you can find in $\R^d$ such that any two of them have a negative inner product is $d+1$ (with equality for the vertices of a regular simplex centred at the origin). So I wondered about choosing $d$ in such a way that $d+1<(3/2)\log_2n$ and then choosing $n$ random unit vectors $u_1,\dots,u_n$ in $\R^d$, joining two of them if and only if they have negative inner product. This felt to me as it had a chance of working, because although the matrix with entries $\langle u_i,u_j\rangle$ has rank $d$ (with probability 1), we are applying to each matrix entry the rather violent and unalgebraic operation of replacing it by 1 if it is negative and 0 otherwise. 

However, GPT-5 Pro almost immediately gave me a test for detecting matrices created this way, based on the fact that the larger $\langle u_i,u_j\rangle$ is, the more likely it is for a random vector to have a positive inner product with both of them. Indeed, if $u_i$ and $u_j$ form an angle $\theta$, this probability is $1/2-\theta/2\pi$. From simple calculations it follows that in any graph obtained by my proposed method, there will be noticeably more independent sets of size 3 than one would expect in a random graph, and there will also be many pairs of vertices with neighbourhoods that intersect a lot more than expected. 

I then noticed that a test of this form also applies to the $\mathbb F_2^n$ construction, since if $x+y+z+w=0$ then any vertex $v$ that is joined to $x,y$ and $z$ is automatically joined to $w$. Passing to a subset of size $2^{\alpha n}$ with $\alpha$ fairly close to 1 leaves many quadruples $x,y,z,w$ of vertices that add to 0, which will have neighbourhood intersections that are much too large.

\subsection*{Third idea: replace linear algebra by polynomials}

I wondered rather vaguely whether one could rescue the more algebraic approach by devising some polynomial condition for two elements $x,y$ of $\mathbb F_2^n$ to be joined. That is, perhaps I could find some suitable polynomial map $P:\mathbb F_2^n\times\mathbb F_2^n\to\mathbb F_2$ (symmetric in $x$ and $y$) and put an edge between $x$ and $y$ if and only if $P(x,y)=1$.

This time GPT-5 Pro gave me a longish answer explaining that unless the degree of $P$ was huge the resulting matrix would still have low rank. That seemed believable but I did not check the argument carefully. Asking GPT-5 Pro the same question again (just about the rank of such a matrix) I obtain the following simple argument: this one would probably have taken me a few minutes to find for myself, so once again there is speed-up but not as much as with some of the other questions. Let $P$ have degree at most $d$. If we consider the $x$th row of the matrix, then its values are $P(x,y)$ as $y$ ranges over $\mathbb F_2^n$. For fixed $x$, $P(x,y)$ is a polynomial in $y$ of degree at most $d$, so it is spanned by the monomials $x_{i_1}\dots x_{i_s}$ with $s\leq d$, and therefore the row is spanned by these monomials when we consider them as functions from $\mathbb F_2^n$ to $\mathbb F_2$. In particular, the row space has dimension at most $\sum_{s=0}^d\binom ns$. For this not to be noticeably smaller than $2^{\alpha n}$ (I am imagining that I have also passed to a random subset of the vertices of size $2^{\alpha n}$) we would need $d$ to be linear in $n$.

But GPT-5 Pro also told me that if the degree of $P$ was as large as that, there would be no obvious reason for the graph not to contain a clique of size $(3/2)\log_2N$, where $N$ is the number of vertices. If I remember correctly, it didn't give a particularly compelling argument for that, but it agrees with my intuition to the point that I do not see a need to defend the assertion in detail.

\subsection*{Attempting to go in the other direction}

Given that I had a uniform explanation for the failure of my first two attempted examples, I then decided to have a conversation about whether a graph that passes certain strong tests for randomness based on the sizes of the intersections of neighbourhoods could be shown to contain a clique of size $(3/2)\log_2(n)$. The answers I received were unhelpful, which was not very surprising as my question was too open ended and very little appears to be known about quasirandomness conditions that give information about the counts of subgraphs of size as high as $(3/2)\log_2n$. 

\section*{Conclusion}

As a research supervisor I have a rule of thumb for when a contribution I make to the research of one of my PhD students is at the level where I should be a joint author. The rule is that if the student comes to discuss the problem with me and I have, in the course of that discussion, an idea that comes more naturally to me than to them and that turns out to be helpful, then that is not enough for joint authorship. But if I spend time struggling with the problem (of course, I will do this only if the project is officially a joint one) and during the course of that struggle I come up with an idea that required more than just standard expertise that I happened to have, then I have made a genuine contribution to the work. 

My experience so far with LLMs is that they are capable of playing this knowledgeable-research-supervisor role with me, which can be extremely useful given just how much knowledge they have, but that they are not yet at the level (or at least have not yet exhibited that level in my own interactions with them) at which a human mathematician who follows my convention above would ask for joint authorship.

%% file: content/07-Scherrer.tex
\section{Power spectra of gravitational radiation from cosmic strings -- Robert Scherrer}
\label{sec:\secprefix-07}

\subsection{The problem in context}

Cosmic strings are hypothetical one-dimensional topological defects that might
have formed at a cosmological phase transition in the early universe (see,
e.g., \cite{hindmarsh1995cosmic} for a review). After a burst of interest in the 1980s, interest in cosmic strings declined sharply.  However, there has been a recent revival of interest in these objects because they can serve as copious emitters of gravitational radiation such as the stochastic background recently observed by the Pulsar Timing Array.

In a previous paper \cite{StormScherrer2024}, a graduate student (David Storm) and I examined the spectrum of radiation that might arise from the class
of cosmic strings first examined by Garfinkle and Vachaspati 
\cite{GarfinkleVachaspati1987}.
While the total power emitted per unit solid angle of the gravitational radiation for these cosmic strings
was investigated in \cite{GarfinkleVachaspati1987}, the quantity relevant for the stochastic
background is the total power $P_n$ as a function of the $n$th harmonic.
The power emitted
at frequency $\omega_n = 4\pi n/L$ (where $L$ is the length of the loop) by a cosmic string with a mass per unit length $\mu$ in the direction of the unit vector $\hat r$ is given by \cite{GarfinkleVachaspati1987}
\begin{equation}
\label{Pn}
    P_n = \frac{32 G\mu^2}{\pi^3 n^2} \int d\Omega \frac{[1- (-1)^n \cos(n \pi e_1)][1- (-1)^n \cos(n \pi e_2)]}{(1-e_1^2)(1-e_2^2)}, 
\end{equation}
where $e_1 = \hat r \cdot \hat{a}$ and $e_2 = \hat r \cdot \hat{b}$,
and $\hat{a}$ and $\hat{b}$ are unit vectors whose directions characterize the string trajectory.  Since $\hat{a}$ and $\hat{b}$ are three-dimensional unit vectors, the problem is characterized entirely by the angle $\alpha$ between $\hat{a}$ and $\hat{b}$.  Square loops have $\alpha = \pi/2$, while smaller values of $\alpha$ correspond to more prolate rectangular loops.

In our earlier paper \cite{StormScherrer2024}, we examined a range of values for $\alpha$ and found numerically that ${dP_n}/{d\Omega}$ did not scale as $n^{-2}$ at large $n$, as had been previously argued \cite{DamourVilenkin2001}, but seemed
to diverge from the predicted result.  As our results disagreed with received wisdom, it was important to confirm them analytically.

\paragraph{Human route.} After roughly six months attacking the problem, I was able to derive a partial result, but only for the simplest case:  $\alpha = \pi/2$.  Furthermore, my result applied only to even $n$.  In this case, I found
\begin{equation}
\label{finalPn}
P_n = \left[ \frac{128 G\mu^2}{\pi^2 n^2}\right ] \left[\gamma + \ln \pi
+ \ln(n)\right].
\end{equation}
where $\gamma$ is Euler's constant.
The details of the derivation can be found in the appendix of \cite{StormScherrer2024}.

\subsection{The task given to GPT-5 Pro:  goal, prompts, and model behavior}

\paragraph{Goal.} While my eventual goal was to determine analytically the power spectrum for all $\alpha$, as a first step I asked GPT-5 Pro to derive $P_n$---or rather, the integral appearing in Eq.~\eqref{Pn}---for odd values of $n$.  I knew from our numerical results that in the large-$n$ limit, it almost certainly had the same form as Eq.~\eqref{finalPn}, but the methods I used to derive this result could not be applied to the odd-$n$ case.

\paragraph{Prompt.} In my initial prompt, I asked GPT-5 Pro to evaluate the integral in Eq.~\eqref{Pn} for odd $n$, and give the asymptotic behavior for large $n$.  The program hung up for quite a long time, giving me no details about its thought process.  After several hours I became frustrated and killed it.  I then repeated the prompt but with an added reference to the Storm and Scherrer paper.  The exact prompt was:

\begin{chatbox}
    \begin{userturn}
        I would like to analytically integrate the following integral over the sphere, so that theta goes from 0 to $\pi$ in spherical coordinates, and phi goes from 0 to $2\pi$.  The integral is integral of $[1+ \cos (n \pi e_1)][1+ \cos(n \pi e_2)]/[(1 - e_1 ^2)(1-e_2^2)]$ where $n$ is an odd integer, and  $e_1 = \cos(\theta)$ and $e_2 = \sin(\theta) \sin(\phi)$, where $\theta$ and $\phi$ are the usual coordinates in spherical coordinates: $\theta$ is the angle relative to the $z$ axis, and $\phi$ is the azimuthal angle.  I would like an exact solution for all odd $n$, but you may assume that $n$ is much larger than 1 if you cannot solve the general case.  Note that a solution for even $n$ was derived by Scherrer and Storm in their paper in Physical Review D.
    \end{userturn}
\end{chatbox}

I was particularly pedantic about the definitions of $\theta$ and $\phi$ because in spherical coordinates, physicists use the opposite convention from mathematicians, and I was not sure which convention GPT-5 Pro would default to.  I also took $n$ to be odd in Eq. (\ref{Pn}) and dropped the $(-1)^n$ factors.

\paragraph{Outcome.}  After reasoning for 40 minutes, GPT-5 Pro produced a result for large odd $n$ identical to the asymptotic result I had previously derived for large even $n$ [Eq.~\eqref{finalPn}].  It used a completely different method of solution from my own.  While my own solution involved splitting the integral into three pieces and taking an ad hoc approach to each piece, GPT-5 used an expansion in Legendre polynomials, and then extended this to a Bessel function expansion.  I had suspected that some sort of solution in Bessel functions was possible but had never found it myself. GPT-5 Pro also gave the leading order correction term to this formula, so that it becomes
\begin{equation}
4 \pi [\ln(n\pi) + \gamma - {\rm Ci}(n \pi)],
\end{equation}
where Ci is the cosine integral function.  I was not aware of this correction term, which is relevant for small values of $n$.
Further, GPT-5 Pro provided a table illustrating the numerical integration in comparison with the large-$n$ asymptotic result, showing that they diverge from each other at small $n$ (less than 5) but are almost identical at large $n$.  (I should mention that even the numerical integral is difficult---the integrand in angular coordinates is so spiky that it resembles a sea urchin).  GPT-5 Pro also asserted (correctly) that the integral in the prompt does not converge for even $n$, but this is because I left out the $(-1)^n$ factors from Eq. (\ref{Pn}).

For reference, the full interaction is available here: \cite{ScherrerGPT}.

\subsection{Result, implications, and next steps}

\paragraph{Results.} GPT-5 Pro derived a new analytical result that confirmed our earlier numerical simulations for the gravitational wave power spectrum from a class of cosmic strings.  Had this result been available to me earlier, I would have incorporated it into Ref. \cite{StormScherrer2024}.

\paragraph{Implications.} GPT-5 Pro is capable of solving complex analytic integrations that are beyond the reach of symbolic manipulation programs such as Mathematica.

\paragraph{Next Steps.}  The obvious next step will be to solve the general case for arbitrary $\alpha$.  If successful, this will provide a sufficiently important result to motivate a research publication.  I have accumulated a number of such unsolved interesting mathematical problems that have frustrated me over my 40-year research career.  Many of these seem particularly well-suited to AI solution.  I have long waited for this moment to arrive.

%% file: content/08-Spears.tex
\section{AI-assisted reduced-physics modeling of thermonuclear burn propagation -- Brian Keith Spears}
\label{sec:\secprefix-08}
\subsection*{Premise}

GPT-5 has evolved to be a quite capable research assistant. It is right now remarkable for its moderate depth of knowledge (relative to, say, a 2-decade veteran PhD researcher) striped across an incredibly broad range of topics. We test and exploit this capability here by recruiting its assistance to formulate an inertial confinement fusion problem that probes this breadth.

We aim to develop an improved understanding of the local physics of fusion ignition and burn propagation at the interface between hot and cold nuclear fuel -- a critical complex of processes that lead to the production of more energy by fusion than the energy absorbed by the ICF target \cite{ignition}. To this end, we set up a reduced-physics model that isolates the generation of a thermonuclear burn wave from the complex implosion dynamics of a full ICF implosion simulation \cite{hydra}.

We are interested in AI-driven assistance with several problem phases:
\begin{itemize}
    \item The setup of a simplified physics model with which we can develop intuition;
    \item The numerical simulation and solution of such a model;
    \item The design of numerical experiments to help us elicit the key physics sensitivities;
    \item Development of a theory that explains the numerical findings.
\end{itemize}

We have developed a simplified, static (no fluid motion) reaction--diffusion model for an ICF deuterium--tritium (DT) capsule to investigate the ignition and propagation of a thermonuclear burn wave into cold, dense fuel. This framework balances fidelity with transparency: it isolates the relative roles of thermal conduction, alpha-particle energy deposition, radiation losses, and density-profile shaping on wave propagation.

\subsection{Introduction}

To get going, I prompted GPT-5 to help me set up a model environment that I could use to explore thermonuclear burn propagation in inertial confinement fusion (ICF). This physics couples complex microphysics with steep thermodynamic gradients, and I would like to intentionally remove the dynamics that continually and crucially changes the power balance. The goal is simply to find the ideal fusion interface in the static case; later we can figure out how to map it into a real implosion and control it (via shock timing, drive preheating, etc.).

I start in my usual way, with what is probably (perhaps not affectionately) known as a ``boomer prompt'':
\clearpage

\begin{chatbox}
    \begin{userturn}
I want to explore the dynamics of a thermonuclear burn wave propagating into cold deuterium-tritium fuel in an imploding ICF capsule. I want to build a simplified, static (no material motion) model of the ignition and propagation of such a wave in spherical geometry.

Let's build a reaction diffusion equation to model it. I want the model to have 3 regions. The first region is a hot, low-density DT plasma (hot spot) that will initiate the burn wave. It should have properties similar to the central hot spot of a NIF implosion at peak ion temperature and should have similar radius. The next region will be relatively cool, dense DT fuel with a quadratic density profile (one parameter for average slope and one for curvature), ramping from the density of the hot spot to high-density cold fuel and a similar temperature ramp from hot to cold. The third and final region is uniform density (relatively) cold fuel, also at the conditions and thickness of a well shock-timed NIF implosion at peak shell compression (after rebound shock passage).

The reaction diffusion equation dynamics should capture clearly three effects: first, thermal energy should diffuse according to a good approximation of the density- and temperature-dependent thermal diffusivity; second, the reaction term should capture energy production by alpha particle generation and the non-local deposition of that alpha energy consistent with the condition-dependent stopping power of alpha particles in the nearby material; finally, there should be radiation loss which is transported far away from the local heat front.

With this model, we want to ask a key question. What density profile in region 2 most easily allows us to propagate a burn wave? We presume that for very shallow ramps, the hot spot can initially heat region two easily, but wave propagation ``flames out.'' For very steep ramps, we presume it requires more energy in the hot spot to get the wave going, but that it propagates more easily. The intuition, then, is that there is an optimal profile (average slope and curvature).

Set up this model with appropriate physical constants and assumptions. Embed it in an optimization framework that delivers the profile slope that provides the most robust initiation and propagation of a burn wave. Demonstrate the results with a few plots showing temperature wave propagation under a few scenarios, and add a figure-of-merit plot with axes difference in propagation depth (or speed) vs.\ fuel profile (slope and curvature).
\end{userturn}
\end{chatbox}

This initial prompt led to an extended conversation with GPT-5 Pro, with a lot of back and forth which I omit here. The remainder of this section summarizes the work that I was able to carry out in conjunction with GPT-5 Pro. Most of the results given here were produced by GPT-5; I merely provided it with some feedback and guidance. Interested readers may consult this long exchange with GPT-5 Pro here: \cite{SpearsGPT}.

\subsection{Model Formulation}

The result was a pretty good start at a static model -- the formulation is good, the terms are aptly formed and named, and the physics property choices are relevant. After some dialogue, we settled on familiar and more detailed physics data, all of which were offered by GPT-5 in advance of my requesting them. This is how the conversation would go with a national lab colleague and expert.

Together, we consider spherical symmetry with no material motion. The dependent variable is the local DT temperature \(T(r,t)\) in units of keV. Energy conservation for the DT plasma (neglecting \(PdV\) work) is expressed as
\begin{equation}
  \rho C_v \frac{\partial T}{\partial t}
  = \frac{1}{r^2} \frac{\partial}{\partial r}
    \left( r^2 \rho \,\chi(T,\rho)\,\frac{\partial T}{\partial r} \right)
  + Q_\alpha(T,\rho)
  - Q_{\text{rad}}(T,\rho),
\end{equation}
where \(C_v = \tfrac{3}{2}k_B/m_i\) is the specific heat per unit mass, \(\chi\) is the thermal diffusivity, \(Q_\alpha\) the alpha-particle heating source, and \(Q_{\text{rad}}\) the volumetric radiation loss.

For thermal transport, we adopt a Spitzer--H\"arm scaling for electronic heat conduction \cite{Spitzer1962,Harm1956}:
\begin{equation}
  \chi(T,\rho) = \chi_0 \frac{T^{5/2}}{\rho},
\end{equation}
with \(\chi_0=0.12\) chosen to approximate magnetized DT plasma diffusivity near 5--10~keV and \(\rho \approx 1~\mathrm{g\,cm^{-3}}\). (The choice of magnetized values is debatable, and not conservative, but we will let it slide for our experiment.)

The fusion power density follows from the Bosch--Hale \cite{BoschHale1992} fit for DT reactivity \(\langle \sigma v\rangle\):
\begin{equation}
  P_{\text{fus}} = n_D n_T \langle \sigma v\rangle E_{\text{fus}},
  \qquad
  E_{\text{fus}} = 17.6~\mathrm{MeV},
\end{equation}
where \(n_D = n_T = \rho/(2m_i)\). The alpha-particle heating source is
\begin{equation}
  Q_\alpha = f_\alpha P_{\text{fus}},
\end{equation}
with \(f_\alpha = E_\alpha/E_{\text{fus}} = 3.5/17.6 \approx 0.199\).

We model finite-range alpha transport using a nonlocal Helmholtz operator \cite{Atzeni2004}:
\begin{equation}
  \left( I - \lambda_\alpha^2\nabla^2 \right) q_\alpha = f_{\text{dep}} P_\alpha,
\end{equation}
where \(\lambda_\alpha(T,\rho) = R_m(T)/\rho\) and \(R_m(T)\) is the alpha mass range (in g\,cm\(^{-2}\)). The physically realistic table adopted was
\begin{equation}
  R_m(T) = 0.30\left(\frac{T}{3~\mathrm{keV}}\right)^{1/2}~\mathrm{g\,cm^{-2}},
\end{equation}
consistent with 3.5~MeV alpha stopping data. Deposition was limited to \(90\%\) of total energy (\(f_{\text{dep}}\le 0.9\)) to mimic escape.

Finally, for radiation losses, we use an approximate gray diffusion with an escape probability \cite{Mihalas1984,Braginskii1965}:
\begin{equation}
  Q_{\text{rad}} = C_{\text{rad}}\,\rho^2\sqrt{T}\,
  \frac{1}{1+\kappa_0\rho R_{\text{out}}},
\end{equation}
where \(C_{\text{rad}}=3\times10^{-36}~\mathrm{W\,m^3\,kg^{-2}\,keV^{-1/2}}\) and \(\kappa_0=5\times10^{-3}\) approximate Rosseland mean opacities for DT.

GPT-5 is quite good at this kind of model development and setup. It required little intervention on my part to make this plausible and complete. That would change in the numerics.

\subsection{Numerical Implementation}

GPT-5, without explicit prompting, selected for a partial differential equation (PDE) solving a second-order Crank--Nicolson implicit scheme on a uniform radial grid (\(N\approx150\) points). Time integration proceeded with adaptive substeps up to 10~ns. Boundary conditions:
\begin{equation}
  \frac{\partial T}{\partial r}=0 \quad \text{at } r=0, \qquad
  \frac{\partial T}{\partial r}=0 \quad \text{at } r=R_{\text{outer}}.
\end{equation}
Reaction, diffusion, and radiative source terms were split via operator splitting. Alpha deposition was solved by a tridiagonal Helmholtz inversion at each step.

These are reasonable choices for the kind of testbed we seek to develop. It should be noted that this took a few minutes to code up. While I could have done this, I do not do it every day, and I suspect it would have taken me a few work days to really get solid code working. The point is that I can now deliver in minutes as if I were at the highest level I have ever been at for this kind of work. There are those who will say this implementation is not remarkable, and it is not. But, to have it in minutes is remarkable.

It was at this point that my human labor and intelligence were required. I needed to suggest some parameterization for the density and temperature profiles at the fusion boundary. GPT-5 also needed me to help it find the region of parameter space that exhibited the kinds of threshold behaviors we were looking for. Make the hot spot too strong, and the whole thing singularly explodes. Make it too weak, and no burn wave will initiate and propagate. To understand the effect of the profile we are imagining, we need a delicately balanced marginal system. Given the nonlinearities in ICF, this takes not only intuition, but some numerical empiricism. GPT-5 made me work here.

For the density profile, three regions were specified:
\begin{itemize}
  \item Region 1 (hot spot): \(r \le R_{hs}\), \(\rho = \rho_{hs} \approx 1.2~\mathrm{g\,cm^{-3}}\).
  \item Region 2 (ramp): \(R_{hs} < r \le R_{hs}+L_{\text{ramp}}\),
  \begin{equation}
    \rho(r) = \rho_{hs} + s(r-R_{hs}) + c(r-R_{hs})^2,
  \end{equation}
  scaled to ensure \(\rho(R_{hs}+L_{\text{ramp}})=\rho_{\text{cold}}\).
  \item Region 3 (cold fuel): \(\rho = \rho_{\text{cold}} \approx 150~\mathrm{g\,cm^{-3}}\).
\end{itemize}

The ramp is controlled by the slope \(s\) and curvature \(c\), spanning
\[
s \in [1.5,6.5]\times10^9~\mathrm{kg\,m^{-4}}
\quad\text{and}\quad
c \in [-6.5,0]\times10^{13}~\mathrm{kg\,m^{-5}}.
\]

I spent a few hours of my time adjusting hot spot conditions, cold fuel conditions, conductivities, alpha particle stopping powers, and profile designs to get a physically reasonable result. This activity is what we in ICF might call ``design work'' -- getting to the place where the physics are interesting and useful. GPT-5 was a bad designer, offering results that were null, noisy, or invalid (NaNs), while claiming that glory had been achieved. However, when re-prompted to examine a pathological result or null signal, GPT-5 offered quite sophisticated solutions, including different implementations of FFTs to prevent aliasing, improved resolutions to track burn fronts, altered performance metrics to amplify signals. If a user gave up at the first wacky response, they would miss out on some rather powerful capability of the model. It needs to be highlighted that this requires you to be a very confident physicist. You have to know that an assertion is wrong, you have to confidently push for a better solution, and you have to be good enough to know when a real solution has been reached. I often find that GPT-5 offers solutions that I have to think deeply about to determine if they are actually correct, or just convenient.

We settled on simulations that were performed at a fixed hot-spot temperature \(T_{hs}=1.5~\mathrm{keV}\). For each (slope, curvature) pair, the mean front speed of the 5~keV isotherm between 1--3~ns was measured by least-squares fitting:
\begin{equation}
  v_{\text{mean}} =
  \frac{\displaystyle\int_{t_0}^{t_1} \frac{dr_f}{dt}\,dt}{t_1-t_0}.
\end{equation}

The resulting \(v_{\text{mean}}(s,c)\) field was smoothed with a Gaussian kernel (10\% bandwidth) to remove interpolation noise. Contours of 40--100~km\,s\(^{-1}\) clearly revealed a propagation ridge centered around
\begin{equation}
  s^{*} \approx 3.0\times10^9~\mathrm{kg\,m^{-4}}, \qquad
  c^{*} \approx -4.5\times10^{13}~\mathrm{kg\,m^{-5}},
\end{equation}
with local maxima up to \(v_{\text{mean}}\approx 100~\mathrm{km\,s^{-1}}\).

\begin{figure}[ht]
  \centering
  \includegraphics[width=\textwidth]{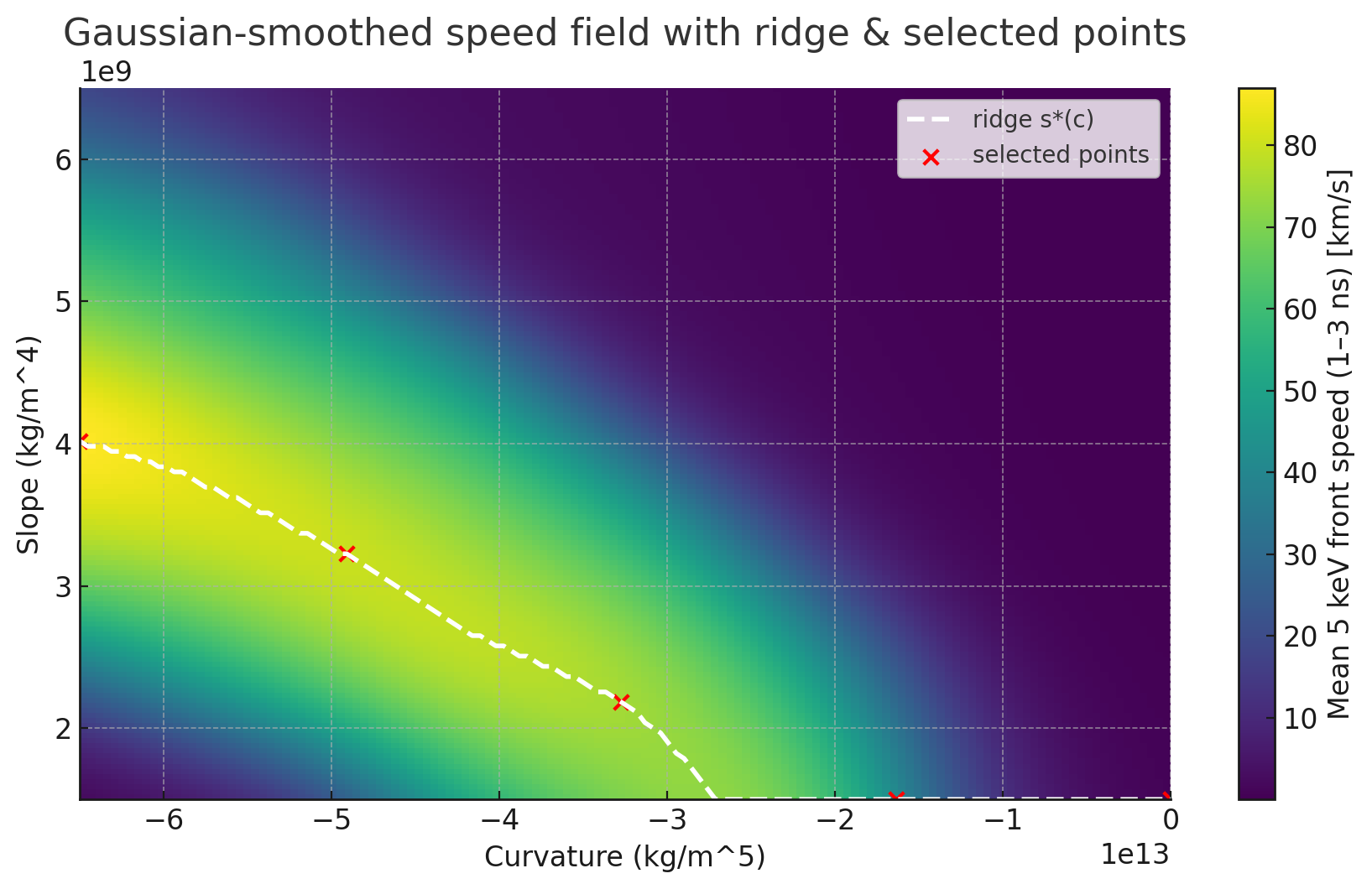}
  \caption{The color map represents the mean propagation speed of a thermonuclear burn wave propagating into the cold ICF fuel.  That wave speed depends on the profile of the interface between hot and cold DT fuel.  More specifically, both the average slope and the curvature play a key role.  Together they set an optimum profile for delivering the highest speed wave under the given physics assumptions.  These optimal profiles lie along the yellow ridge.}
  \label{fig:figure2}
\end{figure}

We spent some time examining the density profiles along the ridge. Five equally spaced points along this ridge were selected; their parameters are listed in Table~\ref{tab:ridge-points}.

\begin{table}[h]
  \centering
  \begin{tabular}{ccc}
    \hline
    \# & slope \((\mathrm{kg/m^4})\) & curvature \((\mathrm{kg/m^5})\) \\
    \hline
    1 & \(4.02 \times 10^{9}\) & \(-6.5 \times 10^{13}\) \\
    2 & \(3.23 \times 10^{9}\) & \(-4.91 \times 10^{13}\) \\
    3 & \(2.18 \times 10^{9}\) & \(-3.27 \times 10^{13}\) \\
    4 & \(1.50 \times 10^{9}\) & \(-1.64 \times 10^{13}\) \\
    5 & \(1.50 \times 10^{9}\) & \(0\) \\
    \hline
  \end{tabular}
  \caption{Five equally spaced points along the propagation ridge in \((s,c)\) space.}
  \label{tab:ridge-points}
\end{table}

Figure~\ref{fig:figure2} illustrates that as curvature becomes less negative, the density near the hot-spot boundary rises faster -- stiffening the thermal coupling and slightly reducing the front speed. Optimal profiles exhibit a gentle initial density rise, allowing alphas to preheat the ramp before losses dominate deeper in the cold fuel.

Within a few hours of noodling, we had a decent numerical picture of a relationship between the parameters chosen to describe the interface and their optimal values for propagating rapid burn fronts.

\subsection{Theoretical Explanation}

While the numerics show a trend, a good design physicist will aim to explain that trend using the theoretical tools at his disposal. I asked GPT-5 to help do this here for two reasons. First, it is useful to have this theoretical picture for forming the intuition we are looking for. Second, it provides an alternative check that GPT-5 has not hallucinated the physics and that I have not been fooled. This is a much-needed verification process.

GPT-5 quite quickly settled (following my nudging prompt) on a traveling-wave power balance to understand what sets the ridge.

We settled on a quasi--steady burn front, written in a co-moving coordinate \(\xi=r-vt\). Writing \(\rho=\rho(\xi)\) from the imposed ramp, the energy equation reads
\begin{equation}
  -v\,\rho C_v\,\frac{dT}{d\xi}
  = \frac{d}{d\xi}\left(\rho\,\chi(T,\rho)\,\frac{dT}{d\xi}\right)
  + Q_\alpha\bigl[T,\rho;\rho(\cdot)\bigr]
  - Q_{\mathrm{rad}}\bigl(T,\rho,\rho R_{\rm out}\bigr).
\end{equation}
It identified, I believe correctly, key scalings:
\begin{itemize}
  \item Thermal transport: \(\chi \sim \chi_0 T^{5/2}/\rho\). Low \(\rho\) implies strong preheat (large \(\chi\)), but also dilute heat capacity (smaller \(\rho C_v\)).
  \item Fusion source: \(Q_\alpha \sim f_\alpha E_{\rm fus}(n_D n_T)\langle\sigma v\rangle(T)\,\mathcal{D}[\rho]\). With \(n\propto\rho\), this gives \(Q_\alpha \propto \rho^2 \langle\sigma v\rangle(T)\) times a nonlocal deposition factor \(\mathcal{D}[\rho]\).
  \item Radiation losses: \(Q_{\rm rad}\sim C_{\rm rad}\rho^2 \sqrt{T}/(1+\kappa_0 \rho R_{\rm out})\). Higher \(\rho\) amplifies emissivity but also increases optical depth and reduces escape.
\end{itemize}

The ridge (maximal mean front speed) consists of \((s,c)\) pairs for which the net power into the preheat zone -- from conduction plus alpha deposition -- most efficiently overcomes radiative loss and the enthalpy needed to lift cold fuel to the 5~keV ignition threshold. In other words, where
\begin{equation}
\underbrace{
  \begin{array}{l}
    \text{(conduction from hot side)}\,+\\
    \text{(nonlocal alpha deposition)}
  \end{array}
}_{\text{effective drive}}
\;\;\;\text{is maximized against}\;\;\;
\underbrace{
  \begin{array}{l}
    \text{(radiative loss)}\,+\\
    \text{(downstream conduction)}\,+\\
    \rho C_v v \Delta T
  \end{array}
}_{\text{sinks + enthalpy}}
\end{equation}

It argued for an optimal slope \(s\). Write the ramp (endpoint-normalized) as
\begin{equation}
  \rho(\xi)\approx \rho_{\rm hs} + s\,\Delta + c\,\Delta^2,\quad
  \Delta\equiv (\xi-\xi_{\rm hs})\in[0,L_{\rm ramp}].
\end{equation}

Make \(s\) smaller (shallower): the near-front \(\rho\) stays lower for longer distances.
\begin{itemize}
  \item Pros: larger \(\chi\propto 1/\rho\) implies better conductive reach; longer alpha stopping length \(\lambda_\alpha\sim R_m/\rho\) implies alphas preheat ahead of the front.
  \item Cons: \(Q_\alpha\propto \rho^2\) is weaker at the front; also the hot preheat layer is optically thinner, so radiative escape improves.
\end{itemize}

Make \(s\) larger (steeper): the front sees higher \(\rho\) sooner.
\begin{itemize}
  \item Pros: stronger \(Q_\alpha\propto \rho^2\), larger heat capacity \(\rho C_v\) to ``catch and hold'' heat, greater \(\rho R\) so alpha energy deposits more locally (bigger \(f_{\rm dep}\)).
  \item Cons: \(\chi\) falls as \(1/\rho\), so preheat is poorer; emissivity (\(\propto\rho^2\)) rises, giving a larger radiative sink (even if escape probability shrinks).
\end{itemize}

Net: there is a sweet spot in \(s\) where conductive/alpha reach is still good, but \(\rho\) at the front is just high enough to bootstrap self-heating without being overly lossy.

It then moved on to optimal curvature \(c\) and why ``moderately negative'' values win.

Curvature redistributes where along the ramp the density rises:
\begin{itemize}
  \item Negative curvature (\(c<0\)) produces a gentle initial rise near the hot-spot edge and a steeper rise later. This creates a ``preheat shelf'': a low-\(\rho\), optically thinner, high-\(\chi\) region right where alpha deposition from the hot spot wants to land (a few \(\lambda_\alpha\) out). It boosts nonlocal alpha coupling and conductive lift to 5~keV in the ramp before losses take over.
  \item Positive curvature (\(c>0\)) produces a steep early rise that quickly increases \(\rho\) just outside the hot spot. That chokes preheat (low \(\chi\)), increases emissivity, and shortens \(\lambda_\alpha\) so more alpha energy deposits behind the front rather than ahead of it -- good for peak \(T\) behind, bad for front advance.
\end{itemize}

This is why the data favor moderately negative curvature at nearly any slope that lies on the ridge: it positions the right \(\rho\) at the right distance so the alpha--conduction synergy is maximized where the front needs it.

Ultimately, a simple power-balance criterion predicts an ``ideal'' \(c\) for each \(s\). A helpful mental model is to focus on a control volume of thickness \(\delta\) centered a distance \(\Delta^\star\) ahead of the hot-spot edge, where the front is forming. Choose \(\Delta^\star\sim\) a few local alpha ranges and conductive penetration depths during 1--3~ns. Then require net positive power there to reach and hold \(T_{\rm ig}\approx 5~\mathrm{keV}\):
\begin{equation}
  \underbrace{Q_\alpha^{\rm in}(\Delta^\star)}_{\text{nonlocal from hot side}}
  +
  \underbrace{Q_{\rm cond}^{\rm in}(\Delta^\star)}_{\propto \chi\,\Delta T/\ell}
  \gtrsim
  \underbrace{Q_{\rm rad}(\rho(\Delta^\star),T)}_{\propto \rho(\Delta^\star)^2 \sqrt{T}/(1+\kappa_0\rho R_{\rm out})}
  +
  \underbrace{\rho(\Delta^\star) C_v \frac{T_{\rm ig}-T_0}{\tau}}_{\text{enthalpy over timescale }\tau}.
\end{equation}

Two closure relations connect \(\rho(\Delta^\star)\) to \((s,c)\):

\begin{itemize}
  \item Alpha reach: \(\lambda_\alpha(\Delta^\star)\sim R_m/\rho(\Delta^\star)\) should be commensurate with \(\Delta^\star\), otherwise alphas either overshoot (too low \(\rho\)) or deposit behind (too high \(\rho\)):
  \begin{equation}
    \rho(\Delta^\star)\,\Delta^\star \sim R_m \quad\text{(order unity in areal density)}.
  \end{equation}
  For a fixed \(s\), this picks out a \(\rho(\Delta^\star)\) and thus fixes the needed curvature \(c\) via
  \(\rho(\Delta^\star)=\rho_{\rm hs}+s\,\Delta^\star+c\,{\Delta^\star}^2\).
  \item Conduction match: the conductive preheat length in time \(\tau\) is \(\ell_c \sim \sqrt{\chi\,\tau}\sim\sqrt{(\chi_0 T^{5/2}/\rho)\tau}\). Efficient preheat wants \(\ell_c\sim\Delta^\star\), i.e.
  \begin{equation}
    \frac{\chi_0 T^{5/2}}{\rho(\Delta^\star)} \sim \frac{{\Delta^\star}^2}{\tau}.
  \end{equation}
  This again fixes \(\rho(\Delta^\star)\) (hence \(c\)) at a given \(s\) and \(\tau\) (1--3~ns in our metric).
\end{itemize}

Put together, for each slope \(s\) there is a best \(\rho(\Delta^\star)\) that simultaneously (i) matches alpha areal-range (\(\rho\Delta^\star \sim R_m\)) and (ii) conductive reach (\(\ell_c\sim\Delta^\star\)). Because \(c\) is the degree of freedom that reshapes \(\rho(\Delta)\) locally at \(\Delta^\star\) while keeping the ramp endpoint fixed, you get an ideal curvature \(c^\star(s)\). This \(c^\star(s)\) traces the observed ridge.

Intuition from that criterion:
\begin{itemize}
  \item If \(s\) is too shallow, \(\rho(\Delta)\) stays small; you need a less negative \(c\) (or even positive) to raise \(\rho\) at \(\Delta^\star\) enough to meet \(\rho\Delta^\star\sim R_m\).
  \item If \(s\) is too steep, \(\rho(\Delta)\) climbs quickly; you need more negative \(c\) to delay the rise so alphas/conduction can reach and heat ahead of the front before losses spike.
\end{itemize}

The ridge curvature is thus the locus where \(\rho(\Delta^\star)\) hits the Goldilocks window: big enough for \(Q_\alpha\) and confinement, small enough for reach and low radiative drain.

We also developed tests to be sure we were clear about how the ridge should shift with physics knobs.

This picture immediately predicts how the ridge moves if you change microphysics:
\begin{itemize}
  \item Longer alpha range \(R_m\): pushes the optimum toward shallower \(s\) and more negative \(c\) (to keep \(\rho\Delta^\star \sim R_m\)).
  \item Stronger radiation (larger \(C_{\rm rad}\) or lower opacity): favors shallower densities near the front, so more negative \(c\) at a given \(s\).
  \item Lower hot-spot temperature: reduces source, so the optimum shifts toward slightly larger \(s\) (more \(\rho\) where the front forms) but still prefers negative \(c\) to preserve reach.
\end{itemize}

Finally, we ended with a compact ``engineering rule.'' Pick a target preheat distance \(\Delta^\star\) tied to your timing window (e.g., 1--3~ns) and desired 5~keV crossing. Then design \((s,c)\) so that
\begin{equation}
  \rho(\Delta^\star)\,\Delta^\star \approx R_m \quad\text{and}\quad
  \frac{\chi_0 T_{\rm ig}^{5/2}}{\rho(\Delta^\star)} \approx \frac{{\Delta^\star}^2}{\tau}.
\end{equation}
This reduces to a single condition on \(\rho(\Delta^\star)\), hence a curvature \(c^\star\) at each slope \(s\). In practice you will land on shallow slopes with moderately negative curvature, exactly what the simulations highlighted.

\begin{figure}[ht]
  \centering
  \includegraphics[width=\textwidth]{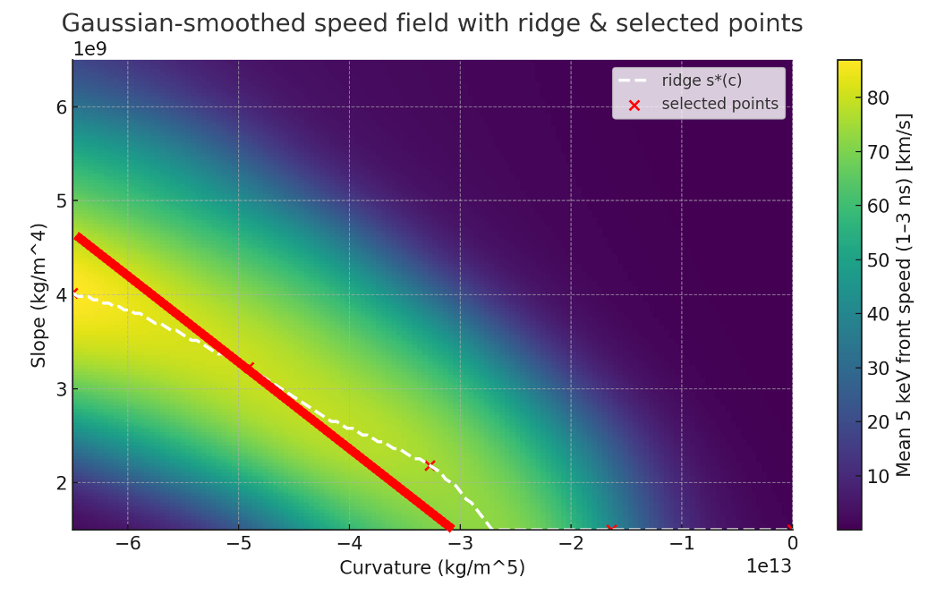}
  \caption{The red line represents the theoretical ridge slope developed by our theoretical exposition.  It fits reasonably well to the numerical ridge.  The combination of a numerical result verified by a theoretical power balance argument increases confidence in results.}
  \label{fig:figure2t}
\end{figure}

This formula offers a check on our results, using microphysics and power balance to estimate the expected \(c^\star(s)\) relationship. As shown in Figure \ref{fig:figure2t}, the red engineering rule and the numerically obtained ridge are pretty close. I invested about 6 hours in this project (it is quite interesting, but the exercise was to be quick), and I ended with both a numerical result and a theoretical explanation that give me confidence to proceed. As I mentioned earlier, this process itself is not novel. It is the workflow of a design physicist operating in the cutting-edge field of ICF. However, executing this workflow from concept, to numerical exploration, to theoretical supporting statement in hours is rather amazing.

\subsection{Physics Conclusions and Study Outcomes}

At the outset, I was looking for a set of physics goals, and we delivered:
\begin{itemize}
  \item A static, 1-D spherical reaction--diffusion model captures the essential burn-wave physics of ICF hot-spot ignition and propagation.
  \item Including realistic alpha stopping and nonlocal deposition with gray radiative loss yields credible propagation thresholds (marginal ignition near 1~keV).
  \item The ridge analysis identifies a family of fuel ramps -- shallow slope, moderately negative curvature -- that maximize front speed at fixed \(T_{hs}\).
  \item A theoretical analysis explains the ridge in density profile space from a fundamental power balance in a co-moving wave front.
  \item The resulting framework serves as a physics-informed testbed for developing theory, validating higher-fidelity hydrodynamic simulations, and designing future experiments.
\end{itemize}

This is a representative use of GPT-5 for my experience. The model took the seed of an idea that I had -- a static model to get a feel for key physics at the initiation and propagation of a thermonuclear burn wave -- and it helped me very quickly build a partial differential equation representation of the model. On command, it incorporated detailed stopping power and familiar (to the expert) physical data known to the ICF community. It discretized the problem, set up the optimization routine I imagined, and delivered the result I thought we would expect. This took me about 6 hours from start to finish -- idea to writing. There is not too much expertise that I could not find by querying my colleagues at the national labs, doing internet search, deriving some things by hand, recalling some undergraduate and graduate numerical tricks, and writing a few hundred lines of code. And, that is amazing. It made me a one-person army of experts all rolled up, as if I were the best I had ever been at everything I have ever thought about ICF. I can name the world experts it felt like I was talking to or the ones I would have had to reach out to to help me with some small step. But, they were right there in my laptop (more or less).

The theoretical explanation that I worked up with it is probably the best part. This explanation directly ties known physics relations together to give a closed-form predictor of what we had already found. This is a good way of ensuring that the results built over hours of elicitation with the model are not wrong and that the model had not seduced me into buying its arguments.

The results, I must say, do not come without work. The model, in its eagerness to please, often introduces numerical duct tape to smooth over a thorny issue, silently swaps out detailed numerical solves for approximations with trends it knows I want, and confidently declares victory when numerical signals are still obviously noise. When pointed to these errors and unwanted outcomes, the model is very good at actually fixing the issue. But, users must be expert enough to catch the oversimplifications, must persist to get the model to reconsider, and must be vigilant to ensure that the refined results are driving in the right direction.

Overall, as I have noted many times, I feel like my 6 hours of work here yielded something I could have done over a month or two with a very good pair of postdocs -- one for theory and one for numerics. So, say 6 person-months reduced to 6 person-hours. That is a compression of about a factor of 1000. I can only imagine what two postdocs and I could do if we collaborated with each of us working in this AI-accelerated mode.

\subsection{Part II --- Final Ridge Map Code}
Below is the complete Python code used to generate the Gaussian-smoothed ridge map and extract the five density profiles.

\lstset{
    backgroundcolor=\color{codebg},   
    commentstyle=\color{codegreen},
    keywordstyle=\color{magenta},
    numberstyle=\tiny\color{codegray},
    stringstyle=\color{codepurple},
    basicstyle=\ttfamily\footnotesize,
    breakatwhitespace=false,         
    breaklines=true,                 
    captionpos=b,                    
    keepspaces=true,                 
    numbers=left,                    
    numbersep=5pt,                  
    showspaces=false,                
    showstringspaces=false,
    showtabs=false,                  
    tabsize=2
}
\begin{lstlisting}[language=Python,caption={Ridge map generation and profile extraction.}]
import numpy as np
import matplotlib.pyplot as plt
from numpy.linalg import lstsq

# ------------------ Physical Constants ------------------
MeV_to_J = 1.602176634e-13
m_u = 1.66053906660e-27
m_ion = 2.5 * m_u
k_B = 1.380649e-23
K_per_keV = 1.1604518e7
Cv = 1.5 * k_B / m_ion
E_fus = 17.6 * MeV_to_J
E_alpha = 3.5 * MeV_to_J
alpha_fraction = E_alpha / E_fus

# ------------------ Regions & Profiles ------------------
class Regions:
    def __init__(self):
        self.R_hs = 50e-6
        self.L_ramp = 45e-6
        self.L_cold = 90e-6
        self.rho_hs = 1200.0
        self.rho_cold = 1.5e5
        self.T_cold_keV = 0.3

def make_regions(): return Regions()

def density_profile_norm(r, regs, slope, curvature):
    R1, R2 = regs.R_hs, regs.R_hs + regs.L_ramp
    rho = np.empty_like(r)
    rho[r <= R1] = regs.rho_hs
    mask = (r > R1) & (r <= R2)
    d = r[mask] - R1
    ramp = regs.rho_hs + slope*d + curvature*(d**2)
    end0 = regs.rho_hs + slope*(R2-R1) + curvature*(R2-R1)**2
    scale = (regs.rho_cold - regs.rho_hs)/(end0 - regs.rho_hs)
    ramp = regs.rho_hs + scale*(ramp - regs.rho_hs)
    rho[mask] = np.clip(ramp, regs.rho_hs, regs.rho_cold)
    rho[r > R2] = regs.rho_cold
    return rho

# ------------------ Transport / Sources ------------------
def thermal_diffusivity(T, rho, chi0=0.12):
    return chi0 * np.maximum(1e-6, T)**2.5 / np.maximum(1e-3, rho)

def bosch_hale_DT_cm3s(T_keV):
    C1 = 1.17302e-9
    C = [0.0, C1, 1.51361e-2, 7.51886e-2, 4.60643e-3, 1.35e-2,
         -1.0675e-4, 1.366e-5]
    B_G = 34.3827; mr_c2 = 1124656.0
    T = np.maximum(1e-6, T_keV)
    th = T/(1 - (T*(C[2]+T*(C[4]+T*C[6])))/(1+T*(C[3]+T*(C[5]+T*C[7]))))
    eta = (B_G**2/(4.0*th))**(1/3)
    return C[1]*th*np.sqrt(eta/(mr_c2*T**3))*np.exp(-3*eta)

def fusion_power_density(T, rho, burn):
    sv = bosch_hale_DT_cm3s(T)*1e-6
    n_i = rho/m_ion
    return 0.25*n_i**2*(1-burn)**2*sv*E_fus

def radiation_loss_gray(T, rho, rhoR, C_rad=3e-36, kappa0=5e-3):
    emiss = C_rad*rho**2*np.sqrt(np.maximum(1e-6, T))
    Pesc = 1/(1 + kappa0*np.maximum(0, rhoR))
    return emiss*Pesc

# realistic alpha mass range
def mass_range_Rm(T): return 0.30*(T/3)**0.5  # g/cm^2
def alpha_range_length(T, rho):
    return (mass_range_Rm(T)/np.maximum(1e-6, rho/1000))*1e-2

def rhoR_outward(r, rho):
    dr = r[1]-r[0]; rho_gcc = rho/1000
    return np.cumsum(rho_gcc[::-1])*dr*100.0[::-1]

# ------------------ Finite-Hotspot Solver ------------------
def step_CN_finite(T, B, r, rho, dt, chi0=0.12):
    N = len(T); dr = r[1]-r[0]
    chi = thermal_diffusivity(T, rho, chi0)
    chi_i = 0.5*(chi[:-1]+chi[1:])
    a=np.zeros(N-1);b=np.zeros(N);c=np.zeros(N-1)
    for i in range(1,N-1):
        Aip=((r[i+1]+r[i])/2)**2*chi_i[i]/dr
        Aim=((r[i]+r[i-1])/2)**2*chi_i[i-1]/dr
        vol=r[i]**2; a[i-1]=Aim/vol; c[i]=Aip/vol; b[i]=-(Aim+Aip)/vol
    b[0]=-2*chi_i[0]/dr**2; b[-1]=-2*chi_i[-1]/dr**2
    rhs = T + dt*((fusion_power_density(T,rho,B)*alpha_fraction
                  - radiation_loss_gray(T,rho,0))/ (Cv*rho*K_per_keV))
    # simple implicit diffusion
    M=np.zeros((N,N)); np.fill_diagonal(M,1-dt*b)
    for i in range(N-1): M[i,i+1]=-dt*c[i]; M[i+1,i]=-dt*a[i]
    Tnew=np.linalg.solve(M,rhs)
    return np.clip(Tnew,0,400),B

def run_sim_finite(regs,slope,curv,Ths,t_end=3.5e-9,N=150,Nt=140):
    Rtot=regs.R_hs+regs.L_ramp+regs.L_cold
    r=np.linspace(0,Rtot,N)
    rho=density_profile_norm(r,regs,slope,curv)
    T=np.full_like(r,regs.T_cold_keV); T[r<=regs.R_hs]=Ths
    B=np.zeros_like(T)
    times=np.linspace(0,t_end,Nt+1); dt=t_end/Nt
    Th=np.zeros((Nt+1,N)); Th[0]=T
    for n in range(1,Nt+1):
        T,B=step_CN_finite(T,B,r,rho,dt)
        Th[n]=T
    class Result: pass
    res=Result(); res.r=r; res.times=times; res.T_hist=Th; res.rho=rho; res.meta={'regs':regs}
    return res

# ------------------ Speed Metric ------------------
def interp_front(r,T,Tthr=5.0):
    idx=np.where(T>=Tthr)[0]
    if len(idx)==0: return np.nan
    i2=idx[-1]; i1=max(i2-1,0)
    T1,T2=T[i1],T[i2]; r1,r2=r[i1],r[i2]
    return r1+(Tthr-T1)/(T2-T1+1e-12)*(r2-r1)

def mean_speed(res,Tthr=5.0,t0=1,t1=3):
    t=res.times*1e9; r=res.r
    mask=(t>=t0)&(t<=t1)
    rf=[interp_front(r,res.T_hist[i],Tthr) for i in np.where(mask)[0]]
    rf=np.array(rf); ok=~np.isnan(rf)
    if ok.sum()<3:return 0
    A=np.vstack([t[mask][ok]*1e-9,np.ones(ok.sum())]).T
    v,_=lstsq(A,rf[ok],rcond=None)[0]
    return max(0,v)

# ------------------ Ridge Map Generation ------------------
regs=make_regions(); T_hs=1.5
slopes=np.linspace(1.5e9,6.5e9,11)
curvs=np.linspace(-6.5e13,0,11)
V=np.zeros((len(slopes),len(curvs)))
for i,s in enumerate(slopes):
    for j,c in enumerate(curvs):
        res=run_sim_finite(regs,s,c,T_hs)
        V[i,j]=mean_speed(res)

# Gaussian kernel smoother
def gauss_smooth(V,sigma=1):
    from scipy.ndimage import gaussian_filter
    return gaussian_filter(V,sigma=sigma,mode='nearest')

Vg=gauss_smooth(V,1)

# Ridge extraction
ridge_s=[]; ridge_c=[]
for j in range(Vg.shape[1]):
    i=np.argmax(Vg[:,j]); ridge_s.append(slopes[i]); ridge_c.append(curvs[j])
ridge_s=np.array(ridge_s); ridge_c=np.array(ridge_c)

# Ridge plot
plt.figure(figsize=(8,5))
im = plt.imshow(Vg.T, origin='lower',
                extent=[slopes.min(), slopes.max(), curvs.min(), curvs.max()],
                aspect='auto')
plt.colorbar(im, label='mean front speed (m/s)')
plt.plot(ridge_s, ridge_c, 'k--', linewidth=2, label='ridge')
plt.xlabel('slope (kg/m^4)'); plt.ylabel('curvature (kg/m^5)')
plt.legend(); plt.tight_layout()
plt.show()
\end{lstlisting}

%% file: content/09-SawhneySellke.tex
\section{Erd\H{o}s problems (part 2/2) -- Mehtaab Sawhney and Mark Sellke}
\label{sec:\secprefix-09}

We continue from where we left off in Section~\ref{sec:\secprefix-05} and explain how GPT-5 Pro assisted us in solving Erd\H{o}s Problem \#848.

\subsection{AI--assisted solution to Problem \#848}

Erd\H{o}s Problem \#848 is in combinatorial number theory, and was posted as Problem 23 in \cite{erdos1992favourite}.
This problem was genuinely open, and has now been solved by the first author and GPT-5, in combination with online comments of van Doorn, Weisenberg and Cambie.
The statement is:

\begin{center}
\fbox{%
\parbox{15cm}{Problem \#848: Let $A$ be a subset $\{1,2,\dots,N\}$ such that $ab + 1$ is not squarefree for all $a,b\in A$ (i.e. $ab+1$ must be divisble by $p^2$ for at least one prime $p$). Then for all sufficiently large $N$, no such set $A$ has more elements than $A = \{1\leq a\leq N~:~a\equiv 7 \pmod {25}\}$.}%
}
\end{center}

Of course, the given set $A$ satisfies the desired property because $7^2+1\equiv 0 \pmod{25}$.
It further turns out that this maximal $A$ is unique except when $\{1\leq a\leq N~:~a\equiv 18 \pmod {25}\}$ has the same size, and for some $\varepsilon>0$ any $A$ containing at least $\frac{(1-\varepsilon)N}{25}$ elements must be contained in one of these two.
A full solution of this problem is included in the appendix. 

This question is in a line of extremal questions raised by Erd\H{o}s at the intersection of elementary number theory and combinatorics. For instance, Erd\H{o}s and Silverman asked what is the densest subset of $[N]$ such that $A+A$ avoids a perfect square; this was solved by \cite{KLS02} who showed that $|A|\le (11/32 + o(1))N$. Erd\H{o}s similarly asked (also within Problem 23 of \cite{erdos1992favourite}) about the largest collection $A\subseteq [N]$ such that $ab$ is never square--free. 

The starting point of the solution is to observe that the diagonal constraint that $a^2+1$ not be squarefree already places a substantial restriction on $A$.
Indeed, classical number theory implies that only primes $p\equiv 1\pmod 4$ can divide $a^2+1$ twice, and for each such prime only $2/p^2$ fraction of integers $a$ have $a^2+1$ divisible by $p$.
Since divisibility by powers of different primes behaves approximately independently, this already implies the simple upper bound 
\[
1-\prod_{\substack{p \text{ prime},\\ p\equiv 1\text{(mod 4)}}}
\Big(1-\frac{2}{p^2}\Big)
\approx 0.105
\]
for the density of $A$, which is within a factor of $3$ of the truth.

What is more challenging is to use the off-diagonal constraints on $ab+1$ to non-trivially improve this upper bound. 
For this the first author asked GPT-5 Pro for suggestions, via the prompt displayed in Figure~\ref{fig:figure5.2} below (which itself built on online comments of van Doorn, Weisenberg and Cambie).

The idea suggested by GPT-5's reply in Figure~\ref{fig:figure5.2} gives a method to use any \emph{single} number $b\in A$ to obtain similarly harsh constraints on all other $a\in A$.
For example, let $A^{\ast}$ be the subset of $A$ consisting of numbers congruent to neither $7$ nor $18$ mod $25$.
If $A^{\ast}$ is non-empty, then fixing any $b\in A^{\ast}$, we have for any $a\in A$ that both $a^2+1$ or $ab+1$ must be divisible by $p^2$ for some prime $p$.
But since $b\in A^{\ast}$ it follows that only one of them can be divisible by $5$ itself.
This is a much stronger restriction on $a\in A$ and already yields the improved bound:
\begin{align*}
\Big(\frac{23}{25}\Big)&\Big(1 - \prod_{\substack{p\equiv 1\pmod 4\\p\ge 13}}\Big(1-\frac{2}{p^2}\Big)\Big) + \frac{2}{25}\Big(1 - \prod_{\substack{p\neq 5}}\Big(1-\frac{1}{p^2}\Big)\Big)\\
&\le .02517 + .0294 \approx .05457.
\end{align*}
Indeed for the $(23/25)$-fraction of integers not congruent to $7$ or $18$ mod $25$ we apply the diagonal argument above, and if $a\equiv 7,18\pmod{25}$ then $ab+1$ must be divisible by $p^2$ for $p\neq 5$.
This bound is already not far from the truth, which suggests that pushing it further ought to solve the entire problem.
GPT-5 made attempts in this direction but had numerous errors in its implementation (as can be seen in the transcript).
Still, the high-level idea was sound and was quickly turned into a complete proof.

\begin{figure}
\begin{chatbox}
\begin{userturn}
Consider the following problem. My suspicion is that there is a more effective way to run the inclusion--exclusion argument. The key point is that if a number satisfies both the congruence modulo $5$ and the congruence modulo $13$, then the standard upper bound will overcount, since such a number would also need an additional prime factor (distinct from $5$ and $13$). My intuition is that a stability-style analysis should help control this overcount, leading to a sharper bound, and potentially to a full proof. Please attempt to develop a proof along these lines.
\end{userturn}

\begin{assistantturn}
[Reasoned for 21m 22s]

Below I'll try to push the “necessary–condition + stability” approach you suggested far enough to reach
the natural extremal answer.

\bigskip

\textbf{Problem}

Let $A \subset \{1,\ldots,N\}$ and assume
\[
(\forall a,b\in A)\qquad ab+1\ \text{is \textbf{not} squarefree}. \tag{P}
\]

Erd\H{o}s asked whether the extremal sets are the residue class 
\[
A=\{n\le N : n\equiv 7\pmod{25}\}
\]
(or symmetrically $n\equiv 18\pmod{25}$), which have density $1/25$.

\textbf{1. The “one-variable necessary condition” (and the sharp constant)}
Setting $b=a$ in (P) shows that for every $a\in A$ there is a prime $p\equiv 1\pmod{4}$ with
$p^2 \mid a^2+1$. For such a prime $p$ there are exactly two residues $r \pmod{p^2}$ with
$r^2 \equiv -1$, so the set of integers $n$ with $p^2\mid n^2+1$ has density $2/p^2$. These events are independent across distinct primes (CRT), hence the exact density of integers $n$ for which $p^2\mid n^2+1$ for \emph{some} $p\equiv 1\pmod{4}$ is
\[
1-\prod_{p\equiv1\pmod4}\left(1-\frac{2}{p^2}\right)=0.10515\ldots
\]

Consequently
\[
|A|\ \le\ (0.10515+o(1))\,N.
\]

This is the sharp inclusion–exclusion bound behind the “0.108” in the notes.

This bound, however, is only a \emph{necessary} condition extracted from the diagonal pairs $(a,a)$. It ignores the pairwise demand in (P). The key is to exploit that demand via a stability argument.
\end{assistantturn}
\end{chatbox}
\caption{The start of the interaction with GPT-5 Pro to derive the improved bound for Problem \#848. In the above, it has just summarized the problem and the argument on \url{https://www.erdosproblems.com}.
Immediately after this, GPT-5 put forward the new idea that led to our solution (see next page).
The full transcript can be found here: \cite{SawhneyGPTbis}.}
\label{fig:figure5.2}
\end{figure}

\begin{figure}
\begin{chatbox}
\begin{assistantturn}

\textbf{2. A uniform squarefree sieve in a fixed 25-class}

Fix $a\ge1$. Let $t\in\{0,\ldots,24\}$ with $t\not\equiv -a^{-1}\pmod{25}$ (so that $25\nmid ab+1$ for every $b\equiv t\pmod{25}$). Consider
\[
B_a(X)\ :=\ \#\{\, b\le X : b\equiv t\pmod{25},\ ab+1\ \text{is squarefree}\,\}.
\]
For any odd prime $p\ne5$ with $p\nmid a$ the congruence $p^2\mid ab+1$ cuts out exactly one residue class modulo $p^2$; when $p\mid a$ it cuts out no class. By the Chinese Remainder Theorem, for any finite set $S$ of such primes we have the exact local density
\[
\frac{\#\{\, b\le X : b\equiv t\pmod{25},\ p^2\nmid ab+1\ \forall p\in S\,\}}{X/25}
=\prod_{\substack{p\in S\\ p\ne5}}\left(1-\frac1{p^2}\right)+O\!\left(\frac1X\right).
\]

Letting $S$ grow and using convergence of $\prod_p (1-1/p^2)$, we obtain a uniform (in $a$) asymptotic:

\medskip

\textbf{Lemma 1 (squarefree density in a fixed 25-class).}

For any fixed $a$ and any $t\not\equiv -a^{-1}\pmod{25}$,
\[
\frac{B_a(X)}{X/25}
= \prod_{p\ne5}\left(1-\frac1{p^2}\right)+o(1)
\ \ge\ \prod_{p\ne5}\left(1-\frac1{p^2}\right)+o(1),
\]
as $X\to\infty$. The rightmost constant is
\[
\delta := \prod_{p\ne5}\left(1-\frac1{p^2}\right)
= \frac{1}{\zeta(2)} \cdot \frac{1}{1-1/25}
= \frac{25}{4\pi^2}
= 0.633257\ldots
\]

Thus, for every fixed $a$ and every 25-class $t$ different from the unique class that makes
$25\mid ab+1$, at least a $\delta-o(1)$ fraction of $b$ in that class give $ab+1$ squarefree.
Equivalently, at most a fraction
\[
1-\delta \;=\; 1-\frac{25}{4\pi^2}\;=\; 0.366742\ldots
\]
of those $b$ can be “rescued’’ by primes other than 5.

This one fact drives the whole stability argument.

\end{assistantturn}
\end{chatbox}
\caption{A continuation of the interaction with GPT-5 Pro to derive the improved bound. Here it proposes the main new idea that led to the solution presented in Subsection~\ref{subsec:full-sol-848}.
Namely if $S$ contains a single ``unexpected'' number, this can be used to strongly constrain \emph{all} other numbers in $S$.
Note that the roles of $a$ and $b$ are swapped compared to Subsection~\ref{subsec:full-sol-848}.
The full transcript can be found here: \cite{SawhneyGPTbis}.
}
\end{figure}

A first step of the eventual solution was to note that if $b\in A^{\ast}$ is moreover even, then $ab+1$ cannot be divisible by $4$ and so the bound above improves to 
\begin{align*}
\Big(\frac{23}{25}\Big)&\Big(1 - \prod_{\substack{p\equiv 1\pmod 4\\p\ge 13}}(1-\frac{2}{p^2})\Big) + \frac{2}{25}\Big(1 - \prod_{\substack{p\notin \{2,5\}}}(1-\frac{1}{p^2})\Big)\\
&\le .02517 + .0125 \approx .03767.
\end{align*}
Since $.03767<1/25$ is strictly smaller than the density of numbers $7$ mod $25$, this already implies that a near-maximal set $A$ cannot contain any even numbers in $A^{\ast}$ at all.
Further casework in this vein allows one to rule out the existence of any $b\in A^{\ast}$ whatsoever, and finally to show that mixing between $7$ and $18$ mod $25$ is inefficient as well.
The main discussion with GPT-5 can be found here: \cite{SawhneyGPTbis}.

We emphasize that the first observation mentioned is due to Wouter van Doorn and the particular prompt given to GPT-5 was based on a comment of Stijn Cambie; these were already present in the comments of the official website \url{https://www.erdosproblems.com/forum/thread/848}. 
Thus, the final proof consists of a key solution step due to GPT-5 sandwiched between two layers of human mathematics!

\subsection{Looking Forward}
Our investigation of Erd\H{o}s problem \#848 highlights that GPT-5 has the ability to serve as an effective mathematical assistant, capable of recalling relevant lemmas, identifying analogies and locating relevant results from vague, ill-specified prompts. However, current models remain limited in perceiving the ``negative space'' of mathematics. While models are able to suggest plausible proof strategies, they often do not realize certain ``obvious'' examples which block progress, and are overly confident in the power of existing methods. This is unsurprising as such discussion is largely absent in mathematical literature itself; mathematicians don't systematically record \emph{why} problems are out reach, or \emph{why} a more obvious strategy doesn't work, or \emph{why} certain techniques are inherently unable to solve certain classes of problems.

\subsection{Formal solution of Problem \#848}
\label{subsec:full-sol-848}

\begin{proposition}\label{prop:main}
There exists an integer $N_0$ such that if $N\ge N_0$ then the following holds. Let $A\subseteq [N]$ with the property that if $a,b\in A$ then $ab+1$ is not squarefree. Then 
\[
|A|\le |\{n\in [N]:n\equiv 7 \pmod {25}\}|,
\]
i.e. $\{n\in [N]:n\equiv 7 \pmod {25}\}$ achieves the maximum possible value of $|A|$.
\end{proposition}
\begin{remark}
Our proof actually allows one to go further and characterize all \emph{approximately} maximal sets $A$. Namely equality is achieved precisely when $A=\{n\in [N]:n\equiv 7 \pmod {25}\}$ or $A=\{n\in [N]:n\equiv 18 \pmod {25}\}$ (the latter may be optimal or suboptimal by $1$ depending on the value of $n$ modulo $25$). Further it turns out that for all $|A|\ge 0.038 \cdot N$ within $5\%$ of optimality, when $N$ is sufficiently large, $A$ must be contained in either $\{n\in [N]:n\equiv 7 \pmod {25}\}$ or $\{n\in [N]:n\equiv 18 \pmod {25}\}$ (and of course all such $A$ satisfy the squarefree condition).
\end{remark}

We first require the following pair of preliminary lemmas; the first was already used by van Doorn (see \cite{BloWeb}). In order to be self-contained, we provide proofs with suboptimal error terms. 
\begin{lemma}\label{lem:sieve-1}
Let $\mathcal{P}$ be a subset of primes with $\max \mathcal{P}\le N^{1/2}$ and let $q$ be a positive integer. For each $p\in \mathcal{P}$, define $\mathcal{R}_p$ to denote a set of residue classes modulo $p^2$ with $|\mathcal{R}_p|\le 2$ and $\mathcal{R}_p = \emptyset$ if $(p,q)\neq 1$. Then 
\begin{align*}
&\Bigg|\Big|\Big(\big\{n\in [N]:n \equiv t\mod q\big\} \bigcap \bigcup_{p\in \mathcal{P}}\{n \pmod {p^2} \in \mathcal{R}_p\}\Big)\Big| - \frac{N}{q} \Big(1-\prod_{p\in \mathcal{P}}\Big(1 - \frac{|\mathcal{R}_p|}{p^2}\Big)\Big)\Bigg| 
\\ 
&\leq  O(N(\log N)^{-1/2}).
\end{align*}
\end{lemma}
\begin{proof}
Let $T = \lfloor \sqrt{\log N}\rfloor$. Note that 
\[\Big|[N]\bigcap \bigcup_{\substack{p\in \mathcal{P}\\T\le p\le N^{1/2}}}\{n \pmod {p^2} \in \mathcal{R}_p\}\Big|\leq O\Big(\sum_{T\le p\le N^{1/2}}\frac{N}{p^2}\Big)\leq O\Big(\frac{N}{T}\Big).\]
Therefore it suffices to estimate 
\[\Big|\Big(\big\{n\in [N]:n \equiv t\mod q\big\} \bigcap \bigcup_{\substack{p\in \mathcal{P}\\p\le T}}\{n \pmod {p^2} \in \mathcal{R}_p\}\Big)\Big|.\]
Note that as $T\le \sqrt{\log N}$, we have that $\prod_{p\le T}p^2\le N^{o(1)}$. Via inclusion exclusion, we have 
\begin{align*}
&\Big|\Big(\big\{n\in [N]:n \equiv t\mod q\big\} \bigcap \bigcup_{\substack{p\in \mathcal{P}\\p\le T}}\{n \pmod {p^2} \in \mathcal{R}_p\}\Big)\Big| \\
&=\sum_{\substack{S\subseteq \mathcal{P}\cap [T]\\S\neq \emptyset}}(-1)^{|S|-1} \cdot \Big|\bigcap_{p\in S}\{n\pmod {p^2}\in \mathcal{R}_p\} \cap \{n\in [N]:n\equiv t\pmod q\}\Big|\\
&=N^{o(1)}+\sum_{\substack{S\subseteq \mathcal{P}\cap [T]\\S\neq \emptyset}}(-1)^{|S|-1} \cdot \frac{N}{q}\cdot\frac{\prod_{p\in S}|\mathcal{R}_p|}{\prod_{p\in S}p^2}\\
&= \frac{N}{q} \cdot \Big(1 - \prod_{p\in \mathcal{P}\cap [T]}\Big(1 - \frac{|\mathcal{R}_p|}{p^2}\Big)\Big) + N^{o(1)}.
\end{align*}
Completing the product, we have the desired result. 
\end{proof}

We next require a variant of the previous lemma which allows us to incorporate the ``off-diagonal'' constraints that $ab+1$ is not squarefree.
\begin{lemma}\label{lem:sf-AP}
Fix a positive integer $N$, residue class $t\pmod q$ with $q$ a perfect square and an integer $b$ such that $1\le b\le N$. Suppose that there does not exist a prime $p$ such that $p^2|q$ and $p^2|(bt + 1)$.

Then we have that 
\[\bigg|\Big|\{a\in [N]: a\equiv t\pmod q \wedge \mu(ab + 1) =0\}\Big| - \frac{N}{q} \cdot \Big(1 - \prod_{\substack{p\\(p,qb) = 1}}\Big(1 - \frac{1}{p^2}\Big)\Big)\bigg| \leq O(\frac{N}{\sqrt{\log N}}).\]
\end{lemma}
\begin{proof}
Note that $ab + 1 \le N^2 + 1$. Therefore if $\mu(ab + 1) = 0$, there exists a prime $p$ such that $p^2|(ab + 1)$. Additionally note that $(p,qb) = 1$. In particular, if $p|b$ then $p\nmid ab + 1$. Furthermore if $p|q$, then observe $ab + 1 \equiv bt + 1\not\equiv 0 \pmod {p^2}$ by the imposed condition. 

We now let $T = \lfloor \sqrt{\log N}\rfloor$. We have that 
\[
\sum_{T\le p\le N}|\{a\in [N]: p^2|(ab + 1)\}|\leq O\Big(\sum_{T\le p\le N}\Big(\frac{N}{p^2} + 1\Big)\Big)\leq O\Big(\frac{N}{T}\Big).
\]
Therefore it suffices to handle $p\le T$ with $(p,qb) = 1$. The condition that $p^2|ab + 1$ gives one specified residue class (depending on each prime $p$) and therefore we may handle the remaining primes via Lemma~\ref{lem:sieve-1} and we may conclude the proof.
\end{proof}

We now give the remainder of the analysis to conclude the proof. The proof consists of breaking the set into the parts which are $7\pmod {25}$, those that are $18 \pmod {25}$, and all the rest. We then apply combinations of Lemma~\ref{lem:sieve-1} and Lemma~\ref{lem:sf-AP} based on whether the relevant sets are nonempty. To optimize the numerical factor of $\prod_{(p,qb) = 1}(1-1/p^2)$, we introduce casework based on whether these sets contain an even integer or consist solely of odd numbers. We note that there exist other similar solutions within this framework; indeed our initial version of the solution required more extensive casework modulo numbers including $169$ and $289$. The conversation with GPT-5 \cite{SawhneyGPTbis} in particular uses the auxiliary prime $3$ and therefore the precise computation differs from the draft here. 

\begin{proof}[{Proof of Proposition~\ref{prop:main}}]
Let $\eta > 0$ be a small absolute constant (in fact $\eta=0.002$ suffices), and let $N$ be sufficiently large and suppose that $|A|\ge (1/25 - \eta) \cdot N$. Define
\begin{align*}
A_{7} &= \{a\in A: a\equiv 7 \pmod {25}\}\\
A_{18} &= \{a\in A: a\equiv 18 \pmod {25}\}\\
A^{\ast} &= A\setminus (A_{7} \cup A_{18}).
\end{align*}

We first prove that $A^{\ast}$ is empty. 

Suppose that there exists $b\in A^{\ast}$ with $2|b$. Then for each $x\in A^{\ast}$, there is $p\equiv 1\pmod 4$ with $p\ge 13$ such that $p^2|x^2 + 1$. By breaking $A^{\ast}$ into residue classes modulo $25$ (of which there are $23$ possibilities) and applying Lemma~\ref{lem:sieve-1}, we have that 
\[\frac{|A^{\ast}|}{N}\le \Big(\frac{23}{25}\Big)\Big(1 - \prod_{\substack{p\equiv 1\pmod 4\\p\ge 13}}\Big(1-\frac{2}{p^2}\Big)\Big) + o(1).\]
For $a\in A_7\cup A_{18}$, there exists $p\neq 2, 5$ such that $p^2|ab + 1$. This implies by Lemma~\ref{lem:sf-AP} that 
\[\frac{|A_7 \cup A_{18}|}{N}\le \Big(\frac{2}{25}\Big)\Big(1 - \prod_{\substack{p\neq 2,5}}\Big(1-\frac{1}{p^2}\Big)\Big) + o(1).\]
Combining these bounds we have 
\begin{align*}
\frac{|A|}{N} &\le \Big(\frac{23}{25}\Big)\Big(1 - \prod_{\substack{p\equiv 1\pmod 4\\p\ge 13}}\Big(1-\frac{2}{p^2}\Big)\Big) + \frac{2}{25}\Big(1 - \prod_{\substack{p\neq 2, 5}}\Big(1-\frac{1}{p^2}\Big)\Big)+o(1)\\
&\qquad\qquad\qquad\qquad\le 0.0252 + 0.0125 = 0.0377< 0.04 -\eta;
\end{align*}
here we have absorbed the $o(1)$ error assuming that $N$ is sufficiently large. 
This contradicts our assumption that $|A|\geq (1/25-\eta)N$.

We have now reduced to the case that $A^{\ast}$ consists only of odd elements. By breaking $A^{\ast}$ into residue classes modulo $50$ and noting there are $23$ valid possibilities and applying 
Lemma~\ref{lem:sieve-1}, we have that 
\[\frac{|A^{\ast}|}{N}\le \Big(\frac{1}{2}\Big)\Big(\frac{23}{25}\Big)\Big(1 - \prod_{\substack{p\equiv 1\pmod 4\\p\ge 13}}\Big(1-\frac{2}{p^2}\Big)\Big) + o(1).\]
We now split further into cases based on whether or not $A_7 \cup A_{18}$ contains even elements. First suppose that no element of $A_7 \cup A_{18}$ is even; then $A_7 \subseteq \{7, 57\} \pmod {100}$ and $A_{18}\subseteq \{43, 93\} \pmod {100}$. Fixing $b\in A^{\ast}$, for each of $A_7$ and $A_{18}$ one of the residue classes $\pmod {100}$ will have $2^{2}$ as a divisor and the other will not. Therefore applying Lemma~\ref{lem:sf-AP} to these progressions $\pmod {100}$, we have 
\[\frac{|A_7 \cup A_{18}|}{N}\le \frac{1}{50} + \Big(\frac{1}{50}\Big)\Big(1 - \prod_{\substack{p\neq 2,5 }}\Big(1-\frac{1}{p^2}\Big)\Big) + o(1).\]
This gives another contradiction as 
\begin{align*}
\frac{|A|}{N} &\le \Big(\frac{1}{2}\Big)\Big(\frac{23}{25}\Big)\Big(1 - \prod_{\substack{p\equiv 1\pmod 4\\p\ge 13}}\Big(1-\frac{2}{p^2}\Big)\Big) + \frac{1}{50} + \Big(\frac{1}{50}\Big)\Big(1 - \prod_{\substack{p\neq 2,5 }}\Big(1-\frac{1}{p^2}\Big)\Big)+o(1)\\
&\qquad\qquad\qquad\qquad\le 0.0126 + 0.0200 + 0.0032 = 0.0358 < 0.04 -\eta.
\end{align*}
Therefore we have reduced to the case where $A^{\ast}$ is nonempty, consisting of odd elements and at least one of $A_{7}$ or $A_{18}$ has an even element. We now fix $b\in A^{\ast}$ and $b'\in A_{7}$ with $2|b'$ (with the case where $A_{18}$ has the even element being completely analogous). The size of $|A^{\ast}|$ is bounded as before. By using $b$ and Lemma~\ref{lem:sf-AP}, we have that 
\[\frac{|A_7|}{N}\le  \frac{1}{25}\Big(1 - \prod_{\substack{p\neq 5}}\Big(1-\frac{1}{p^2}\Big)\Big) + o(1).\]
Additionally, using $b'\in A_7$ and Lemma~\ref{lem:sf-AP}, we have that 
\[\frac{|A_{18}|}{N}\le  \frac{1}{25}\Big(1 - \prod_{\substack{p\neq 2,5}}\Big(1-\frac{1}{p^2}\Big)\Big)+o(1).\]
Therefore we have that 
\begin{align*}
\frac{|A|}{N} &\le \Big(\frac{1}{2}\Big)\Big(\frac{23}{25}\Big)\Big(1 - \prod_{\substack{p\equiv 1\pmod 4\\p\ge 13}}\Big(1-\frac{2}{p^2}\Big)\Big) + \frac{1}{25}\Big(1 - \prod_{\substack{p\neq 5}}\Big(1-\frac{1}{p^2}\Big)\Big) +\frac{1}{25}\Big(1 - \prod_{\substack{p\neq 2,5}}\Big(1-\frac{1}{p^2}\Big)\Big) +o(1)\\
&\qquad\qquad\qquad\qquad\le 0.0126 + 0.0147 + 0.0063 = 0.0336 < 0.04 -\eta.
\end{align*}

Therefore we have now reduced to the case where $A^{\ast} = \emptyset$. Suppose that $A_{7}$ and $A_{18}$ are nonempty; fixing $b\in A_7$ and $b'\in A_{18}$ and applying Lemma~\ref{lem:sf-AP} twice we have that 
\begin{align*}
\frac{|A|}{N} &\le \frac{2}{25}\Big(1 - \prod_{\substack{p\neq 5}}\Big(1-\frac{1}{p^2}\Big)\Big)+o(1)\le 0.0294 < 0.04 -\eta.
\end{align*}
Therefore $A$ is fully contained inside one of $A_{7}$ or $A_{18}$. This immediately gives the desired bound. 
\end{proof}

We record various numerical constants to 4 digits, rounding to preserve the stated inequalities.

\begin{itemize}
\item $\prod_p\Bigl(1-\frac{1}{p^2}\Bigr)=\frac{6}{\pi^2} \ge 0.6079$, so $1-\frac{6}{\pi^2} \le 0.3921$.
\item $\prod\limits_{p\ne 5}\Bigl(1-\frac{1}{p^2}\Bigr)=\frac{25}{4\pi^2} \ge 0.6332$, so $1-\frac{25}{4\pi^2} \le 0.3668$.
\item $\prod\limits_{p\ne 2,5}\Bigl(1-\frac{1}{p^2}\Bigr)=\frac{25}{3\pi^2} \ge 0.8443$, so $1-\frac{25}{3\pi^2}\le 0.1557$.
\item $1-\prod\limits_{\substack{p\equiv 1\,(4)\\ p\ge 13}}\Bigl(1-\frac{2}{p^2}\Bigr)\le 0.0274$.
\end{itemize}

%% file: content/10-Coester.tex
\section{New online algorithms lower bounds -- Christian Coester}
\label{sec:\secprefix-10}

This section describes how GPT-5 assisted in proving new bounds on the performance of algorithms for a fundamental problem in online optimization called \emph{convex body chasing}. The problem was introduced by~\cite{FriedmanL93} and captures how to make sequential decisions under uncertainty when each step restricts feasible options in a geometric way. Formally, one is given a sequence of convex sets $K_1, K_2, \ldots, K_T\subset \mathbb{R}^d$, revealed one by one. Upon seeing $K_t$, an algorithm must choose a point $p_t \in K_t$ before learning the next set. The goal is to minimize the total movement $\sum_{t=1}^T \|p_t - p_{t-1}\|$. The algorithm is called \emph{online} as it has to make decisions on the fly, without knowledge of the future. In contrast, an \emph{offline} algorithm is one that knows the entire input upfront, allowing it to move optimally. The standard measure of performance of an online algorithm is its \emph{competitive ratio}, defined as the worst-case (i.e., maximum) ratio between the online algorithm's cost and the optimum (offline) cost on any instance.

Intuitively, convex body chasing formalizes the tension between adaptivity and stability: each new convex body represents a changing feasible region (e.g., updated constraints, new data, or evolving system state), and the online algorithm must adjust with minimal movement. It arises naturally in contexts like control theory (tracking feasible states), machine learning (online convex optimization with movement costs), and operations research (dynamic resource allocation). In theoretical computer science, it serves as a clean abstraction for understanding how well an online decision process can perform relative to an omniscient optimum, with close connections to other central problems in online optimization \cite{BansalC22}.

When introducing the problem, \cite{FriedmanL93} showed that the competitive ratio of any online algorithm is at least $\sqrt{d}$ in $d$-dimensional Euclidean space. Following a sequence of breakthroughs in recent years (\cite{BubeckLLS19,ArgueGTG21,Sellke20}), it is now established that the optimal competitive ratio lies in the interval $[\sqrt{d},d]$. An important special case that received significant attention is chasing \emph{nested} convex bodies, where $K_1\supset K_2\supset\dots \supset K_T$ (\cite{BansalBEKU20,ArgueBCGL19,BubeckKLLS20}).

The following sections show how GPT-5 refuted competitiveness of a natural algorithm in response to a single prompt, and how it assisted in raising the lower bound on the competitive ratio of arbitrary online algorithms from $\sqrt d$ to $\frac{\pi}{2}\sqrt{\lfloor \frac{d}{2}\rfloor}\approx 1.11\sqrt{d}$.

\subsection{Lower bound for Follow-the-Leader}\label{sec:cc-FTL}
When the initial location is $p_0=0$, observe that for \emph{nested} convex body chasing, the optimal \emph{offline} algorithm will move to $\arg\min_{p\in K_T}\|p\|$, which is feasible for all time steps and minimizes cost. So a natural online algorithm is to move to $p_t = \arg\min_{p\in K_t}\|p\|$ at each time $t$, i.e., tracking the currently optimal offline location. I did not necessarily expect this algorithm to perform well as this type of follow-the-leader algorithm is known to have poor competitive ratio for related problems, and the reason is typically trivial. However, I could not see any such easy argument to refute the algorithm, and what made me curious is that it does achieve the optimal competitive ratio of $\pi/2$ on the specific hard instance described in Section~\ref{sec:cc-generalLB} that yields a $\pi/2$ lower bound for arbitrary algorithms in the two-dimensional case. Given a single short prompt, GPT-5 produced a rather non-obvious counter-example against the algorithm. It shows that the cost of the algorithm can be made arbitrarily large even in the two-dimensional nested case when the initial body is the unit ball around the start location (and thus proves that the competitive ratio is infinite, since the optimal offline cost is at most 1 here):

\clearpage

\begin{chatbox}
  \begin{userturn}
    Consider nested convex body chasing in 2-dimensional Euclidean space, where the initial location is 0 and the initial convex body is the unit ball centered at 0. Prove or disprove that the cost of the follow-the-leader algorithm (i.e., always go to the point in the current convex body whose norm is minimal) is at most pi/2.
  \end{userturn}
  \begin{assistantturn}
        [Reasoned for 16m 19s]

        \textbf{Short answer:} the statement is \textbf{false}.

        [...]

        \subsubsection*{1) Two basic facts about the greedy minimizer}
        Let \(K_0 \supseteq K_1 \supseteq K_2 \supseteq \cdots \subset \mathbb{R}^2\) be a nested sequence of nonempty closed convex sets, with \(K_0\) the closed unit disk centered at \(0\).  
        
        Let \(p_t\in K_t\) be the unique point minimizing \(\|x\|\) over \(K_t\) (uniqueness follows from strict convexity of the Euclidean norm and convexity of $K_t$).
        
        Write \(p_t=r_t u_t\) with \(r_t=\|p_t\|\in[0,1]\) and \(u_t\in\mathbb{S}^1\).
        
        Two standard properties hold:
        
        \begin{enumerate}
            \item \textbf{Half‑space support}: for every \(t\),
           \[
             K_t \subseteq H(p_t):=\{x:\langle p_t,x\rangle \ge r_t^2\}.
           \]
           Equivalently, for all \(s\ge t\),
           \[
             \langle p_t,p_s\rangle \;\ge\; r_t^2 
             \quad\Longleftrightarrow\quad 
             \cos(\angle(u_t,u_s)) \;\ge\; \frac{r_t}{r_s}.
             \tag{1}
           \]
           (This is just the normal cone condition at a Euclidean projection.)
        
            \item \textbf{Realizability}: conversely, if a sequence \((p_t)\) with nondecreasing radii \(r_t\) satisfies (1) for all \(s\ge t\), then setting
           \[
             K_t := \overline{B(0,1)}\cap \bigcap_{i\le t} H(p_i)
           \]
           yields a nested sequence for which the unique minimum‑norm point in \(K_t\) is exactly \(p_t\).
           (Indeed, \(p_t\in K_t\) by (1), and \(H(p_t)\) excludes every point of norm \(<r_t\); convexity then forces uniqueness.)
        \end{enumerate}
        Thus, to disprove any universal upper bound on the greedy cost, it suffices to \textbf{exhibit a sequence} \((p_t)\subset \overline{B(0,1)}\) satisfying (1) whose polygonal length
        \(\sum_t \|p_{t+1}-p_t\|\) exceeds that bound.
    \end{assistantturn}
\end{chatbox}

\clearpage
\begin{chatbox}
    \begin{assistantturn}
        \subsubsection*{2) An alternating construction}
        
        Pick a nonincreasing sequence of angles \(\{\varphi_m\}_{m\ge 1}\) with
        \[
        0<\varphi_{m+1}\le \varphi_m < \frac{\pi}{8},\qquad 
        \sum_{m=1}^\infty \varphi_m = \infty,\qquad
        \sum_{m=1}^\infty \varphi_m^2 < \infty.
        \]
        (For instance, \(\varphi_m=\frac{c}{m}\) with a small fixed \(c\in(0,\pi/8)\) works.)
        
        Choose a small \(r_1>0\). Define a sequence of points by alternating signs and increasing radii:
        \[
        \begin{aligned}
        p_{2m-1} &= r_{2m-1}\,(\cos(-\varphi_m),\sin(-\varphi_m)),\\
        p_{2m}   &= r_{2m}\,(\cos(\ \varphi_m),\sin(\ \varphi_m)),
        \end{aligned}
        \]
        with radii chosen \textbf{recursively} so that the immediately preceding constraint in (1) is tight:
        \[
        \boxed{
        \begin{aligned}
        r_{2m}   &= \frac{r_{2m-1}}{\cos(\varphi_m+\varphi_m)} 
                 = \frac{r_{2m-1}}{\cos(2\varphi_m)},\\[2mm]
        r_{2m+1} &= \frac{r_{2m}}{\cos(\varphi_m+\varphi_{m+1})}.
        \end{aligned}}
        \tag{2}
        \]
        (Geometrically, each new point lies on the boundary of the supporting half‑space defined by the previous minimizer, with the largest possible angular change allowed by that one constraint.)

        [...]
    \end{assistantturn}
\end{chatbox}

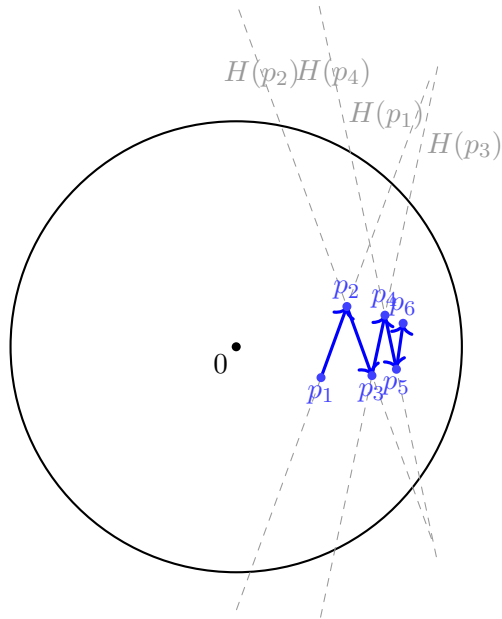
\begin{figure}[h]
\centering
\begin{tikzpicture}[scale=3, line cap=round]
  % Unit circle
  \draw[thick] (0,0) circle (1);

  % Origin
  \fill (0,0) circle (0.02) node[below left=-1pt] {$0$};

  % Define angles (in degrees) and radii (precomputed from the recurrence)
  % Angles: theta1=-20, theta2=+20, theta3=-12, theta4=+12, theta5=-8, theta6=+8
  \def\aone{-20}
  \def\atwo{20}
  \def\athree{-12}
  \def\afour{12}
  \def\afive{-8}
  \def\asix{8}

  % Radii (approx.): r1=0.400, r2=0.522, r3=0.615, r4=0.674, r5=0.717, r6=0.747
  \coordinate (p1) at ({0.400*cos(\aone)},{0.400*sin(\aone)});
  \coordinate (p2) at ({0.522*cos(\atwo)},{0.522*sin(\atwo)});
  \coordinate (p3) at ({0.615*cos(\athree)},{0.615*sin(\athree)});
  \coordinate (p4) at ({0.674*cos(\afour)},{0.674*sin(\afour)});
  \coordinate (p5) at ({0.717*cos(\afive)},{0.717*sin(\afive)});
  \coordinate (p6) at ({0.747*cos(\asix)},{0.747*sin(\asix)});

  % Supporting lines H(p_t): <p_t, x> = ||p_t||^2
  % Each is perpendicular to p_t and passes through p_t.
  \foreach \P/\A in {p1/\aone, p2/\atwo, p3/\athree, p4/\afour}{
    \draw[dashed,gray!80] ($(\P)+(\A+90:1.4)$) -- ($(\P)+(\A-90:1.1)$);
  }

  % Draw the greedy path
  \draw[very thick,blue,->] (p1) -- (p2);
  \draw[very thick,blue,->] (p2) -- (p3);
  \draw[very thick,blue,->] (p3) -- (p4);
  \draw[very thick,blue,->] (p4) -- (p5);
  \draw[very thick,blue,->] (p5) -- (p6);

  % Points
  \foreach \P/\N in {p1/{$p_1$}, p3/{$p_3$}, p5/{$p_5$}}{
    \fill[blue!75] (\P) circle (0.02) node[below=-1pt] {\N};
  }
  \foreach \P/\N in {p2/{$p_2$}, p4/{$p_4$}, p6/{$p_6$}}{
    \fill[blue!75] (\P) circle (0.02) node[above=-1pt] {\N};
  }

  % Labels for some halfspaces
  \node[gray!80] at ($ (p1) + (\aone+96:1.2) $) {$H(p_1)$};
  \node[gray!80] at ($ (p2) + (\atwo+90:1.1) $) {$H(p_2)$};
  \node[gray!80] at ($ (p3) + (\athree+80:1.1) $) {$H(p_3)$};
  \node[gray!80] at ($ (p4) + (\afour+90:1.1) $) {$H(p_4)$};

\end{tikzpicture}
\caption{Beginning of a hard instance for the follow-the-leader algorithm. Figure made by GPT-5.}\label{fig:cc-FTL-LB}
\end{figure}

Figure~\ref{fig:cc-FTL-LB} contains an illustration of the construction. GPT goes on to justify the feasibility of the construction: ``We must verify that (1) holds not only for the immediate predecessor but \textbf{for all} earlier indices.'' The argument it gives for this is actually incorrect, but a correct argument is easy to see: For $s\ge t+1$, we obtain (1) from the following chain of inequalities:
\[
\cos(\angle(u_{t},u_s)) \ge \cos(\angle(u_{t},u_{t+1})) = \frac{r_{t}}{r_{t+1}} \ge \frac{r_t}{r_s}.
\] 

GPT completes the proof (within the same response) as shown in the next chat box. The first line in it contains a slight inaccuracy, since we need the inequality not only for \(|x|\le \pi/8\) but for \(|x|\le \pi/4\). It holds even in that case though, so the overall proof is not affected.

\begin{chatbox}
    \begin{assistantturn}
        Finally, because \(\sum\varphi_m^2<\infty\) and \(1/\cos x\le \exp(x^2)\) for \(|x|\le \pi/8\), the product in (2) stays bounded:
        \[
        \sup_t r_t \;\le\; r_1 \cdot 
        \exp\!\Big(C\sum_{m\ge1}\varphi_m^2\Big) < \infty.
        \]
        Choosing \(r_1>0\) small enough we ensure \(r_t\le 1\) for all \(t\), so every \(p_t\) lies in the unit disk.
        
        \subsubsection*{3) The cost explodes}
        
        The distance traveled in the two types of steps is explicit, because in (2) the immediately previous constraint is tight:
        
        \begin{itemize}
            \item \textbf{Flip step} \(p_{2m-1}\to p_{2m}\) (angles \(-\varphi_m \to +\varphi_m\)):
          \[
          \|p_{2m}-p_{2m-1}\|
          =\sqrt{r_{2m}^2-r_{2m-1}^2}
          =r_{2m-1}\tan(2\varphi_m).
          \]
            \item \textbf{Cross step} \(p_{2m}\to p_{2m+1}\) (angles \(+\varphi_m \to -\varphi_{m+1}\)):
          \[
          \|p_{2m+1}-p_{2m}\|
          =\sqrt{r_{2m+1}^2-r_{2m}^2}
          =r_{2m}\tan(\varphi_m+\varphi_{m+1}).
          \]
        \end{itemize}
        
        Therefore, after \(N\) pairs the greedy cost satisfies
        \[
        \begin{aligned}
        L_N
        &=\sum_{m=1}^N
        \Big(r_{2m-1}\tan(2\varphi_m)+r_{2m}\tan(\varphi_m+\varphi_{m+1})\Big)\\
        &\ge \sum_{m=1}^N
        \Big(2r_{2m-1}\varphi_m+r_{2m}(\varphi_m+\varphi_{m+1})\Big)
        \quad\text{(since }\tan x\ge x\text{ for }x\in[0,\pi/2])\\
        &\ge c\sum_{m=1}^N\varphi_m
        \end{aligned}
        \]
        for some constant \(c>0\) depending only on \(r_1\) and \(\{\varphi_m\}\) (because the radii are nondecreasing [...]).  
        Since \(\sum_m \varphi_m=\infty\), we obtain \(L_N\to\infty\) as \(N\to\infty\). [...]
    \end{assistantturn}
\end{chatbox}
 Thus the ``follow‑the‑leader'' cost can be made arbitrarily large while all requests stay nested inside the unit disk. The original interaction is available at \cite{CoesterGPT}.

\subsection{Improved lower bounds for general algorithms}\label{sec:cc-generalLB}
We now discuss how GPT-5's assistance helped in proving a better lower bound on the competitive ratio of arbitrary online algorithms. The idea for this lower bound is based on a previous 1.5358 lower bound due to~\cite{BienkowskiBCCJK19} for chasing lines (i.e., where each $K_t$ is a line), which was the only previous known improvement over the aforementioned $\sqrt{d}$ (when $d=2$). The proof in \cite{BienkowskiBCCJK19} is rather ad-hoc, giving a small number of request lines issued in three steps, using six rational constants to specify those lines. It was acknowledged by \cite{BienkowskiBCCJK19} that 1.5358 does not seem to be tight, since a generalization of their idea, using more steps and request lines, should achieve a better bound. However, this would make their analysis even more involved.

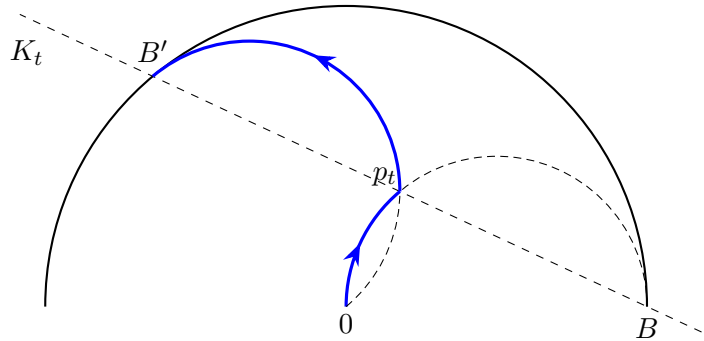
\begin{figure}
\centering
\begin{tikzpicture}[
  scale=4,
  >=Stealth,
  dot/.style={circle, fill, inner sep=0.8pt},
  algpath/.style={line width=1.2pt,
                  postaction={decorate},
                  decoration={markings, mark=at position 0.5 with {\arrow{Stealth}}},
                  blue}
]

% ---------- parameter: rotation of K_t (degrees) ----------
\def\phi{-25} % negative = clockwise from the x-axis

% ---------- base points ----------
\coordinate (O) at (0,0);
\coordinate (B) at (1,0);

% ---------- trigonometric helpers ----------
\pgfmathsetmacro{\cosa}{cos(\phi)}
\pgfmathsetmacro{\sina}{sin(\phi)}

% ---------- line K_t through B with direction (cos phi, sin phi) ----------
\coordinate (Kta) at ($(B) + (0.2*\cosa, 0.2*\sina)$);
\coordinate (Ktb) at ($(B) - (2.3*\cosa, 2.3*\sina)$);

% ---------- other intersection B' of K_t with unit circle ----------
% Solve (B + t u) on x^2 + y^2 = 1 -> t = -2 cos(phi), t=0 is B
\pgfmathsetmacro{\tBprime}{-2*\cosa}
\pgfmathsetmacro{\xBprime}{1 + \tBprime*\cosa}
\pgfmathsetmacro{\yBprime}{0 + \tBprime*\sina}
\coordinate (Bp) at (\xBprime,\yBprime);

% ---------- projection point p_t: midpoint of B and B' ----------
\coordinate (P) at ($0.5*(B)+0.5*(Bp)$);

% ---------- Thales circle centers (radius 1/2) ----------
\coordinate (C1) at (0.5,0);          % diameter OB
\coordinate (C2) at ($(O)!0.5!(Bp)$);  % diameter OB'

% Helpful angles (degrees):
%  - On the circle with center C1, P sits at angle 180 + 2*phi (w.r.t. C1)
%  - On the circle with center C2, B' sits at angle 180 + 2*phi, and P sits at angle 0
\pgfmathsetmacro{\endOne}{180 + 2*\phi}
\pgfmathsetmacro{\endTwo}{180 + 2*\phi}

% Angle from C2 to O (so we can draw the OB' semicircle from O to B')
\pgfmathsetmacro{\alpha}{atan2(-\yBprime/2,-\xBprime/2)} % equals angle of vector O - C2

% ================== DRAWING ==================

% Upper semicircle of the unit circle (center O)
\draw[line width=0.8pt] (1,0) arc (0:180:1);

% Request line K_t
\draw[dashed] (Kta) -- (Ktb) node[pos=0.95, below left] {$K_t$};

% First Thales semicircle: diameter OB (center (1/2,0), radius 1/2)
\draw[densely dashed] (C1) ++(0:0.5) arc (0:180:0.5);

% Algorithm path so far (bold): O -> p_t along the first semicircle
% Start at O (angle 180° at C1), end at angle 180 + 2*phi (p_t)
\draw[algpath] (C1) ++(180:0.5) arc (180:\endOne:0.5);

% Second Thales semicircle: diameter OB' (radius 1/2), drawn for reference
\draw[densely dashed] (C2) ++(\alpha:0.5) arc (\alpha:{\alpha+180}:0.5);

% "Switch-and-pivot" path (bold): p_t -> B' along the second semicircle
% At C2, p_t sits at angle 0°, B' at angle 180 + 2*phi
\draw[algpath] (C2) ++(0:0.5) arc (0:\endTwo:0.5);

% Points and labels
\draw[dot] (O)  node[below=1pt] {$0$};
\draw[dot] (B)  node[below=1pt] {$B$};
\draw[dot] (Bp) node[above=1pt] {$B'$};
\draw[dot] (P)  node[above left=1pt] {$p_t$};

\end{tikzpicture}
\caption{Illustration of $\pi/2$ lower bound idea. Figure made by GPT-5.}\label{fig:cc-pi2-LB}
\end{figure}

My goal was to derive a better and cleaner lower bound using a generalization of their approach where the request line moves continuously. My initial attempts to prompt GPT-5 to help with this construction were unsuccessful: It only reproved the known 1.5358 lower bound, failing to understand why my suggested continuous generalization would be helpful. Curiously though, the reasoning trace preview mentioned at one point that it was aiming for a lower bound of $\pi/2 \approx 1.5708$, a quantity I had not thought of before (the prompt of Section~\ref{sec:cc-FTL} took place at a later time). Although the reasoning trace preview did not substantiate why it would aim for this number specifically, and the responses did not mention it any more, it may have contributed to inspiring me towards the following, more concrete plan for a lower bound:

\paragraph{Lower bound idea (see Figure~\ref{fig:cc-pi2-LB}).} Consider an instance with initial location $(0,0)$, each request set is a line, initially the line $y=0$. Start rotating the line clockwise around $B=(1,0)$. Observe that the projection $p_t$ of the origin onto the current line $K_t$ is the north-western intersection of $K_t$ with the circle of radius $1/2$ around $(1/2,0)$. This is essentially due to Thales' theorem. If the online algorithm follows this curve until the request line is vertical, then it moves along the semicircle and its cost is exactly $\pi/2$ whereas the optimal cost is $1$. So an algorithm has to deviate from this curve in order to achieve a smaller competitive ratio. In fact, since we can force the algorithm to end at $B$, it has to go ``below'' the semicircle curve at some point, otherwise its path is even longer. Consider the first time $t$ when the online algorithm moves ``below'' the semicircle curve. Let $B’$ be the intersection other than $B$ between $K_t$ and the circle of radius 1 around the origin. Observe that $p_t$ is the midpoint of the line segment $\overline{BB'}$, and the minimal offline cost to reach any point $x\in \overline{BB'}$ is the norm $\|x\|$, whose value is symmetric around $p_t$ on $\overline{BB'}$. So we can switch and pivot around $B’$ instead of $B$, counter-clockwise. The semicircle curve from before has a corresponding semicircle curve on the other side of $p_t$. The fact the online algorithm went ``below'' the previous semicircle curve means that it is now above/outside the new semicircle curve, and getting back onto the curve only makes it more expensive for the online algorithm. Repeating this argument, the algorithm should eventually reach a point $p_t$ at distance $1$ from the origin after having moved distance $\ge \pi/2$.

\paragraph{The task for GPT-5.} A subtle difficulty in the preceding argument lies in defining the precise moment of pivot switching. If the switch occurs exactly when the algorithm crosses below the semicircle, then at that moment the algorithm is still located on the semicircle, at the location where it intersects the new semicircle. But then it might also try to go below the new semicircle immediately, leading to a possibly infinite sequence of pivot switches at the same instant, effectively freezing the process.

My initial plan for resolving this was to argue that the algorithm must move at a temporarily faster speed in order to go below the semicircle, allowing the pivot switch to be delayed until its speed decreases again; but formalizing this approach seemed technically cumbersome. GPT-5 suggested a much simpler and cleaner solution: trigger the switch once the algorithm is below the semicircle at distance $\ge \epsilon$ from $p_t$. %of radius $1/2 - \epsilon$ instead of $1/2$, for $\epsilon>0$. 
This avoids the freeze, and as $\epsilon\to 0$ we obtain the same lower bound of $\pi/2$. This adjustment allows to make the argument rigorous and complete. The idea seems obvious in hindsight, yet far more elegant than the overcomplicated solution I was initially attempting. Since the interaction took place using Oxford's ChatGPT Edu license, due to sharing restrictions of this license a link to the full interaction is unfortunately unavailable. An excerpt is shown in Figure~\ref{fig:cc-pi2-excerpt}.

\begin{figure}
\begin{chatbox}
    \begin{assistantturn}
        [Reasoned for 10m 38s]
        
        [...]
        
        \textbf{A robust fix: $\varepsilon$-hysteresis switching}

Pick any $\varepsilon \in (0,1)$. Run the adversary with the following rule.

\begin{itemize}
    \item \textbf{While pivoting around $P$} (clockwise), keep pivoting as long as
    \[
        u_P(\phi)= t(\phi) - \cos\phi \ge -\varepsilon.
    \]
    When the \textbf{first} time $\tau$ with $u_P(\tau) = -\varepsilon$ occurs, \textbf{switch} pivot to  
    \[
        Q := \text{the north-west intersection of } L_\tau \text{ with the unit circle},
    \]
    and reverse the rotation direction.

    \item \textbf{While pivoting around $Q$} (counter-clockwise), keep pivoting as long as
    \[
        u_Q(\psi) := s(\psi) - \cos\psi \ge -\varepsilon
    \]
    (here $s(\psi)$ is the signed distance from $Q$ along the current line). When the first time with $u_Q = -\varepsilon$ occurs, switch to the new unit-circle intersection on that line, and so on.

    \item \textbf{Stop} as soon as the current line becomes tangent at the current pivot (which gives $\min w = 1$).
\end{itemize}

\textbf{Key geometry at a switch}

At the switch time $\tau$ from $P$ to $Q$, the midpoint $M$ of $[P,Q]$ is exactly $p^{\star}(\tau)$ (the work-function minimizer on $L_\tau$). Since $u_P(\tau) = -\varepsilon$, the online point lies $\varepsilon$ \textbf{closer} to $P$ than $M$ along $L_\tau$. Measured from the other endpoint $Q$, that means it lies $\varepsilon$ \textbf{farther} than $M$:
\[
    u_Q(\tau) = +\varepsilon.
\]

So after every switch we begin \textbf{strictly outside} the new Thales circle, with a uniform gap $\varepsilon$. This removes the ``immediately afterwards'' ambiguity and prevents instantaneous back-switching.

[...]

    \end{assistantturn}
\end{chatbox}
\caption{An excerpt from GPT-5 Pro's response, in which it suggested the introduction of a small parameter $\varepsilon$ to avoid technicalities on when to switch in continuous time.
(Note that the notation used in this response differs from our explanations in this section.)
}\label{fig:cc-pi2-excerpt}
\end{figure}

\paragraph{Extension to higher dimension.} Finally, I was hoping to extend the lower bound to arbitrary dimension. My own approach (so far unsuccessful) was to first aim for an extension of the above idea to three dimensions, but it does not seem obvious how to do this. I posed the problem to GPT-5, with a prompt that included the lower bound idea above (without the $\epsilon$ details) plus the following sentence:

\begin{chatbox}
    \begin{userturn}
        [...] Try to generalize this idea for $d=2$ in order to improve over the known $\sqrt{d}$ lower bound for arbitrary dimension $d$.
    \end{userturn}
\end{chatbox}

Within its first response, GPT-5 proposed an approach yielding a lower bound of $\frac{\pi}{2} \sqrt{\lfloor \frac{d}{2}\rfloor} \approx 1.11 \sqrt{d}$. The idea is simple: Consider an orthogonal decomposition of $\R^d$ into $\lfloor d/2\rfloor$ two-dimensional components $\R^2$, and apply the $\pi/2$ lower bound one by one in the components. Although there were errors in GPT-5's proposed execution of this idea, it is not hard to make rigorous. The full chat is available at \cite{CoesterGPTbis} (containing also subsequent unsuccessful attempts to push the ratio further). Notably, my initial focus on three dimensions might have caused me to miss this simple decomposition idea.

\subsection{Reflection}

GPT-5 assisted the research process on this problem at several stages. Perhaps the most impressive part is its proof refuting the follow-the-leader algorithm, produced from a single prompt without any guidance on how to approach the task. This result saved significant time I might have otherwise spent unsuccessfully searching for an upper bound, and it is unclear if I would have been able to come up with the lower bound, which contained several non-trivial steps. The aid it provided for the lower bound on arbitrary algorithms was on the one hand the inspiring appearance of the number $\pi/2$ in its reasoning trace, and on the other hand a technical idea to resolve the ``freezing'' issue in the proof. While I would have presumably found \emph{some} solution to the freezing myself, it would have taken longer to formalize and likely been less elegant. GPT-5's extension to arbitrary dimension was simple, but beyond the scope of my approach of focusing on three dimensions first.

Naturally, several limitations are also still apparent: it initially failed to understand how viewing the problem in continuous time could yield stronger bounds, and upon later attempts to push the lower bounds further, it presented arguments containing serious flaws. In the proofs presented in this section, GPT-5's responses also contained some errors, but these were easy to fix for a human, overall accelerating the research process.

%% file: content/11-BubeckSellkeYin.tex
\section{Inequalities on subgraph counts in trees  -- Sébastien Bubeck, Mark Sellke, and Steven Yin}
\label{sec:\secprefix-11}

We conduct here another experiment, this time in graph theory, on a problem formulated and studied by the first author of this section, together with Nati Linial, in a 2013 preprint that was later published in 2016 in the Journal of Graph Theory (\cite{bubeck2016local}). The paper lists 7 open problems, and the one we consider here is the first open problem, arguably the one we were most interested in solving back in 2013:
\begin{itemize}
\item The open problem is about a conjectured infinite set of inequalities for subgraph counts in trees, that if true would determine a certain convex set of interest.
\item The follow-up paper \cite{bubeck2016paths} shows that, except possibly for the first two, none of these inequalities are actually true. Moreover the main result of \cite{bubeck2016paths} is a proof that the first inequality is indeed true. The second inequality is left as an open problem.
\item Aided by somewhat sophisticated scaffolding (aimed at doing mathematics research), GPT-5 was able to reprove the first inequality, and then build on this to also prove the second (open) inequality. Both of GPT-5's proofs are quite different from any of the arguments in \cite{bubeck2016local, bubeck2016paths}.
We present the re-proof of the first inequality in Section \ref{sec:one} and the new proof of the second in Section \ref{sec:two}.
\item Both proofs below are entirely AI-generated, aside from minor editing for clarity. No human input was needed beyond first asking to prove the first inequality, and then (with the solution to the first inequality in context) asking to prove the second inequality. (A few incorrect proofs were also generated and rejected by human checking.) See Figure \ref{fig:trees_1} and Figure \ref{fig:trees_2} for the precise prompts.
\item We moreover note that GPT-5 did not have access to web search in this experiment, and did not seem to be aware of the existing proof in \cite{bubeck2016paths} (indeed the latter paper has remained quite niche, with only 3 citations in 10 years). 
The pre-existing (human-generated) proof rests on a somewhat cumbersome analysis, that analyzes several different cases across 4 pages of calculation. In contrast GPT-5's proof is short and elegant, and based on a somewhat miraculous identity. The entire argument in Section \ref{sec:one} fits in a page and a half. 
\end{itemize}

We note that we first attempted to use this research scaffolding on top of GPT-5 to solve a few Erd\H{o}s problems, but with no success. The experiment described here was our first attempt at a problem outside of Erd\H{o}s' problems.

\subsection{Paths, stars and wyes in trees}
\begin{figure}
\centering
\begin{tikzpicture}[
  every node/.style={circle,draw,fill=black,inner sep=1.8pt},
  vertex/.style={circle,draw,fill=black,inner sep=1.8pt},
  labelstyle/.style={font=\small,anchor=north}
]

% --- S: 5-vertex star ---
\begin{scope}[xshift=-2cm]
  % nodes
  \node[vertex] (cS) {};
  \node[vertex] (s1) [above left=10mm of cS] {};
  \node[vertex] (s2) [above right=10mm of cS] {};
  \node[vertex] (s3) [below left=10mm of cS] {};
  \node[vertex] (s4) [below right=10mm of cS] {};
  % edges
  \draw (cS) -- (s1);
  \draw (cS) -- (s2);
  \draw (cS) -- (s3);
  \draw (cS) -- (s4);
\end{scope}

% --- P: 5-vertex path ---
\begin{scope}[xshift=0cm]
  \node[vertex] (p1) {};
  \node[vertex] (p2) [right=9mm of p1] {};
  \node[vertex] (p3) [right=9mm of p2] {};
  \node[vertex] (p4) [right=9mm of p3] {};
  \node[vertex] (p5) [right=9mm of p4] {};
  \draw (p1) -- (p2) -- (p3) -- (p4) -- (p5);
\end{scope}

% --- Y: degree-3 center with one branch of length 2 ---
\begin{scope}[xshift=6cm]
  \node[vertex] (cY) {};
  \node[vertex] (y1) [above left=10mm of cY] {};
  \node[vertex] (y2) [above right=10mm of cY] {};
  \node[vertex] (y3) [right=14mm of cY] {};
  \node[vertex] (y3leaf) [right=9mm of y3] {};
  \draw (cY) -- (y1);
  \draw (cY) -- (y2);
  \draw (cY) -- (y3) -- (y3leaf); % branch of length 2
\end{scope}

\end{tikzpicture}
\caption{The three isomorphism types of trees with 5 vertices. TikZ figure produced by GPT-5.}
\label{fig:tikztrees}
\end{figure}
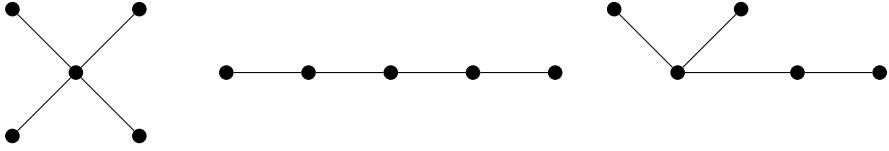

Let $G$ be a finite tree.
We consider the following subgraph counts (see also Figure \ref{fig:tikztrees}):
\begin{itemize}
    \item $S$ is the number of induced star subgraphs on $5$ vertices within $G$.
    \item $P$ is the number of induced path subgraphs on $5$ vertices within $G$.
    \item $Y$ is the number of induced $5$-vertex subtrees of the remaining isomorphism type within $G$ (i.e. $1$ center vertex which is the endpoint of paths of lengths $1,1,2$).
\end{itemize}

\begin{theorem}[\cite{bubeck2016paths}]
\label{thm:main-bsy}
For any finite tree $G$, one has
\begin{equation}
Y-9S-P\leq 6.
\end{equation}
\end{theorem}

\begin{conjecture}[\cite{bubeck2016local, bubeck2016paths}]
\label{conj:main}
There exists some constant $K>0$ such that for any finite tree $G$, one has
\begin{equation}
29Y-42P-144S\leq K.
\end{equation}
\end{conjecture}

The motivation for the above problems is that the set of achievable asymptotic proportions $\Big(\frac{Y}{Y+S+P},\frac{S}{Y+S+P},\frac{P}{Y+S+P}\Big)$ in trees as $Y+S+P\to\infty$ can be shown to be convex (in fact for general subtree counts); thus one would like to understand the linear constraints, and Theorem~\ref{thm:main-bsy} appeared to be a tight example.
By contrast in general (dense) graphs, it is known that the infinite-dimensional \emph{flag algebra} semidefinite program of \cite{razborov2007flag} generates all such inequalities, but that linear inequalities of the form above can still be computationally undecidable \cite{hatami2011undecidability}.
\newline

In some sense Theorem \ref{thm:main-bsy} gives the ``first face" of this convex set, while Conjecture \ref{conj:main} gives the ``second face". We note that Conjecture \ref{conj:main} was also made in \cite{bubeck2016local}, together with an infinite set of conjectured inequalities, one for each face, and those conjectured inequalities were all disproven in \cite{bubeck2016paths} except for Conjecture \ref{conj:main} which we will show here to be actually true.
\newline

\emph{Aside from being checked and rewritten for clarity, the new proofs below are entirely AI-generated.}

\begin{figure}
\begin{chatbox}
  \begin{userturn}
Let $G$ be a finite tree graph. Let $S$ be the number of induced 5-vertex star subtrees, $P$ the number of induced 5-vertex path subtrees, and $Y$ the number of induced 5-vertex subtrees of the remaining isomorphism type (one degree 3 central vertex, and 1 of its 3 neighbors has a 2nd neighbor). Prove that $Y \leq 9*S + P + O(1)$.
  \end{userturn}
\end{chatbox}
\caption{Question posed to (a scaffolded version of) GPT-5 to reprove Theorem \ref{thm:main-bsy}. The answer it provided is written in Section \ref{sec:one}, with minor edits for clarity.}
   \label{fig:trees_1}
\end{figure}

\subsection{New proof of Theorem~\ref{thm:main-bsy}} \label{sec:one}
    For $v\in G$ let $d_v$ be the degree of $v$ (number of neighbors) and:
    \[
    x_v=d_v-1,\quad\quad 
    T_v=\sum_{u\sim v} x_u,
    \quad\quad 
    A_v=A_{d_v}=\binom{d_v-1}{2}
    .
    \]
    It is not hard to see that:
    $$        S=\sum_{v\in G}\binom{d_v}{4}, \quad\quad
        P=\frac{1}{2}
        \sum_{v\in G}\Big[T_v^2 - \sum_{u\sim v}x_u^2\Big], \quad\quad
        Y=\sum_{v\in G}A_{v} T_v.$$
    Indeed the value of $S$ is clear. 
    For $P$, each summand counts the number of paths centered at $v$; note that $T_v^2$ counts the number of pairs of possibly identical length $2$ paths with endpoint at $v$, and $\sum_{u\sim v}x_u^2$ counts the number of such pairs which overlap (at $u\sim v$). 
    Thus $T_v^2 - \sum_{u\sim v}x_u^2$ counts the number of ordered pairs of such paths which do not overlap, so dividing by $2$ gives the number of paths centered at $P$.
    For $Y$, each summand again counts the number of subgraphs centered at $v$. Again $T_v$ counts the number of length $2$ paths with endpoint at $v$; given any such path, to make a $Y$ it suffices to choose $2$ of the remaining $d_v-1$ neighbors of $v$, for a total of $A_{d_v}$ options.

    The first key claim is the identity
    \begin{equation}
    \label{eq:identity-1}
    Y-9S-P
    =
    \sum_{v\in G}
    \Big[F(d_v) -\frac{(T_v-A_v)^2}{2}\Big]
    \end{equation}
    where $F$ is the quartic polynomial:
    \[
    F(k)
    =
    \frac{\binom{k-1}{2}^2}{2} + \frac{k(k-1)^2}{2} - 9\binom{k}{4}
    .
    \]
    To prove \eqref{eq:identity-1}, firstly note that $9\binom{d_v}{4}=9S_v$ and the terms $T_v^2/2$ and $A_v T_v$ directly match up.
    The remaining identity to prove is:
    \[
    \frac{1}{2}\sum_{v}\sum_{u\sim v}x_u^2 
    \stackrel{?}{=}
    \frac{\binom{x_v}{2}^2}{2} + \frac{d_v x_v^2}{2}
    -\frac{A_v^2}{2}.
    \]
    In fact, by definition $\binom{x_v}{2}^2 = A_v^2$, and by changing the order of summation one finds:
    \[
    \sum_{v}\sum_{u\sim v}x_u^2=
    \sum_{u}\sum_{v\sim u}x_u^2=
    \sum_u d_u x_u^2 = \sum_v d_v x_v^2.
    \]
    This establishes \eqref{eq:identity-1}.

    Next, we have the miraculous identity for each $v$:
    \begin{equation}
    \label{eq:miracle}
    -\frac{(T_v-A_v)^2}{2} + F(d_v)
    =
    \frac{5}{2}\sum_{u\sim v} (d_v-d_u)
    -3(d_v-2)
    -
    \frac{d_v(d_v-1)(d_v-3)(d_v-4)}{4}
    -
    \binom{T_v-A_v-2}{2}.
    \end{equation}
    Given \eqref{eq:miracle} the rest of the proof is immediate.
    Indeed the first term on the right-hand side telescopes when summing over $v\in G$ (each term $(d_v-d_u)$ is counted once with sign $1$ and once with sign $-1$), and the third and fourth terms are never positive.\footnote{Note that non-positivity of the third and fourth terms rely respectively on $d$ and $(T_v-A_v)$ being integers. The third in particular appears to be an inspired choice that would be difficult to find for a computer algebra system, though it is perhaps natural given the equality cases (which contain vertices only of degree $1,3,4$).}
    Thus \eqref{eq:miracle} implies that 
    \[
    Y-9S-P
    \leq 
    6|V(G)|-3\sum_{v\in G} d_v
    = 6
    \]
    where the latter equality holds for all trees. 

    To verify \eqref{eq:miracle}, we first compute (see e.g. \href{\detokenize{https://www.wolframalpha.com/input?i=expand+%281%2F2%29*%28x-1%29%5E2*%28x-2%29%5E2%2F4+%2B+%281%2F2%29*x*%28x-1%29%5E2+-+%283%2F8%29*x*%28x-1%29*%28x-2%29*%28x-3%29+%2B+x*%28x-1%29*%28x-3%29*%28x-4%29%2F4}}{Wolfram Alpha}) that 
    \[
    F(k)
    +
    \frac{k(k-1)(k-3)(k-4)}{4}
    =
    \frac{5k^2}{4}-\frac{7k}{4}+\frac{1}{2}.
    \]
    Meanwhile 
    \[
    \frac{(T_v-A_v)^2}{2}
    -
    \binom{T_v-A_v-2}{2}
    =
    \frac{5}{2}(T_v-A_v)-3.
    \]
    Finally 
    \[
    \sum_{u\sim v} (d_v-d_u)
    =
    d_v^2 
    -
    \sum_{u\sim v}(x_u+1)
    =
    d_v^2 - T_v - d_v.
    \]
    Thus \eqref{eq:miracle} is equivalent to:
    \[
    \frac{5d_v^2}{4}-\frac{7d_v}{4}+\frac{1}{2}
    \stackrel{?}{=}
    \frac{5}{2}(d_v^2 - T_v - d_v)
    -3d_v+6
    +
    \frac{5}{2}(T_v-A_v)-3.
    \]
    The $T_v$ terms cancel, and grouping the remaining terms (without yet expanding $A_v$) yields the equivalent:
    \begin{equation}
    \label{eq:equivalent} 
    -\frac{5d_v^2}{4}+\frac{15d_v}{4}-\frac{5}{2}
    \stackrel{?}{=}
    -\frac{5 A_v}{2}.
    \end{equation}
    Finally, expanding
    \[
    A_v=\binom{d_v-1}{2}=\frac{d_v^2 - 3d_v + 2}{2}
    \]
    one finds that all terms in \eqref{eq:equivalent} match.
    This completes the proof.

\subsection{Proof of Conjecture \ref{conj:main}} \label{sec:two}

As mentioned above, the work \cite{bubeck2016local} actually made an infinite sequence of conjectures parametrized by a positive integer $C\geq 2$ (each corresponding to a face of the convex set of achievable tree profiles), with the general statement that 
\[
C(2C^2+7C+7)Y-C(C+1)(C^2+3C+4)P-24(C+1)(C+2)S\leq O(1)
\]
for any fixed $C$.
In addition to confirming the case $C=1$, \cite{bubeck2016paths} actually \emph{disproved} this for $C\geq 3$. The same proof-oriented scaffolding on top of GPT-5 was able to confirm the remaining $C=2$ case is true. 
The proof uses similar ideas but this time requires the non-positivity of a quite complicated non-polynomial function on the integers.

\begin{figure}
\begin{chatbox}
  \begin{userturn}
[Theorem \ref{thm:main-bsy} and Section \ref{sec:one} are put in context.]
\newline

More generally, we can state the inequality above like so \[ C(C+1)(C^2 + 3C + 4)\,P + 24(C+1)(C+2)\,S - C(2C^2 + 7C + 7)\,Y \ge \texttt{cst} \] where the implied constant does not depend on $G$. The case $C=1$ recovers the inequality above. Your task is to now prove or disprove the case for $C=2$.
  \end{userturn}

%   \begin{assistantturn}
% ()
% \end{assistantturn}
\end{chatbox}
\caption{Question posed to (a scaffolded version of) GPT-5 to prove Conjecture \ref{conj:main}.
The answer it provided is written in Section \ref{sec:two}, with minor edits for clarity.}
   \label{fig:trees_2}
\end{figure}

\begin{theorem}
\label{thm:C=2}
For any tree $G$, one has 
\[
29Y-42P-144S\leq 504.
\]
\end{theorem}

\begin{proof}
    We will double the inequality to be proved to conveniently reduce the number of non-integer terms.
    Using the same formulas for $S,P,Y$ as above, we see that 
    \[
    E:= 58Y-84P-288S
    =
    \sum_v 
    \big(58A_vT_v-42T_v^2\big)
    +
    \sum_v \Big(42d_v (d_v-1)^2 - 288\binom{d_v}{4}\Big).
    \]
    We will rewrite this using the identities:
    \begin{align*}
    58 A_v T_v - 42T_v^2 
    &= 
    -29(T_v-A_v)^2 - 13T_v^2 + 29A_v^2,
    \\
    -29(T_v-A_v)^2
    &=
    -58\binom{T_v-A_v-3}{2}-203(T_v-A_v)+348,
    \\
    -13T_v^2 &= -26\binom{T_v-3}{2}-91T_v+156.
    \end{align*}
    Substituting gives:
    \[
    E=
    \sum_v 
    \Big(-58\binom{T_v-A_v-3}{2}-26\binom{T_v-3}{2}-294T_v + L(d_v)\Big)
    \]
    where 
    \[
    L(d)=42d(d-1)^2-288\binom{d}{4}+29\binom{d-1}{2}^2+203\binom{d-1}{2}+504.
    \]
    Here we used $203+91=294$ and $348+156=504$.
    Similarly we above we have 
    \[
    \sum_{u\sim v}(d_v-d_u)
    =
    d_v^2-T_v-d_v
    \]
    and so 
    \[
    -294T_v=294\sum_{u\sim v}(d_v-d_u)-294(d_v^2-d_v).
    \]
    The first term on the right-hand side telescopes to $0$ from summing over $v\in G$. 
    Thus we find that 
    \[
    E
    =
    \sum_v
    \Big(-58\binom{T_v-A_v-3}{2}-26\binom{T_v-3}{2} + J(d_v)\Big)
    \]
    where we have (again see 
    \href{\detokenize{https://www.wolframalpha.com/input?i=expand+42*d*%28d-1%29%5E2+-+288*binomial%28d%2C4%29+%2B+29*binomial%28d-1%2C2%29%5E2+%2B+203*binomial%28d-1%2C2%29+%2B+504+-294%28d%5E2-d%29}}{Wolfram Alpha}) that
    \[
    J(d)=L(d)-294(d^2-d)
    =
    -\frac{19}{4}d^2+\frac{141}{2}d^3-\frac{1257}{4}d^2+\frac{33}{2}d+736.
    \]
    The next step is to maximize each binomial coefficient binomial coefficient above, holding $A_v$ fixed and letting $T_v$ vary over the integers, so that each term in $E$ becomes a univariate function of $d$ (since $A=\binom{d-1}{2}$).
    Writing $Y_v=T_v-3$, we aim to minimize the quantity
    \[
    58\binom{T_v-A_v-3}{2}+26\binom{T_v-3}{2}
    =
    42Y_v^2 - (58A_v+42)Y_v+29A_v(A_v+1).
    \]
    (We are trying to upper bound $E$, but the above quantity is subtracted within $E$.) The minimum is attained at $Y_v=\frac{29A_v+21}{42}$ with value 
    \[
    29A_v(A_v+1)-\frac{(29A_v+21^2)}{42}.
    \]
    However since $T_v$ and thus $Y_v$ must be integral, we actually find that 
    \[
    58\binom{T_v-A_v-3}{2}+26\binom{T_v-3}{2}
    \geq 
    29A_v(A_v+1)-\frac{(29A_v+21^2)}{42}
    +
    \frac{s^2}{42}
    \]
    where $s=\min(r,42-r)$ and $r=29A+21\pmod{42}$.
    (I.e. $s/42$ is the distance from $Y_v=\frac{29A_v+21}{42}$ to the nearest integer.)
    Note that $A$, hence $r$, hence $s$ is a function of $d_v$.
    
    Combining the above, we have bounded $E$ from above by a sum of univariate functions of $d_v$:
    \[
    E
    \leq 
    \sum_{v\in G}
    J(d_v)-m(d_v)
    \]
    where $J$ was defined above and $m(d_v)=29A_v(A_v+1)-\frac{(29A_v+21^2)}{42}
    +
    \frac{s^2}{42}$.
    Since $\sum_{v} (d_v-2)=-2$ in a tree, this means 
    \[
    E-1008
    \leq 
    \sum_{v\in G}
    \Big(J(d_v)-m(d_v)+504(d_v-2)\Big).
    \]
    It remains to show each term on the right-hand side is at most $0$ for any value of $d_v$.
    In fact this does hold, with equality for all $d_v\in \{1,2,4,5\}$.
    This can be verified by an explicit exact evaluation\footnote{If one drops the non-analytic $s^2/42$ term, the resulting (larger) quartic polynomial alternates signs on the sequence $(0,1,3,4,6)$ and is negative at $6$. Since it can have only $4$ roots, this implies it is negative for all $d_v\geq 6$. In particular, this means plugging in the values $\{0,1,2,3,4,5,6\}$ suffices to confirm the numerical claim. See also this \href{\detokenize{https://www.desmos.com/calculator/mpubmlfcdp}}{Desmos} plot.}, and completes the proof.
\end{proof}

%% file: content/11.5-BubeckSellkeYin.tex
\section{COLT problem on dynamic networks  -- Sébastien Bubeck, Mark Sellke, and Steven Yin}
\label{sec:\secprefix-115}

We consider a random tree $(G_t)_{t\ge1}$ constructed iteratively as follows. At time $t=1$ we begin with a single vertex. At each subsequent time $t+1$, a new vertex $Y_{t+1}$ is added and attaches to an existing parent vertex $X_{t+1}\in\{Y_1,\dots,Y_t\}$ chosen by a randomized preferential rule described below.

Each vertex $v$ independently receives a permanent ``attractiveness'' label $a(v)\in\{1,w\}$ with equal probability for the two options.
The \emph{attractiveness-weighted degree} of $x\in V_t$ is
\[
A_t(x) := \sum_{y\sim_t x} a(y).
\]
(Here $\sim_t$ denotes adjacency at time $t$.)
The parent $X_{t+1}$ of $Y_{t+1}$ is chosen according to
\[
\mathbb P(X_{t+1}=x\mid \mathcal F_t) = \frac{A_t(x)}{S_t},
\qquad
S_t := \sum_{x\in V_t} A_t(x).
\]
Here $\mathcal F_t=\sigma(G_1,\dots,G_t)$ is the associated filtration.

\begin{figure}[h]
\centering 
\begin{tikzpicture}[
  node1/.style={circle,draw,thick,fill=white,minimum size=12mm},
  nodew/.style={circle,draw,thick,fill=blue!15,minimum size=12mm},
  newNode/.style={circle,draw,thick,dashed,fill=gray!10,minimum size=12mm},
  probarrow/.style={-Latex, dashed, semithick},
  font=\small
]

% Existing tree
\node[nodew] (A) at (0,1.5) {$w$};
\node[node1] (B) at (-1.5,0) {$1$};
\node[nodew] (C) at (1.5,0) {$w$};

\draw[thick] (A) -- (B);
\draw[thick] (A) -- (C);

% Attractive degree labels
\node[above=1mm of A] {$1+w$};
\node[left=1mm of B] {$w$};
\node[right=1mm of C] {$w$};

% New arriving node
\node[newNode] (D) at (0,-2) {};

% Dashed arrows suggesting attachment probabilities
\draw[probarrow] (D) to[bend left=20] (A);
\draw[probarrow] (D) to[bend right=10] (B);
\draw[probarrow] (D) to[bend left=10] (C);

\end{tikzpicture}
\label{fig:weightedpa}
\caption{Illustration of the process at $t=3$ with two vertices labelled $w$ and one vertex labelled $1$. The attractiveness-weighted degree $A_t$ of each of the existing vertices is displayed next to it. Up to normalization, these are the attachment probabilities for the new fourth vertex. TikZ figure produced by GPT-5.}
\end{figure}
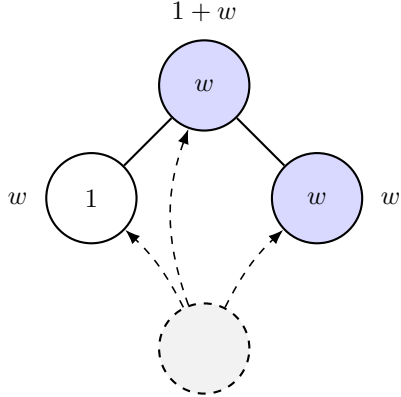

Our goal will be to approximately recover $w$ based on $G_t$ for a single, large value of $t$ \emph{without observing the vertex labels}. 
This was posed as an open problem\footnote{Importantly we note that this is the first COLT open problem that we tried with our scaffolded GPT-5. We chose this paper somewhat randomly, with some preference for it due to the authors' familiarity with preferential attachment processes.} at the 2012 Conference on Learning Theory in \cite{colt_open_problem}, who observed that single-edge label statistics do not help determine $w$ (in fact their question asks for a bit more, see Remark~\ref{rem:colt-problem-details}).
This type of preferential attachment model was introduced in the seminal work \cite{barabasi1999emergence} (in the case $w=1$), with the aim of better modeling the emergence in real networks and explaining observed heavy-tailed behavior. 
Recovering $w$ amounts, in this simplified toy model, to a qualitatively interesting achievement: one has inferred the \emph{dynamical} behavior of an evolving network while only \emph{observing} the network at just $1$ time.
Note that there is a redundancy in the model above: reciprocating $w$ does not change the law of $(G_t)_{t\geq 1}$ because one can reverse all the labels.
Thus one should restrict to $w\geq 1$, and we correspondingly posed the following question to (a scaffolded version of) GPT-5 (see Figure~\ref{fig:colt}).

\begin{figure}
\begin{chatbox}
  \begin{userturn}
Consider a modified preferential attachment tree process, where each vertex is also born with a label in $\{0,1\}$, chosen uniformly at random. The ``attractiveness" of a vertex with label $1$ is $w>0$ and for a label $0$ it is $1$. 
The ``attractive degree" of a vertex is defined as the sum of attractiveness of its neighbors (so if $w=1$ this is simply the usual degree). When a new vertex is added, it connects to a random existing vertex with probability proportional to their attractive degree.
Based upon an observation of a large unlabelled tree generated by this process, is it possible to identify $w$? Either give an algorithm to do so, or prove an information theoretic impossibility.
  \end{userturn}
\end{chatbox}

\caption{Question posed to (a scaffolded version of) GPT-5. It solved this problem by coming up with and proving Theorem \ref{thm:colt}, and observing that this implies Corollary~\ref{cor:colt}. (In particular the decision to study the quantity $L(t)$ required no human input!) The proofs are provided below with minor edits.}
   \label{fig:colt}
\end{figure}

\begin{figure}
    \centering
    \includegraphics[width=\textwidth]{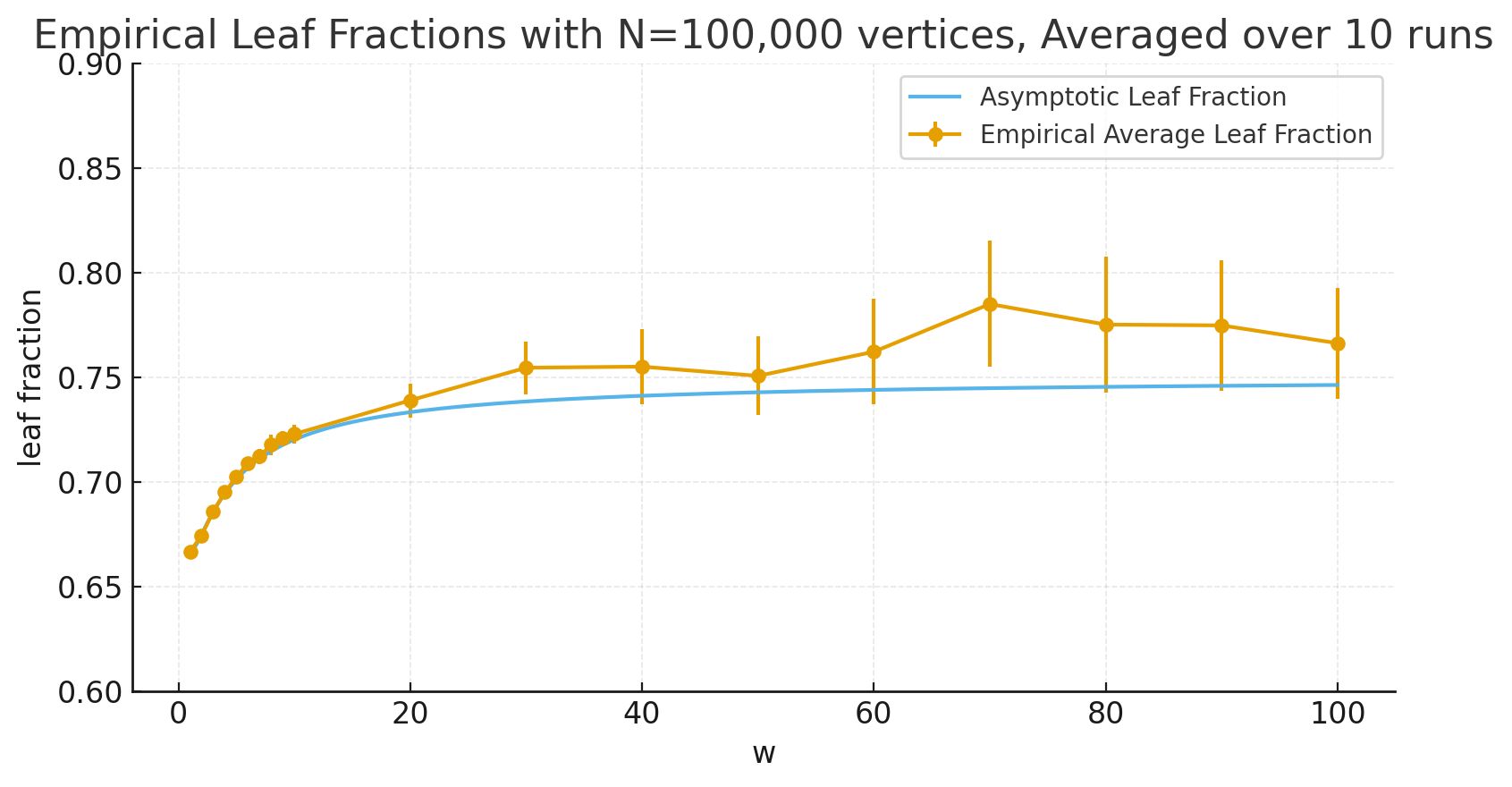}
    \caption{A simulation of the asymptotic result \eqref{eq:limiting-leaf-frac} in Theorem~\ref{thm:colt}.
    The orange plot shows the average (and empirical standard deviation) of the fraction of leaves in $G_{10^5}$, across $10$ independent trials for each $w$. 
    The function $f(w)$ is graphed in blue on $w\in [1,\infty)$, where it strictly increases from $f(1)=2/3$ to $\lim_{w\to\infty} f(w)=3/4$.
    The code used to generate this plot was written by GPT-5.
    }
    \label{fig:colt-v2}
\end{figure}

GPT-5 was able to prove that $w$ is indeed identifiable, stated in Corollary~\ref{cor:colt} below.
This is achieved by way of Theorem~\ref{thm:colt} below, which computes the limiting fraction $f(w)$ of vertices which are leaves at a given large time.
To state the result, let $L_1(t)$ (resp.\ $L_w(t)$) be the number of leaves at time $t$ whose parent (unique neighbor) has attractiveness $1$ (resp.\ $w$).
Then the total number $L(t)=L_1(t)+L_w(t)$ of leaves at time $t$ is observable without observing the labels in $G_t$ (while $L_1(t)$ and $L_w(t)$ are not).

\begin{theorem}
\label{thm:colt}
For any $w>0$, one has almost surely:
\begin{equation}
\label{eq:limiting-leaf-frac}
\lim_{t\to\infty}
\frac{L(t)}{t}
=
f(w)\equiv \frac{3}{4}-\frac{1}{2(w+2)}+\frac{1}{4(2w+1)} 
=
\frac{3(w+1)^2}{2(w+2)(2w+1)}.
\end{equation}
In fact, one has 
\[
\lim_{t\to\infty}
\frac{L^{(1)}(t)}{t}
=
\frac{w+1}{2(w+2)},
\qquad
\lim_{t\to\infty}
\frac{L^{(w)}(t)}{t}
=
\frac{w+1}{2(2w+1)}.
\]
\end{theorem}

The function $f$ defined in \eqref{eq:limiting-leaf-frac} is easily seen to be strictly increasing, with 
\[
f'(w)=
\frac{1}{2}
\bigg(
\frac{1}{(w+2)^2}-
\frac{1}{(2w+1)^2}
\bigg)
>0,\quad\forall w>1.
\]
Identifiability of $w$ is then immediate, as in the following Corollary.

\begin{corollary}
\label{cor:colt}
For any $w\geq 1$ the point estimate $\hat w(t)=f^{-1}(L(t)/t)$
converges to $w$ in probability as $t\to\infty$. Here $f^{-1}:[2/3,3/4]\to [1,\infty]$ is the inverse of $f|_{[1,\infty]}$, and we set $f^{-1}(x)=1$ for $x\leq 2/3$ and $f^{-1}(x)=\infty$ for $x\geq 3/4$.
\end{corollary}

It is worth noting that the above type of algorithm provably fails in a related model of evolving trees studied in \cite{ben2025inference} (in which attachment probabilities depend only on label agreement and not on degree). In their model, the fraction of degree $k$ vertices is asymptotically $2^{-k}$ for all parameter values. This illustrates that GPT-5's decision to focus on the quantity $L(t)$ is already non-obvious.

\begin{remark}
\label{rem:colt-problem-details}
\cite{colt_open_problem} in fact asks for a polynomial rate of convergence, i.e. an estimator $\hat w(t)$ satisfying $\mathbb P[|\hat w(t)-w|\leq \varepsilon]\geq 1-\delta$ for $t$ polynomially large in $(\varepsilon,\delta)$. 
We believe a more precise implementation of our approach suffices to show this stronger statement, but do not pursue this here.
\cite{colt_open_problem} also considers a slightly more general setting in which incoming vertices have probability $p$ to be labeled $w$ (which removes the convenient $w\leftrightarrow 1/w$ symmetry).
\end{remark}

\begin{remark}
    While all major proof ideas below are due to GPT-5, a few details of proof writing are human-supplied.
    Our scaffolded version of GPT-5 autonomously produced exactly the high-level argument below and the same main references, but omitted the proof details for certain intermediate assertions, stating (correctly) that they followed from routine stochastic approximation arguments.
    When we asked (an unscaffolded) GPT-5 to provide more detail for these latter arguments, it made several false starts, stating for example that \cite[Theorem 1]{robbins1971convergence} directly implies $V_t\to 0$ almost surely in the proof of Lemma~\ref{lem:Rt-limit} below, while it in fact only gives convergence of $V_t$.
    After some human pushback, GPT-5 eventually came up with a correct but unnecessarily complicated proof; we chose to present a more streamlined argument (by observing that $\mathbb E[V_t]\to 0$).
\end{remark}

\subsection{Proof Overview}

We assume $w\geq 1$ without loss of generality, and treat $w$ as constant in all $O(\cdot)$ notation. Let
\[
S^{(1)}_t := \sum_{x: a(x)=1} A_t(x),
\qquad 
S^{(w)}_t := \sum_{x: a(x)=w} A_t(x).
\]
Then $S^{(1)}_t+S^{(w)}_t=S_t$, and 
\[
R_t\equiv \frac{S^{(w)}_t}{S_t}=\mathbb P[a(X_{t+1})=w]
\]
is the chance for the next vertex to attach to a $w$-labelled vertex in $G_t$.
Further, let 
\begin{equation}
D_t := S^{(w)}_t - \tfrac12 S_t = \frac{S_t^{(w)}-S_t^{(1)}}{2} = S_t\cdot(R_t-\tfrac{1}{2}),\qquad 
\mu = \mathbb E[a(\text{new vertex})] = \frac{1+w}{2}
\end{equation}
and set (with $E_t=E(G_t)$ the set of edges at time $t$) 
\[
T_t:=\sum_{\{u,v\}\in E_t} a(u)a(v)
=
\frac{S_t^{(1)}+wS_t^{(w)}}{2}.
\]
Then
\begin{equation}
\label{eq:expected-parent-label}
\mathbb E[a(X_{t+1})\mid\mathcal F_t] = \frac{2T_t}{S_t}.
\end{equation}
For notational convenience, given a discrete-time stochastic process $(Z_t)_{t\geq 1}$ we write $\Delta Z_t=Z_{t+1}-Z_t$.

The proof consists of two main steps.
In the first we show $R_t\to 1/2$ and $S_t/t\to w+1$ almost surely (see Lemmas~\ref{lem:Rt-limit} and \ref{lem:St/t}).
In the second we use this to analyze the growth of $L^{(1)}(t)$ and $L^{(w)}(t)$.
Both steps and especially the second employ classical stochastic approximation results in the style of \cite{robbins1951stochastic,robbins1971convergence}.

\subsection{One-Step Dynamics}

When the $(t+1)$-th vertex $y=Y_{t+1}$ attaches to the random vertex $x=X_{t+1}\in \{Y_1,\dots,Y_t\}$, we have
\[
\Delta A_t(x)\equiv A_{t+1}(x)-A_t(x) = a(y),\qquad
\Delta A_t(y)=a(x),
\]
and $A_t(z)$ is unchanged for other vertices $z\in \{Y_1,\dots,Y_t\}\backslash \{x\}$. Consequently
\[
\Delta S_t = a(x)+a(y).
\]
Using \eqref{eq:expected-parent-label} one obtains with some algebra that
\begin{equation}
\label{eq:DeltaS-expectation}
\mathbb E[\Delta S_t\mid\mathcal F_t]
= \mu + \frac{2T_t}{S_t}
=
\frac{D_t(w-1)}{S_t} + (w+1).
\end{equation}
Likewise,
\[
\Delta S^{(w)}_t
= \mathbf1\{a(x)=w\}a(y)+\mathbf1\{a(y)=w\}a(x),
\]
and one can easily compute
\begin{equation}
\label{eq:DeltaD-expectation}
\mathbb E[\Delta D_t\mid \mathcal F_t]
= \mu\big(R_t - \tfrac12\big) 
=
\frac{\mu D_t}{S_t}.
\end{equation}
We also trivially have
\begin{equation}
\label{eq:linear-bounds-S}
2t \le S_t\le 2wt
\end{equation}
and $|\Delta R_t|\leq O(1/S_t)$.

\subsection{Convergence of $R_t$ and $S_t$}

\begin{lemma}
\label{lem:Rt-limit}
$\lim_{t\to\infty} R_t=\tfrac12$ almost surely.
\end{lemma}

\begin{proof}
Define 
\[
V_t:=(R_t-\tfrac12)^2
=
(D_t/S_t)^2.
\]
We will show $V_t\to 0$ holds almost surely.
Letting $Q_t=D_t/S_t$, one has
\[
\Delta V_t
=
2Q_t\Delta Q_t
+
(\Delta Q_t)^2.
\]
Next it is easy to see by Taylor's approximation theorem that the quotient rule holds approximately for discrete differences; since $|\Delta S^{(w)}_t|, |\Delta S_t|\leq O(1)$ we find:
\begin{align*}
    \Delta Q_t
    &=
    \frac{
    S_t\Delta D_t
    -
    D_t\Delta S_t
    }{S_t^2}
    +
    O(S_t^{-2}).
\end{align*}
Combining the previous two displays, we find:
\[
\Delta V_t
=
\frac{2D_t}{S_t^2}\Big(\Delta D_t - \frac{D_t \Delta S_t}{S_t}\Big)
+ O(S_t^{-2}).
\]
Using \eqref{eq:DeltaD-expectation}--\eqref{eq:DeltaS-expectation} yields
\[
\mathbb E[\Delta V_t\mid\mathcal F_t]
=
-\frac{4D_t^2 T_t}{S_t^4}
+ O(S_t^{-2})
=
-V_t \Big(\frac{4T_t}{S_t^2}\Big)+ O(S_t^{-2})
.
\]
Since $T_t\asymp S_t\asymp t$, we get 
\[
\mathbb E[\Delta V_t\mid\mathcal F_t]
\le -\Omega(S_t^{-1})V_t + O(S_t^{-2}).
\]
By \eqref{eq:linear-bounds-S}, we have 
\[
\sum_{t\geq 1} S_t^{-1}=\infty,\qquad
\sum_{t\geq 1} S_t^{-2}=\bar S(w)<\infty.
\]
The almost-supermartingale theorem of \cite[Theorem 1]{robbins1971convergence} implies that $\lim_{t\to\infty} V_t$ exists almost surely.
It remains to identify the limit.
For this, letting $E_t=\mathbb E[V_t]$, we obtain
\[
E_{t+1}\leq (1-cS_t^{-1})E_t + O(S_t^{-2}).
\]
for a uniform constant $c=c(w)>0$.
Then
\[
E_T
\leq 
(E_1+\bar S(w))
\cdot
\prod_{1\leq t<T}
(1-cS_t^{-1}).
\]
This tends to $0$ as $T\to\infty$ because $\sum S_t^{-1}=\infty$.
We conclude that $E_t\to 0$.
Since $V_t\in [0,1]$ for all $t$, its almost-sure $t\to\infty$ limit must therefore equal $0$.
This completes the proof.
\end{proof}

\begin{lemma}
\label{lem:St/t}
$S_t/t\to w+1$ almost surely.
\end{lemma}

\begin{proof}
Recalling \eqref{eq:DeltaS-expectation}, we have
\[
\Delta S_t = \frac{D_t(w-1)}{S_t} + (w+1)+\xi_{t+1}
\]
for a bounded martingale difference $\xi_{t+1}$. Since $D_t/S_t\to 0$ from the previous lemma, Cesàro averaging yields
\[
\frac1t\sum_{i=1}^t \Big(\frac{D_t(w-1)}{S_t} + (w+1)\Big)\to w+1.
\]
The martingale SLLN gives $\frac1t\sum_{i=1}^t \xi_{i}\to0$ almost surely. Thus $S_t/t\to w+1$.
\end{proof}

\subsection{Stochastic Approximation for Leaf Fractions}

We use the following stochastic approximation result, which gives an ODE limit for certain stochastic processes.
It follows by combining \cite[Propositions 4.1 and 4.2]{BenaimSPS1999} (using $q=2$ therein), and specializing to a $1$-dimensional state space.

\begin{theorem}
\label{thm:SA-ODE}
Consider a scalar recursion 
\[
Z_{t+1}=Z_t+\gamma_{t+1}(F(Z_t)+\varepsilon_{t+1}+\delta_{t+1})
\]
driven by:
\begin{itemize}
    \item Deterministic step sizes satisfying 
\[
\gamma_t\to0,\qquad \sum_{t\geq 1}\gamma_t=\infty,\qquad \sum_{t\geq 1}\gamma_t^2<\infty.
\]
\item A martingale-difference sequence $(\varepsilon_t)_{t\geq 1}$ adapted to a filtration $\mathcal F_t$ (i.e. $\mathbb E[\varepsilon_{t+1}|\mathcal F_t]=0$) and satisfying $\mathbb E[\varepsilon_{t+1}^2|\mathcal F_t]\leq C$ for a deterministic constant $C$.
\item 
Asymptotically vanishing adapted perturbations: $\delta_t\to 0$ almost surely.
\item 
Locally Lipschitz $F:\mathbb R\to\mathbb R$.
\end{itemize}
Suppose the ODE $z'(T)=F(z(T))$ has a unique globally attracting equilibrium $z^{\star}$, i.e. for all initial conditions $z(0)\in\mathbb R$ one has $\lim_{T\to\infty} z(T)=z^{\star}$ for the solution to this ODE.
Then almost surely the analogous convergence holds for the discrete system: 
\[
\lim_{t\to\infty} Z_t= z^\star.
\]
\end{theorem}

Recall the definitions of $L_1(t),L_w(t)$ just above Theorem~\ref{thm:colt}. 
It is easy to see that
\begin{equation}
\label{eq:leaf-drifts}
\mathbb E[\Delta L_1\mid\mathcal F_t]
= \frac{S_t^{(1)}-L_1(t)}{S_t},
\qquad
\mathbb E[\Delta L_w\mid\mathcal F_t]
= \frac{S_t^{(w)}-w L_w(t)}{S_t}.
\end{equation}
Indeed $L_1(t)/S_t$ and $L_w(t)/S_t$ are the conditional probabilities (given $\mathcal F_t)$ for $X_{t+1}$ to be a leaf of label $1$ and $w$.
On these events $X_{t+1}$ is no longer a leaf in $G_{t+1}$.

The main quantities of interest will be the leaf fractions
\[
x_t=L_1(t)/t,\qquad y_t=L_w(t)/t.
\]
Since e.g.\ $x_{t+1}=\frac{t x_t}{t+1}$ when $\Delta L_1(t)=0$, we have:
\begin{equation}
\label{eq:leaf-ratio-drifts}
\Delta x_t
= \frac{\Delta L_1(t) - x_t}{t+1},
\qquad
\Delta y_t
= \frac{\Delta L_w(t) - y_t}{t+1}.
\end{equation}
We showed in Lemma~\ref{lem:Rt-limit} and \ref{lem:St/t} above that almost surely
\begin{equation}
\label{eq:summarize-stage-1}
\Big(\frac{S_t^{(1)}}{t},\frac{S_t^{(w)}}{t},\frac{S_t}{t}\Big)
\to 
\Big(\frac{w+1}{2},\frac{w+1}{2},w+1\Big).
\end{equation}
In light of \eqref{eq:leaf-drifts} and \eqref{eq:leaf-ratio-drifts}, we will apply Theorem~\ref{thm:SA-ODE} to $x_t$ and $y_t$ separately, with 
$\gamma_{t+1}=1/(t+1)$ each in case.
We use the vector fields
\[
F_1(x)=\frac12 - x - \frac{x}{1+w},
\qquad
F_w(y)=\frac12 - y - \frac{wy}{1+w}.
\]
and the martingale noises
\[
\varepsilon_{t+1}^X:=\Delta L_1(t) - \mathbb E[\Delta L_1(t)\mid\mathcal F_t],
\qquad
\varepsilon_{t+1}^Y:=\Delta L_w(t) - \mathbb E[\Delta L_w(t)\mid\mathcal F_t].
\]
Focusing without loss of generality on the $x_t$ case, we obtain the recursion
\[
x_{t+1} = x_t + \gamma_{t+1}(F_1(x_t)+\varepsilon_{t+1}^X+\delta_{t+1}^X),
\]
where $\delta_{t}^X$ is defined to make equality hold, and one has $\delta_{t}^X\to0$ almost surely as a consequence of \eqref{eq:summarize-stage-1}.
Since $|\Delta L_1|\leq 1$ almost surely, the required conditions on $\varepsilon_t^X$ are clear.
It is also easy to see that the linear ODE $x'(T)=F_1(x(T))$ has a unique stable, globally attracting fixed point, which is the zero of $F_1$ given by
\[
x^{\star}=\frac{w+1}{2(w+2)}.
\]
Thus we may apply Theorem~\ref{thm:SA-ODE} to conclude the almost sure limit:
\[
x_t/t \to 
\frac{w+1}{2(w+2)}.
\]
Identical reasoning shows that $y_t$ converges almost surely to the unique zero of $F_w$, i.e.
\[
y_t/t\to 
y^{\star}=\frac{w+1}{2(2w+1)}.
\]
This completes the proof of Theorem~\ref{thm:colt}.

%% file: content/99-Conclusion.tex
\chapter*{Conclusion}
\addcontentsline{toc}{chapter}{Conclusion}

AI models are well-known for assisting with routine tasks; what is less well-known is that frontier models like GPT-5 are increasingly able to accelerate novel research. We highlighted examples across a range of scientific disciplines showing GPT-5 assisting in ideation, powering deep literature search, and even proposing and implementing proofs of (appropriately-sized) open problems. None of this would have been possible just twelve months ago. The fact that GPT-5 is already showing it can accelerate scientific research, combined with the rate of improvement of frontier AI, suggests that the coming years will be an exhilarating time in science and mathematics.

\clearpage

\section*{Acknowledgments}

A. Lupsasca is supported in part by the National Science Foundation through the CAREER award 2340457 and grant 2307888, and by the Simons Foundation through the grant SFI-MPS-BH-00012593-09. This research was conducted during the period M. Sawhney served as a Clay Fellow. C. Coester is funded by the European Union (ERC, CCOO, 101165139). Views and opinions expressed are however those of the author(s) only and do not necessarily reflect those of the European Union or the European Research Council. Neither the European Union nor the granting authority can be held responsible for them. D. Unutmaz thanks Oral Alpan, M.D. for critical reading of Section 3 of this report.